\documentclass[11pt,a4paper]{article}
\usepackage[acceptedWithA]{tacl2021v1}
\usepackage{xspace,mfirstuc,tabulary}
\usepackage{times,latexsym}
\usepackage{url}
\usepackage[T1]{fontenc}
\usepackage{todonotes}

\usepackage{scrextend}
\usepackage{colortbl}
\usepackage{rotating}

\usepackage{graphicx}
\usepackage{booktabs}
\usepackage{multirow}
\usepackage{arydshln}
\usepackage{comment}
\usepackage{amssymb}
\usepackage{amsmath}
\usepackage{bclogo}
\usepackage[skins]{tcolorbox}
\usepackage{rotating}
\usepackage{graphicx}
\definecolor{papercolor}{RGB}{0,168,142}
\definecolor{lightgray}{gray}{0.9}

\makeatletter
\def\adl@drawiv#1#2#3{%
        \hskip.5\tabcolsep
        \xleaders#3{#2.5\@tempdimb #1{1}#2.5\@tempdimb}%
                #2\z@ plus1fil minus1fil\relax
        \hskip.5\tabcolsep}
\newcommand{\cdashlinelr}[1]{%
  \noalign{\vskip\aboverulesep
           \global\let\@dashdrawstore\adl@draw
           \global\let\adl@draw\adl@drawiv}
  \cdashline{#1}
  \noalign{\global\let\adl@draw\@dashdrawstore
           \vskip\belowrulesep}}
\makeatother

\newcommand\Holmes{\texttt{Holmes}}
\newcommand\FlashHolmes{\texttt{FlashHolmes}}

\newif\iftaclinstructions
\taclinstructionsfalse 
\iftaclinstructions
\renewcommand{\confidential}{}
\renewcommand{\anonsubtext}{(No author info supplied here, for consistency with
TACL-submission anonymization requirements)}
\newcommand{\instr}
\fi

\iftaclpubformat 

\else

\fi

\title{\texttt{Holmes} \bcloupe\\ A Benchmark to Assess the Linguistic Competence of Language Models}

\author{Andreas Waldis\thanks{* Corresponding author \href{andreas.waldis@live.com}{andreas.waldis@live.com}} $^{1,2}$,
Yotam Perlitz$^{3}$,
Leshem Choshen$^{4,5}$,
Yufang Hou$^{6}$, Iryna Gurevych$^{1}$\\
$^1$Ubiquitous Knowledge Processing Lab (UKP Lab), 
Technical University of Darmstadt\\
 $^2$Information Systems Research Lab, Lucerne University of Applied Sciences and Arts \\
 $^3$IBM Research AI, $^4$MIT CSAIL, $^5$MIT-IBM Watson AI Lab, $^6$IBM Research Europe - Ireland\\
\texttt{\href{http://www.ukp.tu-darmstadt.de/}{www.ukp.tu-darmstadt.de}} \hspace{0.5em} \texttt{\href{http://www.hslu.ch/}{www.hslu.ch}}\\
}

\date{}

\begin{document}
\maketitle
\begin{abstract}
We introduce \Holmes{}, a new benchmark designed to assess language models (LMs) \textit{linguistic competence} - their unconscious understanding of linguistic phenomena.
Specifically, we use classifier-based probing to examine LMs' internal representations regarding distinct linguistic phenomena (e.g., part-of-speech tagging).
As a result, we meet recent calls to disentangle LMs' linguistic competence from other cognitive abilities, such as following instructions in prompting-based evaluations.
Composing \Holmes{}, we review over 270 probing studies and include more than 200 datasets to assess \emph{syntax, morphology, semantics, reasoning,} and \emph{discourse} phenomena.
Analyzing over 50 LMs reveals that, aligned with known trends, their linguistic competence correlates with model size.
However, surprisingly, model architecture and instruction tuning also significantly influence performance, particularly in \textit{morphology} and \textit{syntax}.
Finally, we propose \FlashHolmes{}, a streamlined version that reduces the computation load while maintaining high-ranking precision.
\end{abstract}
\vspace{0em}
\includegraphics[width=2em,height=2em]{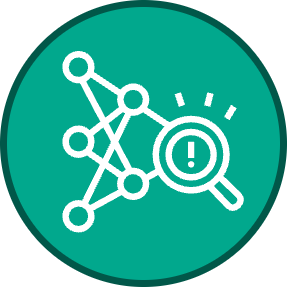}\hspace{.75em}\parbox{15em}{\href{https://holmes-benchmark.github.io}{\vspace*{1.3em}\texttt{holmes-benchmark.github.io}}
}
\vspace{-.5em}

\section{Introduction}
\label{sec:introduction}

Linguistic competence is the unconscious understanding of language \citep{chomsky1965}, like the syntactic structure of a sentence.
As language models (LMs) are trained on simple tasks such as next word prediction \citep{Brown2020LanguageMA}, one might naturally wonder: \textit{What is the linguistic competence of LMs, and how do they differ?}
To answer such questions, contemporary benchmarks estimate cognitive abilities, as done for mathematical reasoning~\citep{cobbe2021gsm8k} or factual knowledge~\citep{DBLP:conf/emnlp/PetroniRRLBWM19,petroni2020how}.
However, such benchmarks rely on LMs' \textit{use of language} (textual responses) known as linguistic performance \citep{Matthews1998TheCO}.
As a result, they conflate abilities tested with specific instructions, as done for syntactic phenomena in \citet{blevins-etal-2023-prompting2}, with latent abilities like producing coherent text or following instructions.
As this entanglement makes it infeasible to draw definitive conclusions \citep{hu-levy-2023-prompting,liang2023holistic,perlitz2024benchmarkagreementtestingright}, recent studies call to assess LMs' linguistic competence comprehensively and isolated \citep{lu2023emergent, mahowald2023dissociating}.

\begin{figure}[]
    \centering
    \includegraphics[width=0.48\textwidth]{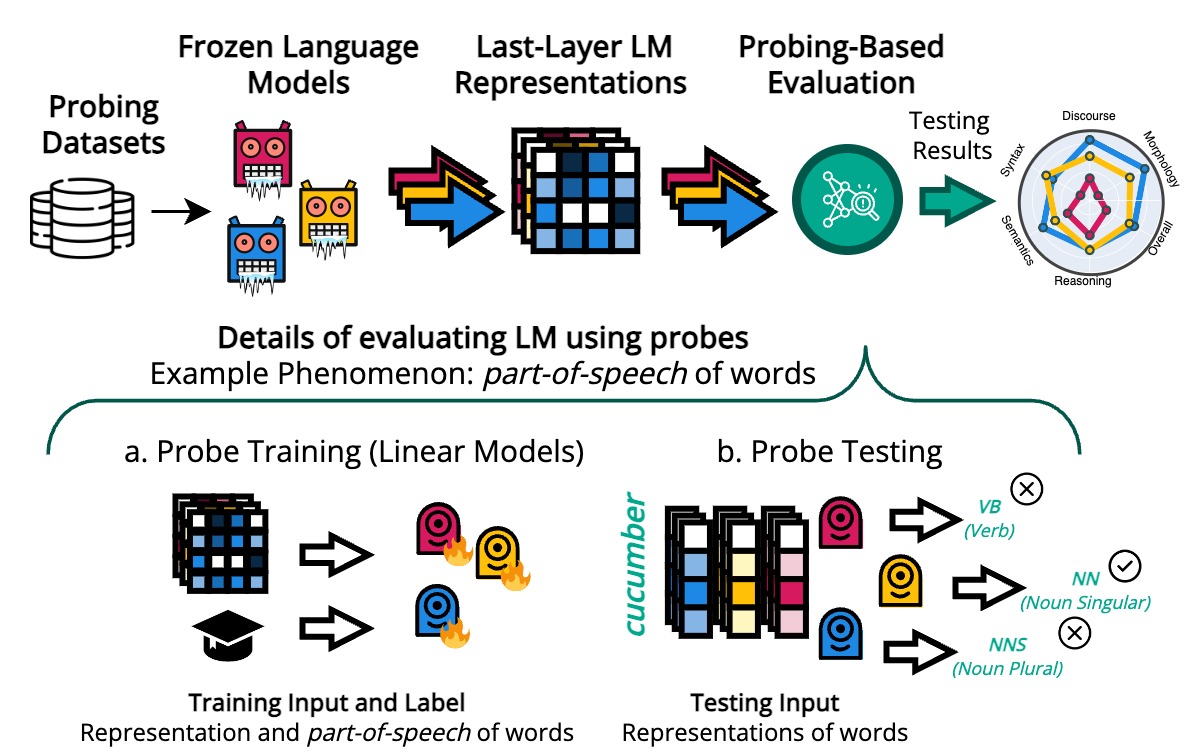}
    \caption{
    In \Holmes{}, we encode examples of probing datasets using frozen LMs. 
    Then, we train probes (linear models) with labels representing the specific linguistic phenomenon under test. 
    Finally, we use the results of testing the probes to approximate the LMs' linguistic competence regarding the tested phenomena. 
    }
    \label{fig:competence-performance}
\end{figure}

\begin{figure*}[t]
    \centering
    \includegraphics[width=0.98\textwidth]{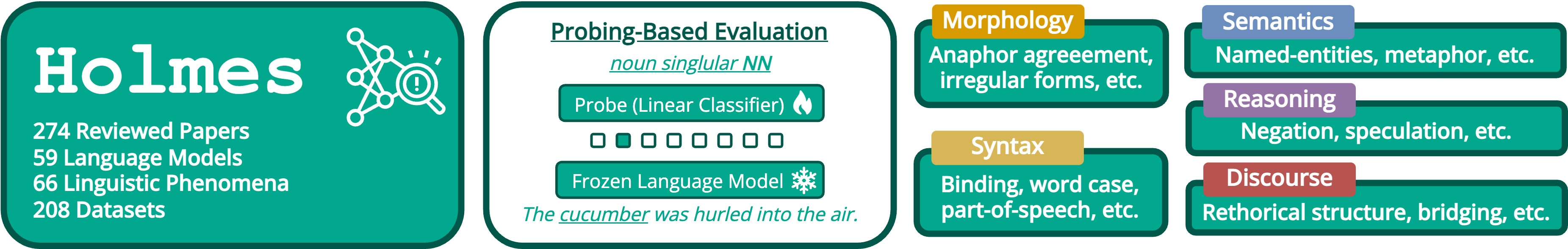}
    \caption{
    Overview of \Holmes{} (left) with the five phenomena types (right) and an example of probing-based evaluations for part-of-speech: encoding the input tokens and predicting the POS tag for \textit{cucumber}, here \textit{NN}.}
    \label{fig:overview_holmes}
\end{figure*}

In this work, we introduce \Holmes{} (\autoref{fig:overview_holmes}),
a benchmark to assess the linguistic competence of LMs (\autoref{fig:top-lms}) regarding numerous linguistic phenomena.
To disentangle LMs understanding of these phenomena from their linguistic performance, we assess the LMs' internals using classifier-based probing \citep{tenney-etal-2019-bert,hewitt-manning-2019-structural,belinkov-2022-probing}.
As illustrated in \autoref{fig:competence-performance} for probing the part-of-speech (POS) tags for words,
we first train linear models (probes) using the internal representations of text inputs from the last model layer to predict the specific phenomena aspects. 
We then approximate the LMs' grasp of these phenomena using the probes' performance, rigorously verified using control tasks \citep{hewitt-liang-2019-designing} and from an information theory perspective \citep{voita-titov-2020-information}.
With this particular and comprehensive scope, we thoroughly address the initially raised questions as follows:

\paragraph{Meta-Study (\autoref{sec:survey})}
The review of over 270 probing studies reveals a gap in comprehensively evaluating linguistic competence.
Despite covering over 200 probing tasks and 150 LMs, individual studies focus on particular tasks and LMs. 
As a result, only three LMs were probed on over 20\% of the tasks, and only one task (POS) was evaluated for more than 20\% of the LMs.
Notably, recent large LMs are significantly underrepresented. 

\paragraph{Benchmark (\autoref{sec:holmes})}
Addressing these identified deficiencies, \Holmes{} offers a structured way to assess LMs' English linguistic competence comprehensively.
It features 208 distinct datasets covering \textit{morphology}, \textit{syntax}, \textit{semantics}, \textit{reasoning}, and \textit{discourse} phenomena, including previously underrepresented ones like negation or rhetoric.

\paragraph{Results and Analysis (\autoref{sec:full})}
From assessing 59 LMs, we find that no LM consistently excels the others.
Further, linguistic competence is more pronounced for \textit{morphology} and \textit{syntax} than the other types of phenomena, and LMs' linguistic competence is fundamentally affected by \textbf{model size}, \textbf{model architecture}, and \textbf{instruction tuning}. 

First, we generalize previous findings \citep{tenney2018what,zhang-etal-2021-need} and show LMs' linguistic competence, particularly \textit{morphology} and \textit{syntax}, scales beyond 350 million parameters.
Second, contrary to the prompting evaluations \citep{lu2023emergent} and aligned with \citet{waldis2024dive} and \citet{Gautam2024RobustPU}, \textbf{model architecture} is critical. 
The linguistic competence of decoder-only LMs lags behind encoder-only ones.
Not even 70 billion decoders produce representations for words with the same stability as encoders with 110 million parameters.
Third, while \textbf{instruction tuning} \citep{ouyang2022training,touvron2023llama,zhou2024lima} aims to align LMs with human interactions, we focus for the first time on its effect on linguistic competence. 
We found that instruction tuning improves \textit{morphology} and \textit{syntax} but has mixed effects on other phenomena types, hinting at a superficial alignment.
Lastly, we compare \Holmes{} with other benchmarks. 
While LM rankings of reasoning-intense downstream tasks \citep{Pham_OpenLLM_Operating_LLMs_2023} correlate with reasoning phenomena, explicitly prompting for linguistic phenomena \citep{liang2023holistic} leads to unreliable results.  
As these results show that \Holmes{} aligns with other benchmarks, its probing-based evaluation is indispensable for explicitly testing LMs' linguistic competence disentangled from their linguistic performance. 




\paragraph{Efficiency (\autoref{sec:efficient})}
Finally, to mitigate the heavy computational burden of evaluating a new LM on \Holmes{}, we form the streamlined version \FlashHolmes{} by selectively excluding samples not significantly influencing overall rankings \citep{perlitz2023efficient}.
Specifically, \FlashHolmes{} approximates \Holmes{} rankings with high precision while requiring only \textasciitilde 3\% of the computation. 

\paragraph{Contributions}
With \Holmes{}, we introduce a comprehensive and thorough benchmark to assess LMs' linguistic competence, providing ground to evaluate them more holistically.
Extensive experiments on \Holmes{} reveal that LMs' linguistic competence is manifold and more pronounced for phenomena targeting words and syntactic structure than semantic, reasoning, or discourse. 
LMs properties like size or architecture crucially account for differences among LMs.
Fostering further research, we provide interactive tools to explore \Holmes{} and straightforward evaluation code for upcoming LMs with efficiency in mind.

\section{Preliminaries}\label{sec:preliminaries}

\paragraph{Language Models (LMs)} 
Language Models compute probabilities for word sequences $i$, enabling tasks such as classifying $i$, textual comparisons between $i$ and another sequence $i'$, and text generation based on $i$.
We consider LMs as any model producing representations of $i$, regardless of their specific type: \textbf{sparse} like bag-of-words \citep{harris1954distributional}; \textbf{static} such as GloVe \citep{pennington-etal-2014-glove}; or \textbf{contextualized} transformers \citep{devlin-etal-2019-bert,raffel2020exploring}.

\paragraph{Linguistic Competence and Performance}
For centuries \citep{robins2013short}, linguists have been fascinated by the processes of language learning, usage, and evolution.
One specific discussion is the differentiation between knowing and using a language.
\citet{saussure1916course} distinguished between language with specific rules and words (\textit{langue}) as an ongoing negotiated fulfillment of the societal need for communication and its usage (\textit{parole}).
Similarly, \citet{chomsky1965} uses the term \textit{linguistic competence} for the unconscious understanding of language and \textit{linguistic performance} for using languages in any utterance.
In this work, we follow Chomsky's terminology and treat LMs as static artifacts of a certain time, omitting ongoing processes of the society considered by de Saussure. 
Specifically, we focus on assessing the linguistic competence of LMs, including specific linguistic phenomena like word dependencies and their distinct parts of speech (POS).
Opposed, contemporary benchmarks \citep{cobbe2021gsm8k,DBLP:conf/emnlp/PetroniRRLBWM19,petroni2020how} assess linguistic performance by providing textual instructions and verifying LMs' textual responses. 
Note that this evaluation protocol can also verify an understanding of specific linguistic phenomena, as done in \citet{blevins-etal-2023-prompting2} or \citet{liang2023holistic} for syntactic structure.
However, such evaluation protocols conflate LMs' linguistic competence with latent abilities (like following instructions.
Thus, \Holmes{} unique evaluation perspective is indispensable to assess linguistic phenomena isolated to assess LMs comprehensively. 

\paragraph{Linguistic Phenomena} We define the linguistic competence of LMs as their ability to understand a diversity of linguistic phenomena.
Specifically, we focus on five phenomena types: \textit{morphology}, the structure of words; \textit{syntax}, the structure of sentences; \textit{semantics}, the meaning of words; \textit{reasoning}, the use of words in logical deduction and other related phenomena like negation or speculation; \textit{discourse}, the context in text like rhetorical structure.
Following \citet{mahowald2023dissociating}, we categorize these phenomena types into two groups:  
\textit{morphology} and \textit{syntax} are \textbf{formal} phenomena, which include understanding grammatical rules and statistical patterns, while \textbf{functional} ones (\textit{semantics}, \textit{reasoning}, and \textit{discourse}) focus on practical abilities like interpreting text sentiment or detecting the existence of speculation.

\paragraph{Datasets} 
We define a dataset as text examples and labels covering a specific aspect of a linguistic phenomenon, like words and their POS tags. 
Typically, these labels are unambiguous, enabling us to assess the specific aspect under test in isolation.

\paragraph{Probes} 
Probing allows to examine what information is encoded in the internal representations of LMs.

We empirically evaluate the linguistic capabilities of language models (LMs) in relation to the linguistic phenomena highlighted in \Holmes{} using probes. To this end, we employ probing tasks based on the well-established classifier-based probing methodology \citep{tenney-etal-2019-bert,hewitt-manning-2019-structural,belinkov-2022-probing}, also referred to as diagnostic classifiers \citep{Veldhoen2016DiagnosticCR,giulianelli-etal-2018-hood}.

Probing tasks involve training a linear model (probe) on a specific dataset to isolate and assess a particular aspect of a linguistic phenomenon. The text examples from the dataset are encoded using a given LM, and these encoded representations are used to train the probe on specific labels that correspond to the linguistic phenomenon being tested.

The probe’s performance serves as an indicator of the LM’s internal representation and understanding of the phenomenon. Higher performance scores suggest that the LM captures relevant patterns internally, which contributes to improved accuracy \citep{tenney2018what}.

Using probes, we empirically assess the linguistic competence of LMs regarding the featured linguistic phenomena in \Holmes{}.
We design probing tasks using the widely recognized classifier-based probing method \citep{tenney-etal-2019-bert,hewitt-manning-2019-structural,belinkov-2022-probing} also known as diagnostic classifiers \citep{Veldhoen2016DiagnosticCR,giulianelli-etal-2018-hood}.
Running such a probing task involves training a probe (linear model) using the specific dataset to test a distinct aspect of a linguistic phenomenon in isolation. 
To do this, we encode the text examples of a dataset with a given LM and use them to train the probe regarding the specific labels representing the tested linguistic phenomenon.
The probe's performance is then used to approximate the LM's understanding of the specific phenomenon.
A higher score indicates that LMs capture patterns relevant to this phenomenon internally, which in turn enhances the accuracy \citep{tenney2018what}.

\section{Meta-Study}\label{sec:survey}
This section summarizes our survey of 274 studies (\autoref{sec:setup}) probing LMs' linguistic competence.
We analyze them regarding their evolution, probing tasks and LMs addressed (\autoref{sec:analysis}), and identify the need to consolidate existing resources (\autoref{subsec:summary}).

\subsection{Scope}\label{sec:setup}
We analyze 28k papers ($P$) from 2015 to August 2023 of major NLP conferences (TACL, ACL, AACL, COLING, EACL, EMNLP, NAACL, and the corresponding workshops) expanded with selected work from other venues such as ICLR.
To identify relevant work, we employ a semi-automatic approach. First, we use automated filtering based on paper metadata and full text\footnote{We use \href{https://web.archive.org/web/20240810051002/https://pypdf2.readthedocs.io/en/3.0.0/}{PyPDF2} v3.0.0, \href{https://web.archive.org/web/20240810051009/https://dblp.org/faq/How+to+use+the+dblp+search+API.html}{DBLP} and \href{https://web.archive.org/web/20240810051101/https://github.com/danielnsilva/semanticscholar}{semanticscholar} API.}, grounded in the occurrence of established terminology related to the specific focus of \Holmes{}, namely disentangling the linguistic competence of LMs by studying their internal representations.
This terminology, including \emph{probing} and \emph{probe}, is commonly found in influential literature surveys \citep{rogers-etal-2020-primer, belinkov-2022-probing} and diverse investigation settings, such as analyzing internal representations using linear classifiers \citep{tenney2018what, conneau-etal-2018-cram,elazar-etal-2021-amnesic} or masked-based approaches focusing on lexical knowledge of LMs \citep{petroni-etal-2019-language, talmor-etal-2020-olmpics, kassner-etal-2021-multilingual, peng-etal-2022-copen}.
Specifically, we define  three criteria to identify relevant papers: $P'=\{\forall p \in P | p \in P_1 \cup p \in P_2\ \cup p \in P_3\}$, where:

$\mathbf{P_1}$: papers with \textit{probing} or \textit{probe} in the title.

$\mathbf{P_2}$: papers with \textit{probing} or \textit{probe} in the abstract and at least five occurrences in the main content.

$\mathbf{P_3}$: papers with \textit{probing} or \textit{probe} occurring at least ten times in the main content.

We identified 493 matching papers ($P'$) by applying these criteria. 
We then manually review the automatically generated candidate list ($P'$) and select studies that examined LMs with one or more specific linguistic phenomena as part of their analysis or as a primary contribution. 
This process involves filtering out papers using the term \textit{probing} in other senses, such as \textit{probing hash tables} in \citet{bogoychev-lopez-2016-n}.
Moreover, we supplement the candidates with a curated selection of highly relevant studies that do not meet the above criteria. 
For example, seminal works published before 2019 which employ terms like ``\emph{diagnostic classifier}'' \citep{giulianelli-etal-2018-hood, ijcai2018p796}, as well as other notable studies \citep{gupta-etal-2015-distributional,shi-etal-2016-string}. 
This comprehensive approach yields 274 relevant papers ($P_r$), which we further analyze subsequently.

\subsection{Analysis}\label{sec:analysis}

\begin{figure}[t]
    \centering
    \includegraphics[width=0.48\textwidth]{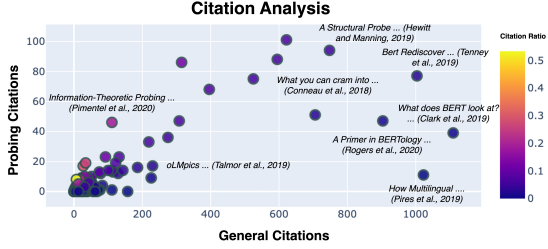}
    \caption{Citation analysis considering \textit{probing citations} originating from the set of relevant work and every other citation (\textit{general citations}). The color scale indicates the ratio ($\alpha$) between them.}
    \label{fig:citations}
\end{figure}

\paragraph{i) Scattered evolution calls for consolidation.}
We begin by examining the evolution of relevant studies in the field, illustrated in \autoref{fig:citations}. 
We analyze the citation patterns among these studies, distinguishing between \textbf{probing citations} ($C_p$), which represent citations between them, and \textbf{general citations} ($C_g$), which encompass all other citations. 
The colorized ratio $\alpha = \frac{|C_p| + 1}{|C_g| + 1}$ visually relates these two measures. 
This analysis reveals that only a small fraction of the works have garnered broad recognition, with 16 papers exceeding 200 general citations. 
Furthermore, probing works cite each other relatively infrequently, with an average probing citation ratio of $\alpha = 0.1$. 
This suggests that other fields have paid limited attention to LMs' linguistic competence. 
The scattered citation patterns and lack of engagement with this topic underscore the need to consolidate existing resources and establish a solid foundation to bootstrap research in this area.

\begin{figure}[t]
    \centering
    \includegraphics[width=0.48\textwidth]{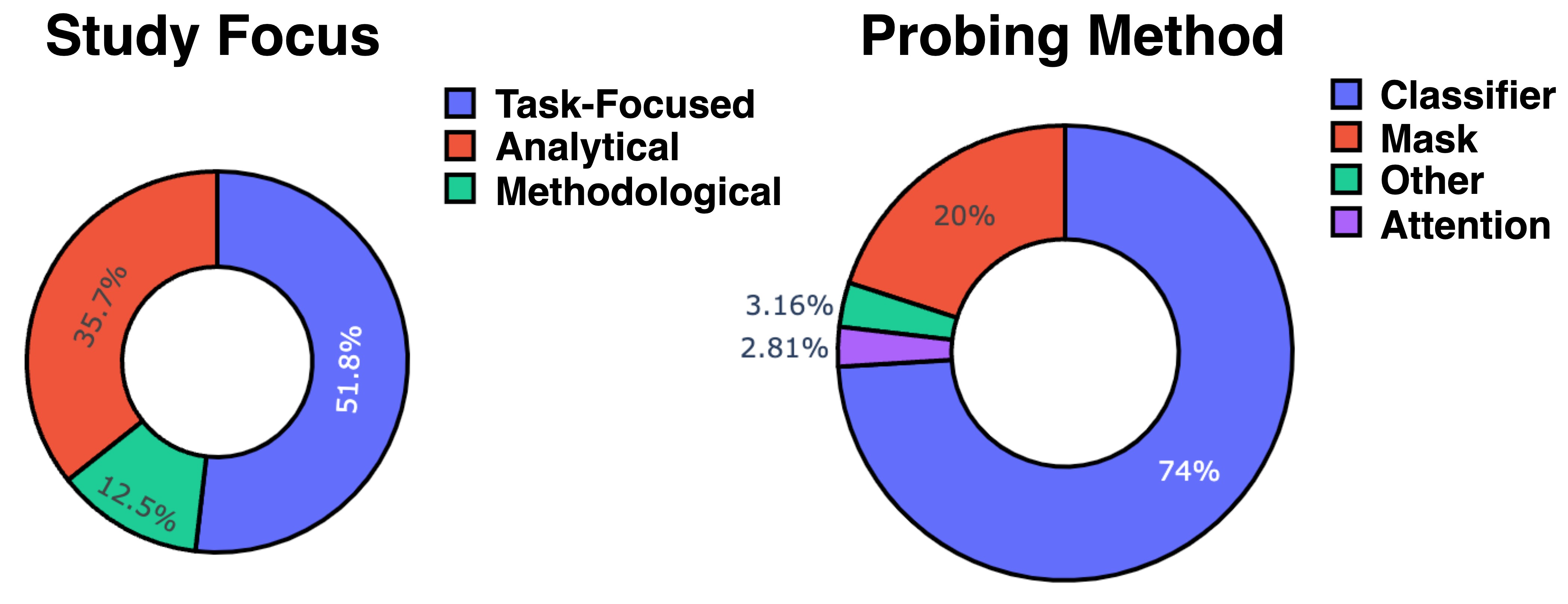}
    \caption{Categorization of the selected studies by their focus and their conducted probing method.}
    \label{fig:focus_approach}
\end{figure}

\paragraph{ii) Probing work prioritizes tasks and analytics over methods.} 
We categorize the selected work according to their probing focus into three categories: \textbf{methodological}, which introduces new methods, such as control tasks \citep{hewitt-liang-2019-designing} or minimum description length \citep{voita-titov-2020-information}; \textbf{task-focused}, which assesses specific linguistic phenomena as main contributions, such as discourse relations in text \citep{koto-etal-2021-discourse}; and \textbf{analytical}, which uses probing tasks to analyze LMs, such as the impact of pre-training data \citep{zhang-etal-2021-need}.
As shown in \autoref{fig:focus_approach}, the majority of studies (51.8\%) focus on specific probing tasks, such as numeric scales \citep{zhang-etal-2020-language}, or morphosyntactic analysis \citep{shapiro-etal-2021-multilabel-approach}. 
A significant proportion (35.7\%) use probing as a supplementary analytical tool, for example, to analyze the effect of fine-tuning \citep{mosbach-etal-2020-interplay,zhu-etal-2022-predicting}. 
The remaining 12.5\% address methodological problems related to probing \citep{wu-etal-2020-perturbed, immer-etal-2022-probing, zhu-etal-2022-data}.

\paragraph{iii) The dominance of classifier-based probing.} 
Next, we analyze the specific employed probing method regarding four categories: (1) \textbf{classifier-based probing}, which uses linear or shallow models to probe internal representations of LMs;  (2) \textbf{mask-based probing}, where LMs fill gaps to verify linguistic phenomena; (3) \textbf{attention-based probing}, which relies on attention patterns; and (4) \textbf{other methods} that do not fit into the previous three categories.
Our analysis indicates that most studies (74\%) utilize the classifier-based probing method, as exemplified in \citet{tenney-etal-2019-bert}.
Additionally, 20\% of studies conduct mask-based probing, as shown in \citet{10.1162/tacl_a_00342}. 
In contrast, only a small portion of work ($\sim$ 3\%) considers attention patterns or other approaches, such as bridging \citep{pandit-hou-2021-probing} or dimension selection \citep{torroba-hennigen-etal-2020-intrinsic}.

\paragraph{iv) Tasks and LMs are barely broadly evaluated.} 
Finally, we examine the tasks and LMs investigated by the relevant studies. 
For example, \citet{tenney2018what} explore BERT on various tasks, including POS tagging, semantic-role labeling (SRL), and others. 
Our analysis reveals that, collectively, these studies cover a remarkable 289 unique tasks and 161 distinct LMs, demonstrating a broad scope of investigation.
Below, we delve into the details and highlight noteworthy findings.

\begin{figure}[]
    \centering
    \includegraphics[width=0.48\textwidth]{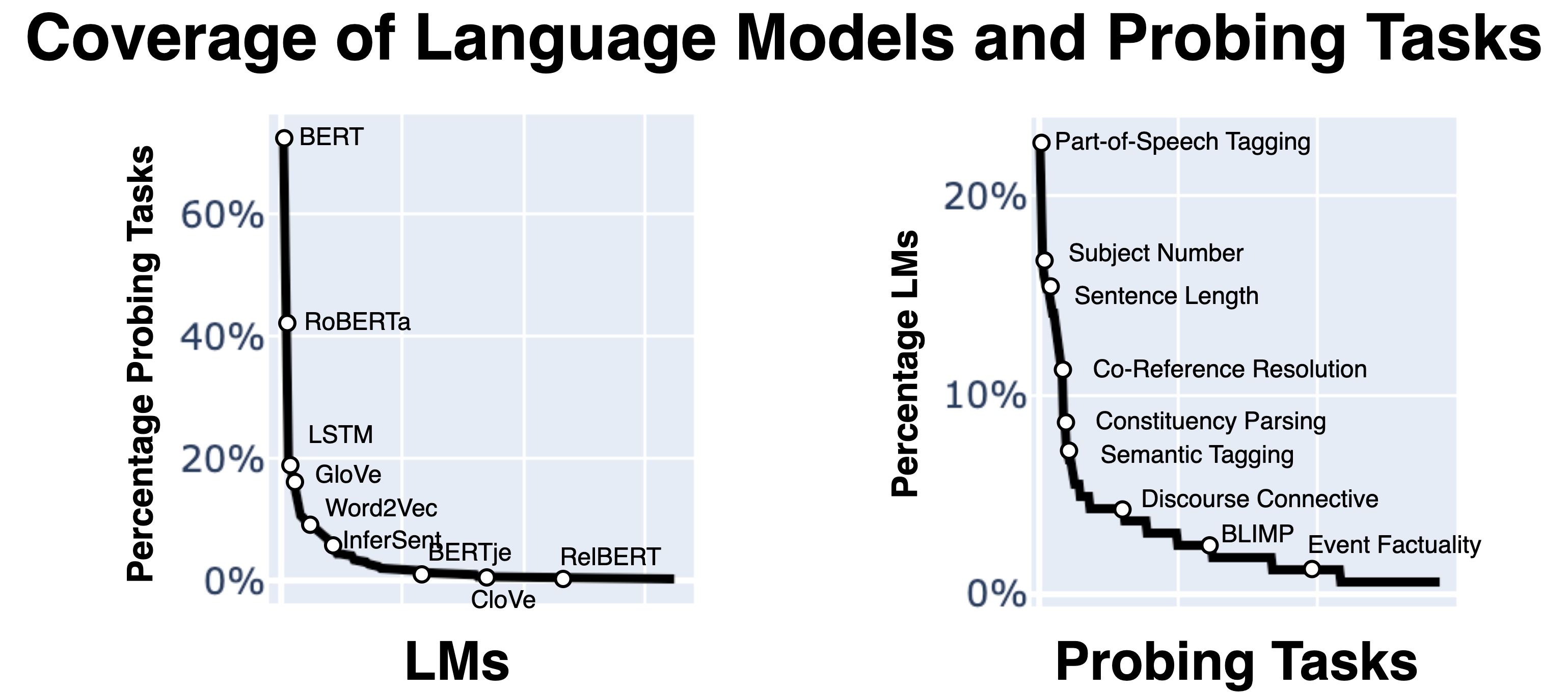}
    \caption{Overview of how many tasks single LMs cover and vice versa - single examples are highlighted.}
    \label{fig:lms_capabilities_coverage}
\end{figure}

We analyze how LMs and tasks are considered jointly in \autoref{fig:lms_capabilities_coverage}.
Despite the broad coverage, single studies, including fundamental ones, maintain a particular focus and consider only a fraction of LMs and tasks. 
For example, while most tasks (72\%) were assessed on BERT, RoBERTa's coverage has already declined to 42\%.
Conversely, part-of-speech tagging (POS), the most probed task, was only evaluated on 23\% of the LMs, excluding prominent examples like BART \cite{lewis-etal-2020-bart}.
In particular, more recently released larger and powerful LMs, like Pythia \citep{biderman2023pythia}, UL2 \citep{tay2022ul2}, or LLAMA-2 \citep{touvron2023llama}, as well as instruction-tuned LMs like FLAN-T5 \citep{chung2022scaling} or LLAMA-2-Chat \citep{touvron2023llama} are missing almost entirely, with only a few recent exceptions \citep{hu-levy-2023-prompting,waldis2024dive}.
Again, these insights underscore the need to consolidate existing resources for more comprehensive coverage.

\begin{figure}[]
    \centering
    \includegraphics[width=0.48\textwidth]{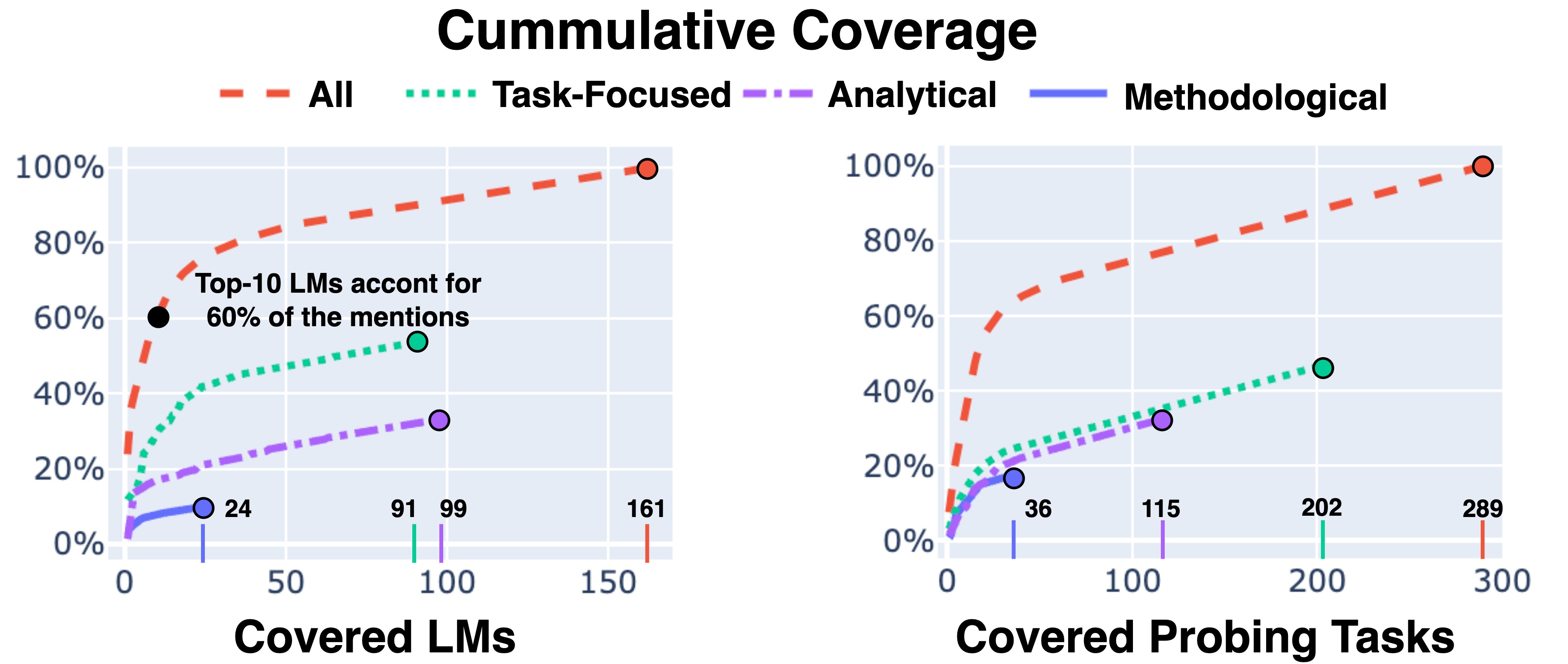}
    \caption{Cumulative coverage of LMs and tasks, considering all relevant studies and their focus.}
    \label{fig:cummulative_coverage}
\end{figure}

\autoref{fig:cummulative_coverage} further highlights this point by sorting LMs and tasks according to their frequency of mention in relevant works and plotting their cumulative coverage.
For example, considering all studies (red line), the top-10 most mentioned LMs account for 80\% of all LMs mentions (black dot), while the remaining 151 unique LMs account for only 40\%.
A comparison of the paper's focus reveals that methodological studies rely only on a limited set of 24 LMs and 36 tasks.
In contrast, task-focused and analytical work cover a similar number of LMs (91 and 99, respectively).
However, due to their distinct focus, task-focused studies cover a significantly larger number of tasks (202) than analytical ones (115).


\begin{table*}[t]
\centering
    \setlength{\tabcolsep}{12pt}
    \resizebox{0.98\textwidth}{!}{%
        \begin{tabular}{lcccccccccc}
        \toprule
             \bf Type & \bf Phenomena & \bf Example & \bf Label\\
        \midrule
      \multirow{2}{*}{\textbf{Morphology}} &\multirow{2}{*}{Subject-Verb Agreement} & \textit{And then, the cucumber \underline{was} hurled into the air.} & \texttt{Correct}\\
     & & \textit{And then, the cucumber \underline{were} hurled into the air.} & \texttt{Wrong}\\
      \textbf{Syntax} & Part-of-Speech & \textit{And then, the \underline{cucumber} was hurled into the air.} & \texttt{NN (Noun Singular)}\\
      \textbf{Semantic} & Semantic Roles & \textit{And then, the cucumber was hurled \underline{into the air}.} & \texttt{Direction}\\
      \textbf{Reasoning} & Negation & \textit{\underline{And then, the cucumber was hurled into the air.}} & \texttt{No Negation}\\
      \textbf{Discourse} & Node Type in Rhetorical Tree & \textit{\underline{And then}, the cucumber was hurled into the air.} & \texttt{Satellite}\\
      
        \bottomrule
        \end{tabular}
    }
    \caption{Example instance of \Holmes{} datasets for every type of linguistic phenomena. The relevant part of the example for the specific label is \underline{underlined}.}
    \label{tab:examples}
\end{table*}

\subsection{Summary}\label{subsec:summary}

Our meta-study emphasizes the need to consolidate existing resources for a comprehensive assessment of the linguistic competence of LMs — a manifold but rather a blind spot in evaluation research.
Apart from more thorough evaluations, such a stimulus can significantly boost future research, as happened in computer vision with ImageNet \citep{Deng2009ImageNetAL} or in NLP with GLUE and SuperGLUE \citep{NEURIPS2019_4496bf24, DBLP:conf/iclr/WangSMHLB19}.

\section{Holmes Benchmark}\label{sec:holmes}

With \Holmes{}, we provide an extensive ground to tackle these identified deficiencies in the existing literature and comprehensively investigate the English linguistic competence of LMs.
Specifically, \Holmes{} features 208 datasets addressing distinct aspects of 66 phenomena covering \textit{morphology}, \textit{syntax}, \textit{semantic}, \textit{reasoning}, and \textit{discourse}.

\subsection{Datasets}\label{subsec:structure}

We provide a comprehensive coverage of linguistic phenomena by covering 208 unique datasets. 
We leverage existing and established resources like OntoNotes \citep{weischedel2013ontonotes}, English Web Treebank \citep{silveira-etal-2014-gold}, or BLiMP \citep{warstadt-etal-2020-blimp-benchmark} to create datasets addressing phenomena like the POS of words, their dependencies or the linguistic acceptability of sentences.
Further, we include a range of less employed data, addressing contextualization of words \citep{klafka-ettinger-2020-spying}, reasoning \citep{10.1162/tacl_a_00342}, semantic decomposition \citep{white-etal-2016-universal,rudinger-etal-2018-neural,rudinger-etal-2018-neural-models,govindarajan-etal-2019-decomposing, vashishtha-etal-2019-fine}, grammatical knowledge \citep{huebner-etal-2021-babyberta}, bridging \citep{pandit-hou-2021-probing}, and rhetorical \citep{carlson-etal-2001} and discourse \citep{webber2019penn} structure in text.
Finally, we cover rarely probed phenomena like negation \citep{vincze2008bioscope, konstantinova-etal-2012-review, vahtola-etal-2022-easy}, or word complexity \citep{paetzold-specia-2016-semeval}.

\subsection{Structure}\label{subsec:structure}
Apart from the comprehensive scope, \Holmes{} provides a clear structure for specific evaluations on different levels of aggregation. 
We first group the datasets according to the linguistic phenomena addressed.
Next we categorize these phenomena into their previously defined five phenomena types (see \autoref{sec:preliminaries}): \textit{morphology}, like the agreement of subject and verb; \textit{syntax}, such as the part-of-speech of words; \textit{semantics}, like semantic roles of words; \textit{reasoning}, such as detecting a negated sentence; and \textit{discourse}, like selecting the correct following sentence.
\autoref{tab:examples} provides examples for every type of phenomenon.
Note that we rely on the categorization provided by the specific studies whenever given - more details in the Appendix \autoref{app:task-categorization}.
For example, \citet{conneau-etal-2018-cram} categorized the tense of the main clause as \textit{semantic}.
This phenomenon could also be categorized as \textit{syntax} if we test the detection of incorrect formulations given a specific tense. 
However, we follow the authors' suggestion and test the detection of the tense on a sentence level, which represents semantic aspects.

\subsection{Experimental Setup}
\Holmes{} evaluation follows the primarily used classifier-based probing paradigm, as described in \autoref{sec:preliminaries}, to analyze the internal representations of the last layer of LMs.\footnote{Please refer to Appendix \autoref{app:details-probing-datasets} and \autoref{fig:app:tasks} for more details about the composition of the internal representations.}
Thereby, we maximally disentangle the understanding of distinct linguistic phenomena from each other and from other cognitive abilities, such as following textual instructions. 
Further, this method allows us to assess any LM tpye, including sparse, static, or contextualized ones. 
Based on the specific dataset, we either select the embeddings of the specific input tokens (like single words for POS tagging) or average embeddings across a span or the whole sentence. 
We define a probing task as training a probe $f_p$ (linear model without intermediate layers) using these embeddings as inputs and the dataset labels as training signals.
If not defined in the original data, we divide the dataset samples into train/dev/test split following a ratio of 70/10/20.
We repeat this procedure five times using different random seeds for a robust measurement.

\subsection{Evaluations}\label{subsec:evaluation}
We approximate how well an LM encodes specific linguistic phenomena using the absolute prediction performance of the probes.
In addition, we rigorously evaluate the reliability of probing results using control tasks and from an information theory perspective \citep{voita-titov-2020-information, hewitt-liang-2019-designing}.
Different from commonly used prompting assessments, this particular evaluation protocol refrains from known fallacies in which the results and conclusions are sensible with specific instructions \citep{mizrahi2023state, min-etal-2022-rethinking} or few-shot examples \citep{lu2023emergent}.

\paragraph{Task Score Metric}
Based on a dataset's specific task type, we use a corresponding performance measure, macro \textbf{$F_1$} for classification or Pearson correlation for regression.
In addition, we calculate the standard deviation $\sigma$ of the probe across multiple seeds.
A lower $\sigma$ indicates a better encoding of a given linguistic phenomenon since the measurement is robust to noise. 
Further, we use the task score for ranking-based evaluation of all evaluated LMs $L=\{l_1, ...,l_m\}$ within \Holmes{}.
We calculate the mean winning rate $mwr$ (in percentage), telling us how many times one LM $l_1$ wins against others \citep{liang2023holistic}.
With a higher $mwr$, we assume an LM encodes tested linguistic phenomena better than others.

\paragraph{Compression}
Next, we evaluate the probes' reliability from an information-theoretic perspective.
Following \citet{voita-titov-2020-information}, we use the compression $I$.
It is the ratio between the minimum description length $mdl$ of encoding $n$ instances with a label space of $K$ compared to applying a uniform encoding $I=\dfrac{u}{mdl}$.
A higher $I$ means fewer bits are needed to encode the instance and their labels, indicating that the given linguistic phenomenon is more clearly encoded in the internal representation of LMs.

\paragraph{Selectivity}
A reliable probe should grasp patterns relevant to the tested phenomena in the internal representations of LMs but should not be able to learn anything else.
Therefore, we expect high performance when evaluating the specific dataset but low performance when we randomize training signals. 
We check this using control tasks introduced in \citet{hewitt-liang-2019-designing}.
Specifically, we calculate the selectivity $S=F_1(y,\hat{y}) - F_1(y',\hat{y'})$ as the difference between the probe trained with the original labels $y$ and the control task where we train the probe with randomly assigned labels $y'$.
With a higher $S$, we assume the detected patterns are relevant for the specific phenomena under test, as random patterns do not lead to similar performance. 

\begin{figure}[]
    \centering
    \includegraphics[width=0.5\textwidth]{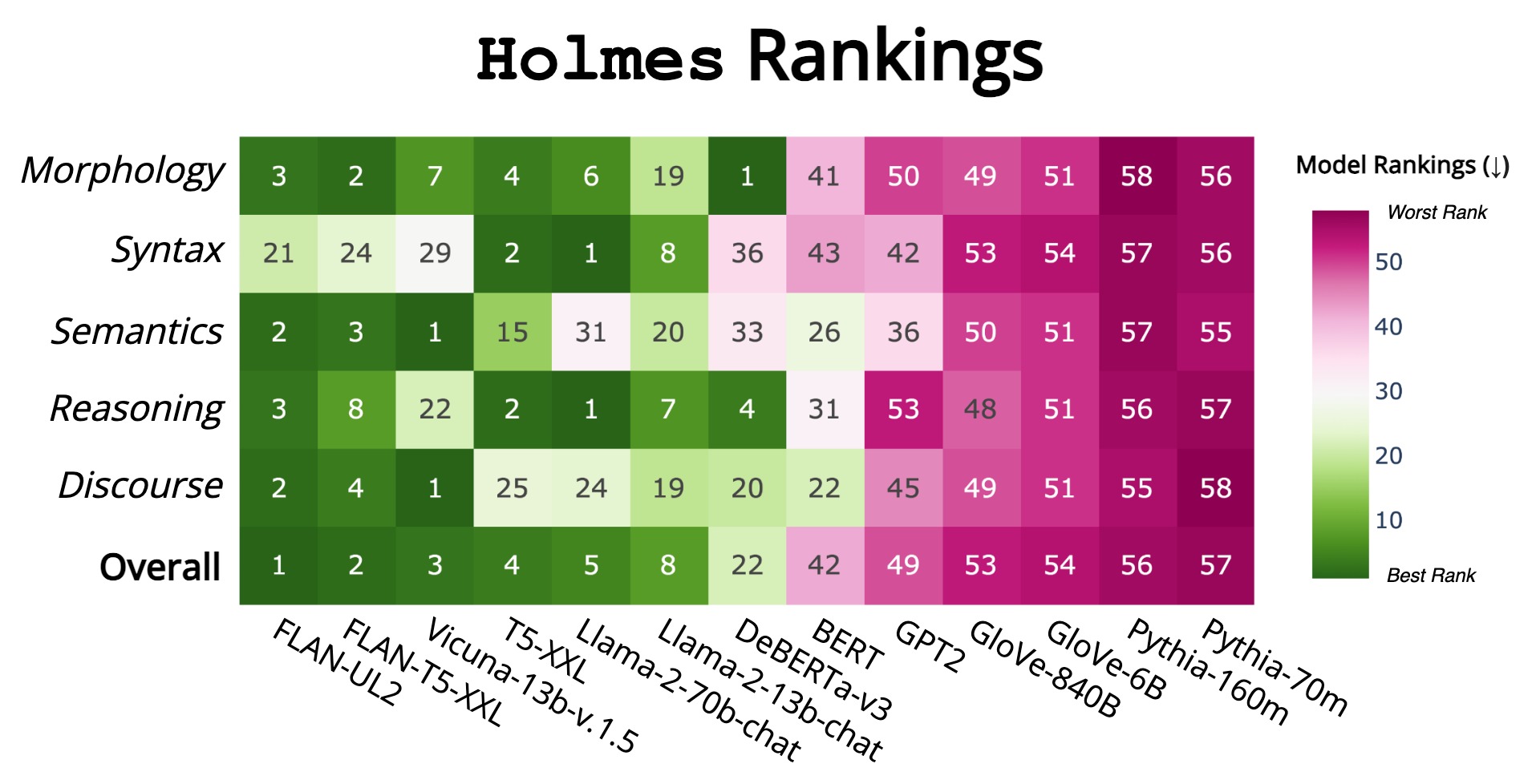}
    \caption{
    A subset of \Holmes{} rankings ($\downarrow$) for various evaluated LMs.
    FLAN-UL2 outperforms the others \textit{overall}, while different LMs prevail for the five distinct types of linguistic phenomena.
    }
    \label{fig:top-lms}
\end{figure}

\section{\Holmes~Results}\label{sec:full}
Using \Holmes{}, we evaluate a diverse collection of 59 LMs.\footnote{Please refer to  Appendix \autoref{app:lms} for a complete list.}
Using the results of these extensive experiments, we first answer the research question: \textit{what is the linguistic competence of LMs?}
In doing so, we discuss the reliability of results (\underline{\hyperlink{paragraph:one}{i}}) and the linguistic competence of LMs concerning the unique structure of \Holmes{} (\underline{\hyperlink{paragraph:two}{ii}}).
Subsequently, we examine \textit{how linguistic competence varies among LMs}, as we find LMs prevailing for different types of linguistic phenomena (\autoref{fig:top-lms}) and delve into the effects of model architecture (\underline{\hyperlink{paragraph:three}{iii}}), size (\underline{\hyperlink{paragraph:four}{iv}}), and instruction tuning (\underline{\hyperlink{paragraph:five}{v}}).
Finally, we show how \Holmes{}' results relate to the \textit{linguistic performance } of LMs by comparing them with the OpenLLM benchmark (\underline{\hyperlink{paragraph:six}{vi}}) and further experiments with the HELM benchmark (\underline{\hyperlink{paragraph:seven}{vii}}).

\begin{figure}[]
    \centering
    \includegraphics[width=0.5\textwidth]{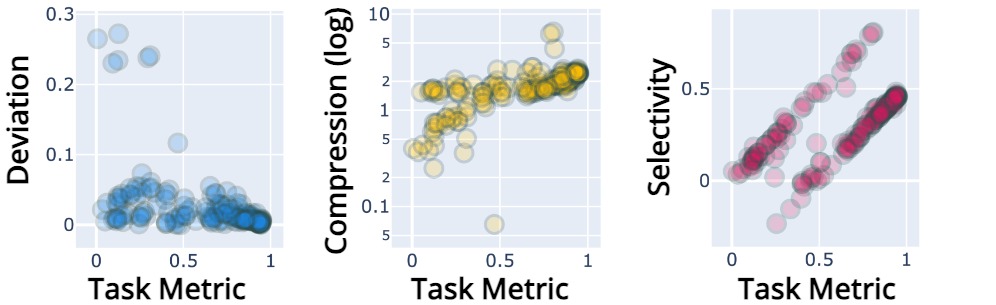}
    \caption{Reliability evaluation \Holmes{} results to ensure low \textit{deviation} across random seeds, high information \textit{compression} (log), and high \textit{selectivity}.
    Every dot represents the averaged results of one probing dataset across LMs. The x-axis represents the task metrics (either person correlation or macro $F_1$).}
    \label{fig:dataset_metrics}
\end{figure}

\paragraph{i) \Holmes{} results are reliable.}\label{paragraph:one}
\autoref{fig:dataset_metrics} shows the reliability of probing-based evaluations using averaged results across random seeds and LMs. 
Single outliers are datasets that are too hard for all LMs, either because the sample size is too small or the linguistic phenomena under test are too complex.
First, a low average \textit{deviation} ($\sigma=0.02$) across five seeds underscores the reliability of probing-based measures. 
These results also highlight the stability of probing results over prompting-based evaluations, where prompt paraphrasing leads to deviations of $\sigma=0.07$ reported in \citet{mizrahi2023state}.
Next, substantial \textit{compression} (average $I=1.9$) and \textit{selectivity} (average $S=0.31$) further confirm the probes' reliability.
Note, for \textit{selectivity}, we consider only base-sized model (10m-200m parameters) for computational efficiency.
Interestingly, two parallel trends emerge. 
More challenging datasets with many labels, like POS tagging, are arranged around a selectivity of 0.1 to 0.4 and a task metric of 0.3.
In contrast, for easier binary classification tasks (such as linguistic applicability), we observe selectivity around 0.2 to 0.5 and a task metric of 0.6 to 0.9.
Furthermore, our analysis reveals a statistically significant positive correlation ($p<0.05$) between the task metrics and both compression ($\tau=0.64$) and selectivity ( $\tau=0.65$). This finding provides strong evidence for the reliability of our task metric, thereby justifying its use as the primary evaluation measure in our study.

\begin{figure}[]
    \centering
    \includegraphics[width=0.5\textwidth]{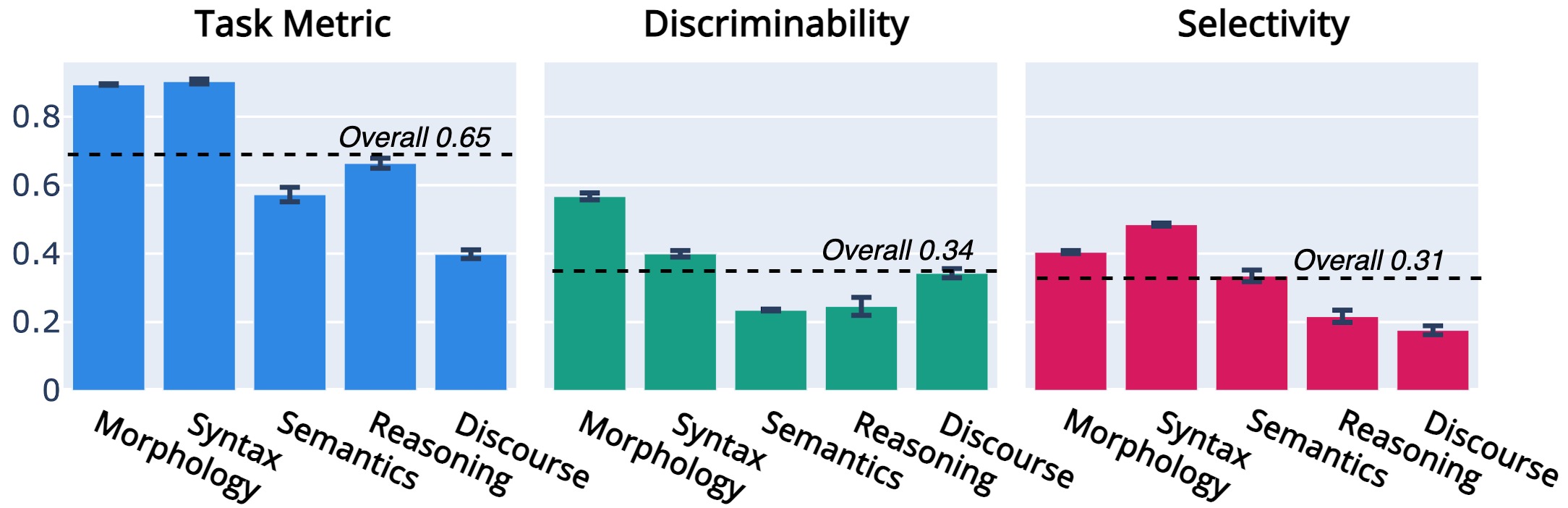}
    \caption{Average \textit{task metric}, \textit{difficulty}, and \textit{discriminability} for each phenomena type. The dashed lines show the average measure over all datasets.}
    \label{fig:competence_metrics}
\end{figure}

\paragraph{ii) LMs' linguistic competence is manifold.}\label{paragraph:two}
We focus on what \Holmes{} tells us in general and regarding formal and functional phenomena, as defined in \autoref{sec:preliminaries}.
We report in \autoref{fig:competence_metrics} the \textit{task metric}, \textit{discriminability}, and \textit{selectivity}, averaged for every phenomena type. 
Note, discriminability \citep{rodriguez-etal-2021-evaluation} quantifies the alignment of LMs ranking of one specific dataset compared to the overall rankings using the Kendall Tau correlation.
Considering these three metrics, all tested LMs strongly encode formal phenomena (\textit{morphology} and \textit{syntax}), which often depend on the local neighborhood of words.
Therefore, we assume that LMs approximate these co-occurrences during pre-training with high precision. 
For example, the specific POS tag of a word, like \textit{man} (\textit{noun}), primarily depends on its surroundings, such as the frequent predecessor \textit{the}.
In contrast, LMs encode less information about functional phenomena (\textit{semantics}, \textit{reasoning}, and \textit{discourse}) since they show a relatively low performance regarding the task metric. 
For these functional phenomena, we assume more complex co-occurrences are required to capture the broad context in language, such as the rhetorical relation of two distant text spans. 
Despite these differences between formal and functional phenomena types, they contribute to the benchmark in a balanced way.
A low to medium discriminability indicates that none of these linguistic phenomenon types dominates the overall LM rankings.

\begin{figure}[]
    \centering
    \includegraphics[width=0.5\textwidth]{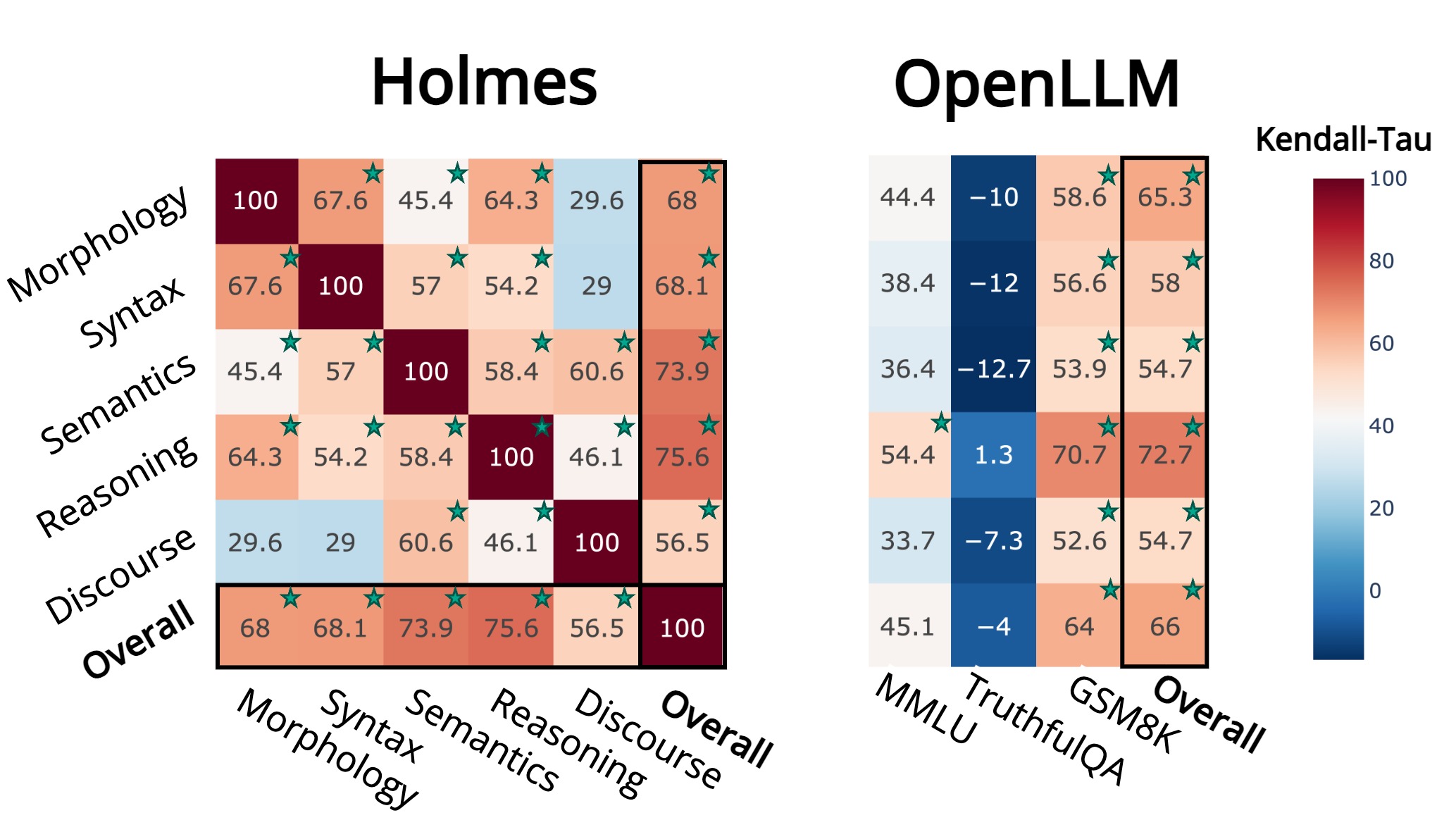}
    \caption{Kendall-tau correlation within \Holmes{} (\underline{left}) and compared to OpenLLM (\underline{right}). Green stars indicate significant correlations ($p<0.05$).} 
    \label{fig:benchmark-heatmap}
\end{figure}

This balanced influence of the five phenomena types is further visible when considering their ranking correlations (\autoref{fig:benchmark-heatmap}, left).
A high average correlation of $68.4\pm7.5$ with the overall results (last column/row) hints that they are facets of a broader occurrence but share common characteristics.
Still, breaking into categories is meaningful, as the phenomena types (first five columns/rows) are medium correlated (average of $54.7\pm13.9$).
Analyzing the results of phenomena types further highlights the value of this distinction.
While results of \textit{semantics} and \textit{reasoning} are similarly correlated with the overall results ($73.9$ and $75.6$), their direct correlation ($58.4$) indicates their supplementary nature.
Further, \textit{discourse} results show the lowest correlation with others ($44.4\pm14.7$), indicating a particular scope.

\begin{figure}[]
    \centering
    \includegraphics[width=0.5\textwidth]{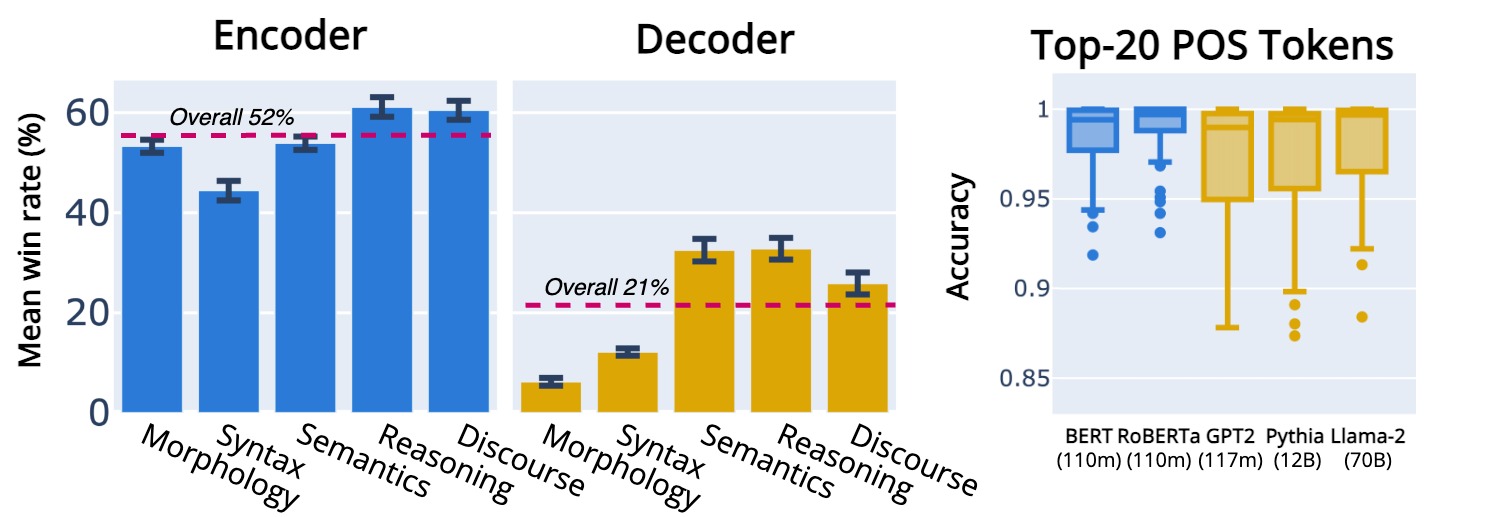}
    \caption{Comparison of the phenomenon types for encoder and decoder LMs (\underline{left}) and on the \underline{right}, the accuracy of the top-20 most common tokens of the three part-of-speech probing datasets for BERT, RoBERTa, GPT2, Pythia, and Llama-2.}
    \label{fig:architecture}
\end{figure}

\paragraph{iii) Encoder architecture equips LMs with high linguistic competence.}\label{paragraph:three}
Next, we discuss the impact of model architecture on the linguistic competence of LMs.
In \autoref{fig:architecture} (left), we compare encoder and decoder LMs.
Due to the absence of big encoder LMs, we consider five \textit{encoder} and six \textit{decoder} LMs with up to 220m parameters.
Encoder LMs show a higher $mwr$ of $52\%$ than decoder LMs ($21\%$).
This observation is the most saturated for \textit{morphology} or \textit{syntax}, encompassing a variety of token-level phenomena, like part-of-speech.
We assume that the missing bi-directional encoding of decoder LMs causes this lower performance because the available context of one token heavily depends on its position. 
Thus, even common tokens, like \textit{the}, have different potential representations - at the beginning or middle of a sentence. 
These instabilities are further evident when considering \autoref{fig:architecture} (right) which reports the accuracy for the top-20 most common POS tokens (such as \textit{the}) based on the \textit{pos}, \textit{xpos}, \textit{upos} dataset.
Given their high frequency, one expects stable prediction performance.
Surprisingly, encoder LMs (BERT and RoBERTa) show higher median accuracy and lower deviations compared to the same-size decoder counterpart (GPT2).
While scaling model size to 12B (Pythia) and 70B (Llama-2) allows for improved accuracy and lower deviations, decoder LMs do not match the encoder performance, even up to \textbf{700 times bigger}.

\begin{figure*}[]
    \centering
    \includegraphics[width=1.0\textwidth]{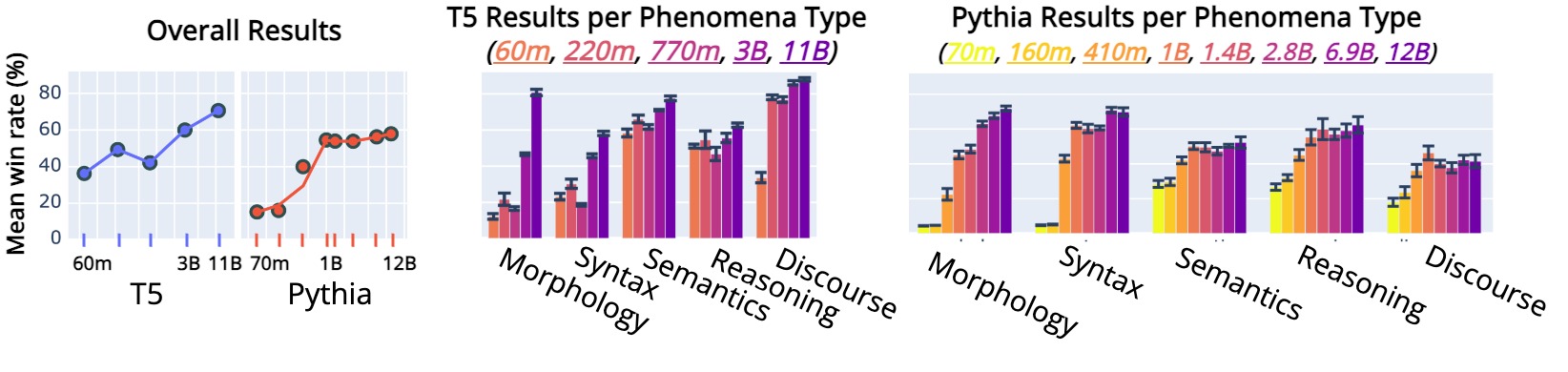}
    \caption{Effect of scaling LM parameters considering the T5 and Pythia model families providing eight and five different sizes. We address the overall scope (\underline{left}) and the different types of linguistic phenomena (\underline{right}).}
    \label{fig:scaling}
\end{figure*}

\paragraph{iv) More parameters improve LMs' linguistic competence.}\label{paragraph:four}
We discuss how the number of parameters influences the linguistic competence of LMs. 
Given the variety of LMs of different sizes, we focus on the Pythia (decoder-only) and T5 (encoder-decoder) families.
From \autoref{fig:scaling}, we observe for both Pythia and T5 that the linguistic competence scales with model size, and it is particularly pronounced after exceeding 0.5B (Pythia) and 1.0B (T5) parameters.
Again, model architecture is crucial, as T5 LMs (encoder-decoder) exhibit a clearly higher mean winning rate of $40-70\%$ than Pythia (decoder-only) ones with $mwr$ of $20-60\%$.
Further, we found formal phenomena evolving differently with increased model size than functional ones.
Specifically, \textit{morphology} and \textit{syntax} start at a lower level, with an apparent performance jump after 0.5B (Pythia) and 1.0B (T5) parameters, followed by slow but steady growth. 
Differently, \textit{semantics}, \textit{reasoning}, and \textit{discourse} start at a higher $mwr$, followed by a continuous improvement as the model size grows.
From these results, we assume that \textbf{more parameters enable language models to better approximate simple word co-occurrences} in nearby contexts. 
While handling formal phenomena like word dependencies, they struggle with more distant and complex co-occurrences, such as rhetorical relations.

\begin{table}[ht!]
\centering
  \resizebox{0.48\textwidth}{!}{%

    \begin{tabular}{lcccccc}
 \bf Model& \bf Morphology   & \bf Syntax  & \bf Semantics   & \bf Reasoning   & \bf Discourse   & \bf Overall  \\
\toprule
  \rowcolor{lightgray}
\multicolumn{7}{c}{\textit{Comparison against \underline{Llama-2}} with 7 billion parameters}\\
Llama-2-Chat & -8\% & +5\% & -6\% & -8\% & -2\% & -2\% \\\midrule
  \rowcolor{lightgray}
\multicolumn{7}{c}{\textit{Comparison against \underline{T5}} with 11 billion parameters}\\
FLAN-T5 & +9\% & +1\% & -3\% & +6\% & 0\% & +1\%  \\\midrule
  \rowcolor{lightgray}
\multicolumn{7}{c}{\textit{Comparison against \underline{Pythia}} with 12 billion parameters}\\
Dolly-v2 & +4\% & -1\%& -9\% & -2\% & +2\% & -3\% \\\midrule
  \rowcolor{lightgray}
\multicolumn{7}{c}{\textit{Comparison against \underline{Llama-2}} with 13 billion parameters}\\
Tülu-2 & +6\% & +3\% & -13\% & +1\% & -13\% & -4\% \\
Orca-2 & 0\% & -4\% & -6\% & +3\% & -2\% & -3\%  \\
Llama-2-chat & +9\% & +6\%  & 0\% & +7\% & +1\% & +4\%\\
Vicuna-v1.5 & +26\% & +9\%& 0\% & +8\% & +2\%& +7\% \\\midrule
  \rowcolor{lightgray}
\multicolumn{7}{c}{\textit{Comparison against \underline{UL2}} with 20 billion parameters}\\
FLAN-UL2 & \textbf{+41\%} & \textbf{+15\%} & \textbf{+6\%} & \textbf{+11\%} & -1\% & \textbf{+12\%} \\
  \rowcolor{lightgray}
\multicolumn{7}{c}{\textit{Comparison against \underline{Mixtral}} with \textasciitilde 47 billion parameters}\\
Mixtral-Instruct & +6\% & +4\% & +1\% & +9\% & +3\% & +4\%   \\\midrule
  \rowcolor{lightgray}
\multicolumn{7}{c}{\textit{Comparison against \underline{Llama-2}} with 70 billion parameters}\\
Tülu-2 & +14\% & 0\% & -9\% & -4\% & +1\% & -2\% \\
Llama-2-Chat & +24\% & +13\% & +3\% & +3\% & \textbf{+13\%} & +10\% \\\midrule\midrule
\textit{\textbf{Average}} & \textit{+10\%} & \textit{+5\%} &\textit{ -3\%} & \textit{+4\%} & \textit{-1\%} & \textit{+2\% }
\\
\bottomrule
    \end{tabular}
  }
  \caption{The mixed effect of instruction tuning on the mean winning rate compared to the pre-trained LMs.}
  \label{tab:instruction}
\end{table}

\paragraph{v) Instruction-tuned LMs get better at mimicking language than understanding it.}\label{paragraph:five}
We focus on how instruction tuning affects LMs' linguistic competence and compare tuned and pre-trained LMs, for example, FLAN-UL2 vs. UL2. 
\autoref{tab:instruction} shows less saturated effects for the overall scope while being more pronounced for the five phenomenon types - again emphasizing the structured and comprehensive evaluation of linguistic competence. 
On average, we found instruction tuning has the highest effect on \textit{morphology} ($+10\%$) followed by \textit{syntax} ($+5\%$), \textit{reasoning} ($+4\%$), and a negative effect for \textit{semantics} $-3\%$ and \textit{discourse} $-1\%$.
These results confirm previous assumptions that \textbf{instruction tuning updates are often superficial} \citep{Yadav2023ComPEFTCF,hershcovitch2024lossless,sharma2023truth} and that LMs get better at mimicking language (formal phenomena) than understanding it, measured with functional phenomena \citep{mahowald2023dissociating}. 
Further, larger models benefit more from instruction tuning. 
Llama-2-70b-Chat and FLAN-UL2 gain up to $+24\%$ and $+41\%$ for \textit{morphology} and $+10\%$ and $+12\%$ on average.
When comparing LMs based on Llama-2-13B, we see that specific fine-tuning methods shape the LMs differently. 
The top-ranked 13B LM for \Holmes{} and OpenLLM, Vicuna, was trained on fewer instructions than others (125k) but of higher quality. 
This high quality seems important as LMs with more instructions but lower quality (Tülu with approx. 330k instructions) lose performance, the same for 70B versions. 
Further and aligned with the previous comparison with OpenLLM results, reasoning specialization (Orca-2) is reflected in the corresponding phenomena.
These insights show again that while providing a particular perspective, \Holmes{} shows apparent differences between LMs and allows us to map them to methodological decisions.

\paragraph{vi) Internals of LMs are partly aligned with their linguistic performance.}\label{paragraph:six}
We analyze the alignment of the probing-based LM rankings of \Holmes{} with prompting-based ones when evaluating downstream using the LMs responses (linguistic performance).
Specifically, we compare against OpenLLM \citep{Pham_OpenLLM_Operating_LLMs_2023}.\footnote{Unlike other benchmarks like HELM \citep{liang2023holistic}, OpenLLM covers many open LMs leading to high overlap with \Holmes{}.}
\autoref{fig:benchmark-heatmap} (right) shows \Holmes{} and OpenLLM rankings of jointly evaluated LMs are medium correlated, hinting that LMs' linguistic competence is partly reflected in their language utterances when solving concrete tasks.
While \textit{syntax}, \textit{semantics}, and \textit{discourse} show similar correlation (54.7 to 58.0), \textit{morphology} and \textit{reasoning} exhibit a substantially higher one of 65.3 and 77.5. 
These results suggest that \textbf{LMs' reasoning abilities are reflected in their internal representations} when evaluating related phenomena like identifying the cause of negations.
These correlation patterns are consistent across the three most meaningful OpenLLM datasets (\textit{MMLU}, \textit{TruthfulQA}, and \textit{GSM8K}).
As \textit{TruthfulQA} shows lower correlations with the linguistic phenomena and other datasets within OpenLLM, we presume this dataset captures distinctly different skills (possibly knowledge).

\paragraph{vii) Prompting is not a substitute for probing when evaluating LMs' linguistic competence.}\label{paragraph:seven}
Finally, we compare probing- and prompting-based LM rankings on the jointly evaluated BLiMP tasks \citep{warstadt-etal-2020-blimp-benchmark} of \Holmes{} and HELM \citep{liang2023holistic}.
Results (Appendix, \autoref{fig:app:helm-holmes}) show apparent discrepancies (rank correlation $\tau=0.05$) when evaluating LMs' internal representations or their responses (linguistic performance) to HELM instructions.
As most prompting-based results from HELM fall below the random baseline, only \textbf{probing-based evaluation can effectively isolate the assessment of linguistic phenomena}.
In contrast, prompting-based methods mix this assessment with other abilities, such as instruction following.
Similar to \citet{hu-levy-2023-prompting}, these insights show the need for a more comprehensive comparison of different evaluation protocols like probing, prompting, or log-probabilities (used in HELM in Figure 33 on page 58 as a workaround for BLiMP). 
Nevertheless, probing provides a unified evaluation protocol assessing the diversity of linguistic phenomena using representations of tokens, spans, or whole texts beyond minimal pair tasks testing whether correct or wrong sentences are preferred.

\section{Efficiency}
\label{sec:efficient}
Seamless, easy, cost-effective integration of new LMs is crucial to adopting benchmarks widely.
As \Holmes{} covers many datasets and examples, it is computationally heavy in encoding text and training the probes.
It takes $\thicksim$ 6 GPU days to encode the 70 million tokens ($\thicksim$230k pages) and two days to run the 208 probes for a 70b model. 
To account for this issue, we introduce \FlashHolmes{}, a streamlined version of \Holmes{}, to evaluate new LMs with a fraction of the compute while maintaining evaluation integrity.

\begin{figure}[]
    \centering
    \includegraphics[width=0.5\textwidth]{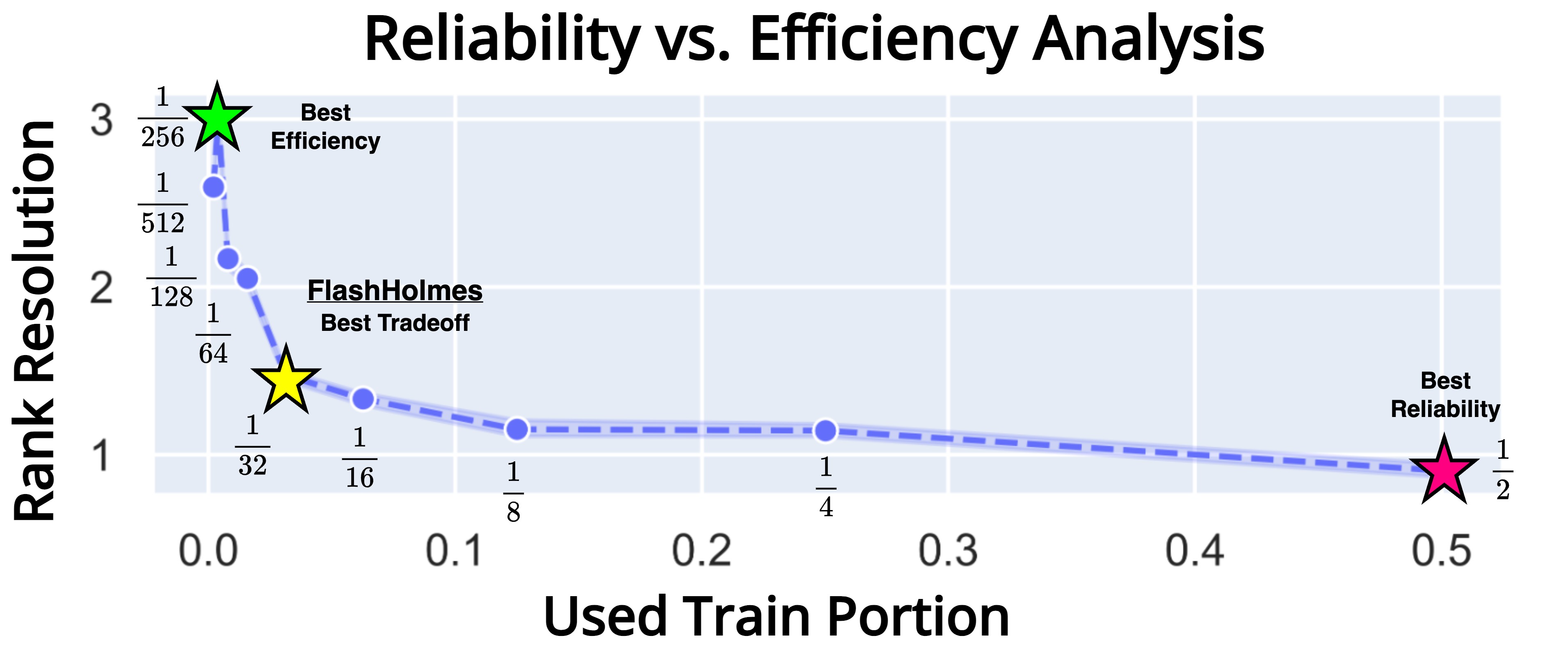}
    \caption{Analysis of the reliability vs. efficiency trade-off when reducing the number of training data.}
    \label{fig:flash_holmes}
\end{figure}

Besides excluding licensed data (18 probing datasets), we analyze the effect of discarding training instances. 
As a result, we reduce the computation for encoding and the actual probing simultaneously. 
We follow \citet{perlitz2023efficient} and calculate the \textit{rank resolution}, $95\%$ CI of model rank difference.
This measure indicates the maximum expected rank deviation from evaluating an LM on \FlashHolmes{} compared to \Holmes{}.
For example, a rank resolution of one means that an LM evaluated on \FlashHolmes{} and \Holmes{} has the same rank or switch place with its neighbors with a probability of 95\%.
\autoref{fig:flash_holmes} shows the resulting rank resolution when training only on a fraction of the instances, from $1/2$ to $1/512$.
Solely focusing on efficiency ($1/512$) still provides a decent rank resolution of \textasciitilde 2.6.
In contrast, considering $1/2$ of the training data results in the best reliability of \textasciitilde 0.9.
To balance benchmark reliability and efficiency, we compose \FlashHolmes{} using $1/32$ of the training instances.
Precisely, it reduces the computation expenses of evaluating LMs to \textasciitilde 3\% of what \Holmes{} would have required while preserving a high rank-correlation of \textasciitilde 1.5.

\section{Related Work}
\label{sec:related_work}

\paragraph{Benchmarking LMs}

Benchmarks approximate LMs abilities like general language understanding \citep{NEURIPS2019_4496bf24,DBLP:conf/iclr/WangSMHLB19}, out-of-distribution generalization \citep{yang-etal-2023-glue,waldis2024handle}, real-world knowledge contradiction \citep{hou2024wikicontradictbenchmarkevaluatingllms}, adversarial scenarios \citep{nie-etal-2020-adversarial,wang2021adversarial}, or retrieval \citep{thakurbeir, muennighoff-etal-2023-mteb}.
With the recent advent of large LMs, the predominant method has shifted to evaluate the obtained linguistic performance of LMs when providing textual instructions \citep{Brown2020LanguageMA,Hendrycks2020MeasuringMM, Srivastava2022BeyondTI}. 
While LMs show substantial performance on application-oriented tasks \citep{liang2023holistic} or mathematical reasoning  
\citep{cobbe2021gsm8k}, such evaluations are sensible to specific formulations \citep{mizrahi2023state} or metrics \citep{DBLP:conf/nips/SchaefferMK23} employed.
Thus, results of different benchmarks were found to disagree substantially \citep{yuan2024probelmplausibilityrankingevaluation,perlitz2024benchmarkagreementtestingright}. 

\paragraph{Assessing the Linguistic Competence of LMs}
Analyzing LMs' linguistic competence started with static word vectors \citep{kohn-2015-whats}, sentence embeddings \citep{conneau-etal-2018-cram,adi2017finegrained}, the internals of translation models \citep{shi-etal-2016-string,bau2018identifying}, or contextualized LMs \citep{tenney2018what, tenney-etal-2019-bert, hewitt-manning-2019-structural}.  
Other methodological work addressed the validity of obtained results with control tasks \citep{hewitt-liang-2019-designing} or from an information theory perspective \citep{voita-titov-2020-information,pimentel2020information}, or studied causal effects \citep{elazar-etal-2021-amnesic}.
While further studies focus on whether LMs follow human understanding of linguistic competence when solving downstream tasks \citep{belinkov-2022-probing,aw2023instruction,mahowald2023dissociating},  \citet{mosbach-etal-2020-interplay-fine} and \citet{waldis2024dive} found that downstream task fine-tuning hurts the understanding of linguistic phenomena.   

In contrast to prior studies, \Holmes{} assesses the linguistic competence of an extensive set of contemporary LMs covering a comprehensive collection of linguistic phenomena.
Unlike other work evaluating linguistic phenomena \citep{blevins-etal-2023-prompting,amouyal2024large} using prompting leading to unreliable results \citep{liang2023holistic}, probing allows \Holmes{} to reliably and comprehensively compare LMs regardless of architecture or pre-training. 
As a result,  \Holmes{} can address recent calls to thoroughly and explicitly evaluate linguistic phenomena \citep{hu-levy-2023-prompting,lu2023emergent,mahowald2023dissociating}.


\section{Conclusion}
\Holmes{} marks the most up-to-date and extensive consolidation of existing resources addressing the need to assess the linguistic competence of LMs in isolation.
Our experiments demonstrate that LMs' linguistic competence is pronounced regarding formal phenomena but lacks functional ones when information about broader textual contexts, such as rhetorical structure, is required.
Simultaneously, size, architecture, and instruction tuning are crucial factors for differences among LMs. 
As LM and resources in linguistics constantly grow, we will actively extend \Holmes{} with new datasets and upcoming LMs.

\section*{Ethical Considerations and Limitations}

\paragraph{Language}
\Holmes{} as well as \FlashHolmes{} solely assess linguistic phenomena for the English language. 
As we plan to expand the benchmark and scope of multilingual data, we focus momentarily on English because of the widespread availability of resources, including curated corpora and the diversity of available LMs.

\paragraph{Last Layer Internal Representation}
Given the extensive scope of the analysis presented in this work, we focus on examining the internal representation of language models (LMs) through the output of their last layer. 
While this analysis provides valuable insights, it only partially captures the complexity inherent in LMs across all their layers. 
To facilitate further research into the comprehensive analysis of LMs, we see \Holmes{} providing groundwork, including the release of the specific tasks in a unifying format and corresponding evaluation code, which can be easily adapted to investigate specific layers of LMs.

\paragraph{Coverage}
We agree with \citet{liang2023holistic} and see one fundamental aspect in composing a benchmark in acknowledging its incompleteness. 
Linguistic phenomena, LMs, and underlying meta-studies are a subset of the variety of available resources. 
We consolidated them carefully to provide a comprehensive scope of the linguistic competence and various LMs. 
However, as benchmarks evolve as tools to assess LMs, we will further expand \Holmes{} both with the existing and upcoming LMs and data resources.

\paragraph{Data Availability}
Linguistic annotations, in particular more complex ones targeting phenomena like \textit{discourse}, are money and time-wise expensive.
Out of 208 datasets included in \Holmes{}, 18 probing datasets are based on licensed resources and are not freely available.  
However, with \FlashHolmes{}, we provide an effective and efficient alternative based on open-access resources. 
Furthermore, upon confirming the granted access, we are happy to share our probing datasets, including those based on the licensed resources. 

\paragraph{Bias}
As \Holmes{} relies on existing resources, it inherits the bias embodied in these datasets.
Examples of such bias are gender equality or gender fairness, like the use of neo pronouns such as \textit{em} in \citet{lauscher-etal-2023-em}.

\paragraph{Dataset Contamination}
\Holmes{} encompasses a large collection of established datasets, like OntoNotes \citep{weischedel2013ontonotes}.
While we solely rely on LMs with open-sourced weights, the training or instruction-tuning data is not known for all of them, as for the Llama-2 \citep{touvron2023llama}, Mixtral \citep{jiang2024mixtral}, or Wizard \citep{DBLP:journals/corr/abs-2304-12244} LMs. 
Therefore, we need to expect that some texts were part of the LMs' pre-training corpora and that specific tasks, such as named-entity recognition (NER), were used during instruction tuning.
However, instruction-tuning aligns LMs' linguistic performance to produce coherent text responding to specific textual instruction provided and does not align LMs' internal representations explicitly \citep{Brown2020LanguageMA, touvron2023llama, jiang2024mixtral}. 
As \Holmes{} evaluates the linguistic competence using LMs' internal representations, it retains its validity even under potential data contamination \citep{balloccu-etal-2024-leak}.
Building upon our results, showing that downstream abilities are partly reflected in LMs' internal representations, one could examine whether instruction-tuning injects task-specific information into LMs' internal representations, thereby detecting task contamination.

\section*{Acknowledgments}
We thank Yongxin Huang, Tim Baumgärtner, Ji-Ung Lee, the action editor, and anonymous reviewers for their valuable feedback and discussions.
Andreas Waldis is supported by the Hasler Foundation Grant No. 21024.

\bibliography{anthology,custom}

\begin{thebibliography}{141}
\expandafter\ifx\csname natexlab\endcsname\relax\def\natexlab#1{#1}\fi

\bibitem[{Adi et~al.(2017)Adi, Kermany, Belinkov, Lavi, and
  Goldberg}]{adi2017finegrained}
Yossi Adi, Einat Kermany, Yonatan Belinkov, Ofer Lavi, and Yoav Goldberg. 2017.
\newblock \href {https://openreview.net/forum?id=BJh6Ztuxl} {Fine-grained
  analysis of sentence embeddings using auxiliary prediction tasks}.
\newblock In \emph{5th International Conference on Learning Representations,
  {ICLR} 2017, Toulon, France, April 24-26, 2017, Conference Track
  Proceedings}. OpenReview.net.

\bibitem[{Aghazadeh et~al.(2022)Aghazadeh, Fayyaz, and
  Yaghoobzadeh}]{aghazadeh-etal-2022-metaphors}
Ehsan Aghazadeh, Mohsen Fayyaz, and Yadollah Yaghoobzadeh. 2022.
\newblock \href {https://doi.org/10.18653/v1/2022.acl-long.144} {Metaphors in
  pre-trained language models: Probing and generalization across datasets and
  languages}.
\newblock In \emph{Proceedings of the 60th Annual Meeting of the Association
  for Computational Linguistics (Volume 1: Long Papers)}, pages 2037--2050,
  Dublin, Ireland. Association for Computational Linguistics.

\bibitem[{Amouyal et~al.(2024)Amouyal, Meltzer-Asscher, and
  Berant}]{amouyal2024large}
Samuel Amouyal, Aya Meltzer-Asscher, and Jonathan Berant. 2024.
\newblock \href {https://aclanthology.org/2024.findings-eacl.12} {Large
  language models for psycholinguistic plausibility pretesting}.
\newblock In \emph{Findings of the Association for Computational Linguistics:
  EACL 2024}, pages 166--181, St. Julian{'}s, Malta. Association for
  Computational Linguistics.

\bibitem[{Aw et~al.(2023)Aw, Montariol, AlKhamissi, Schrimpf, and
  Bosselut}]{aw2023instruction}
Khai~Loong Aw, Syrielle Montariol, Badr AlKhamissi, Martin Schrimpf, and
  Antoine Bosselut. 2023.
\newblock \href {https://doi.org/10.48550/ARXIV.2312.00575} {Instruction-tuning
  aligns llms to the human brain}.
\newblock \emph{CoRR}, abs/2312.00575.

\bibitem[{Balloccu et~al.(2024)Balloccu, Schmidtov{\'a}, Lango, and
  Dusek}]{balloccu-etal-2024-leak}
Simone Balloccu, Patr{\'\i}cia Schmidtov{\'a}, Mateusz Lango, and Ondrej Dusek.
  2024.
\newblock \href {https://aclanthology.org/2024.eacl-long.5} {Leak, cheat,
  repeat: Data contamination and evaluation malpractices in closed-source
  {LLM}s}.
\newblock In \emph{Proceedings of the 18th Conference of the European Chapter
  of the Association for Computational Linguistics (Volume 1: Long Papers)},
  pages 67--93, St. Julian{'}s, Malta. Association for Computational
  Linguistics.

\bibitem[{Bau et~al.(2019)Bau, Belinkov, Sajjad, Durrani, Dalvi, and
  Glass}]{bau2018identifying}
Anthony Bau, Yonatan Belinkov, Hassan Sajjad, Nadir Durrani, Fahim Dalvi, and
  James~R. Glass. 2019.
\newblock \href {https://openreview.net/forum?id=H1z-PsR5KX} {Identifying and
  controlling important neurons in neural machine translation}.
\newblock In \emph{7th International Conference on Learning Representations,
  {ICLR} 2019, New Orleans, LA, USA, May 6-9, 2019}. OpenReview.net.

\bibitem[{Beeching et~al.(2023)Beeching, Fourrier, Habib, Han, Lambert, Rajani,
  Sanseviero, Tunstall, and Wolf}]{Pham_OpenLLM_Operating_LLMs_2023}
Edward Beeching, Clémentine Fourrier, Nathan Habib, Sheon Han, Nathan Lambert,
  Nazneen Rajani, Omar Sanseviero, Lewis Tunstall, and Thomas Wolf. 2023.
\newblock Open llm leaderboard.
\newblock
  \url{https://huggingface.co/spaces/HuggingFaceH4/open_llm_leaderboard}.

\bibitem[{Belinkov(2022)}]{belinkov-2022-probing}
Yonatan Belinkov. 2022.
\newblock \href {https://doi.org/10.1162/coli_a_00422} {Probing classifiers:
  Promises, shortcomings, and advances}.
\newblock \emph{Computational Linguistics}, 48(1):207--219.

\bibitem[{Belinkov et~al.(2017)Belinkov, Durrani, Dalvi, Sajjad, and
  Glass}]{belinkov-etal-2017-neural}
Yonatan Belinkov, Nadir Durrani, Fahim Dalvi, Hassan Sajjad, and James Glass.
  2017.
\newblock \href {https://doi.org/10.18653/v1/P17-1080} {What do neural machine
  translation models learn about morphology?}
\newblock In \emph{Proceedings of the 55th Annual Meeting of the Association
  for Computational Linguistics (Volume 1: Long Papers)}, pages 861--872,
  Vancouver, Canada. Association for Computational Linguistics.

\bibitem[{Biderman et~al.(2023)Biderman, Schoelkopf, Anthony, Bradley, O'Brien,
  Hallahan, Khan, Purohit, Prashanth, Raff, Skowron, Sutawika, and van~der
  Wal}]{biderman2023pythia}
Stella Biderman, Hailey Schoelkopf, Quentin~Gregory Anthony, Herbie Bradley,
  Kyle O'Brien, Eric Hallahan, Mohammad~Aflah Khan, Shivanshu Purohit,
  USVSN~Sai Prashanth, Edward Raff, Aviya Skowron, Lintang Sutawika, and Oskar
  van~der Wal. 2023.
\newblock \href {https://proceedings.mlr.press/v202/biderman23a.html} {Pythia:
  {A} suite for analyzing large language models across training and scaling}.
\newblock In \emph{International Conference on Machine Learning, {ICML} 2023,
  23-29 July 2023, Honolulu, Hawaii, {USA}}, volume 202 of \emph{Proceedings of
  Machine Learning Research}, pages 2397--2430. {PMLR}.

\bibitem[{Birke and Sarkar(2006)}]{Birke2006ACA}
Julia Birke and Anoop Sarkar. 2006.
\newblock \href {https://aclanthology.org/E06-1042} {A clustering approach for
  nearly unsupervised recognition of nonliteral language}.
\newblock In \emph{11th Conference of the {E}uropean Chapter of the Association
  for Computational Linguistics}, pages 329--336, Trento, Italy. Association
  for Computational Linguistics.

\bibitem[{Blevins et~al.(2023{\natexlab{a}})Blevins, Gonen, and
  Zettlemoyer}]{blevins-etal-2023-prompting2}
Terra Blevins, Hila Gonen, and Luke Zettlemoyer. 2023{\natexlab{a}}.
\newblock \href {https://doi.org/10.18653/v1/2023.acl-long.367} {Prompting
  language models for linguistic structure}.
\newblock In \emph{Proceedings of the 61st Annual Meeting of the Association
  for Computational Linguistics (Volume 1: Long Papers)}, pages 6649--6663,
  Toronto, Canada. Association for Computational Linguistics.

\bibitem[{Blevins et~al.(2023{\natexlab{b}})Blevins, Gonen, and
  Zettlemoyer}]{blevins-etal-2023-prompting}
Terra Blevins, Hila Gonen, and Luke Zettlemoyer. 2023{\natexlab{b}}.
\newblock \href {https://doi.org/10.18653/v1/2023.acl-long.367} {Prompting
  language models for linguistic structure}.
\newblock In \emph{Proceedings of the 61st Annual Meeting of the Association
  for Computational Linguistics (Volume 1: Long Papers)}, pages 6649--6663,
  Toronto, Canada. Association for Computational Linguistics.

\bibitem[{Bogoychev and Lopez(2016)}]{bogoychev-lopez-2016-n}
Nikolay Bogoychev and Adam Lopez. 2016.
\newblock \href {https://doi.org/10.18653/v1/P16-1183} {N-gram language models
  for massively parallel devices}.
\newblock In \emph{Proceedings of the 54th Annual Meeting of the Association
  for Computational Linguistics (Volume 1: Long Papers)}, pages 1944--1953,
  Berlin, Germany. Association for Computational Linguistics.

\bibitem[{Brown et~al.(2020)Brown, Mann, Ryder, Subbiah, Kaplan, Dhariwal,
  Neelakantan, Shyam, Sastry, Askell, Agarwal, Herbert{-}Voss, Krueger,
  Henighan, Child, Ramesh, Ziegler, Wu, Winter, Hesse, Chen, Sigler, Litwin,
  Gray, Chess, Clark, Berner, McCandlish, Radford, Sutskever, and
  Amodei}]{Brown2020LanguageMA}
Tom~B. Brown, Benjamin Mann, Nick Ryder, Melanie Subbiah, Jared Kaplan,
  Prafulla Dhariwal, Arvind Neelakantan, Pranav Shyam, Girish Sastry, Amanda
  Askell, Sandhini Agarwal, Ariel Herbert{-}Voss, Gretchen Krueger, Tom
  Henighan, Rewon Child, Aditya Ramesh, Daniel~M. Ziegler, Jeffrey Wu, Clemens
  Winter, Christopher Hesse, Mark Chen, Eric Sigler, Mateusz Litwin, Scott
  Gray, Benjamin Chess, Jack Clark, Christopher Berner, Sam McCandlish, Alec
  Radford, Ilya Sutskever, and Dario Amodei. 2020.
\newblock \href
  {https://proceedings.neurips.cc/paper/2020/hash/1457c0d6bfcb4967418bfb8ac142f64a-Abstract.html}
  {Language models are few-shot learners}.
\newblock In \emph{Advances in Neural Information Processing Systems 33: Annual
  Conference on Neural Information Processing Systems 2020, NeurIPS 2020,
  December 6-12, 2020, virtual}.

\bibitem[{Carlson et~al.(2001)Carlson, Marcu, and
  Okurovsky}]{carlson-etal-2001}
Lynn Carlson, Daniel Marcu, and Mary~Ellen Okurovsky. 2001.
\newblock \href {https://aclanthology.org/W01-1605/} {Building a
  discourse-tagged corpus in the framework of rhetorical structure theory}.
\newblock In \emph{Proceedings of the {SIGDIAL} 2001 Workshop, The 2nd Annual
  Meeting of the Special Interest Group on Discourse and Dialogue, Saturday,
  September 1, 2001 to Sunday, September 2, 2001, Aalborg, Denmark}. The
  Association for Computer Linguistics.

\bibitem[{Chomsky(1965)}]{chomsky1965}
Noam Chomsky. 1965.
\newblock \href
  {http://www.amazon.com/Aspects-Theory-Syntax-Noam-Chomsky/dp/0262530074}
  {\emph{Aspects of the Theory of Syntax}}.
\newblock The MIT Press, Cambridge.

\bibitem[{Chung et~al.(2022)Chung, Hou, Longpre, Zoph, Tay, Fedus, Li, Wang,
  Dehghani, Brahma, Webson, Gu, Dai, Suzgun, Chen, Chowdhery, Narang, Mishra,
  Yu, Zhao, Huang, Dai, Yu, Petrov, Chi, Dean, Devlin, Roberts, Zhou, Le, and
  Wei}]{chung2022scaling}
Hyung~Won Chung, Le~Hou, Shayne Longpre, Barret Zoph, Yi~Tay, William Fedus,
  Eric Li, Xuezhi Wang, Mostafa Dehghani, Siddhartha Brahma, Albert Webson,
  Shixiang~Shane Gu, Zhuyun Dai, Mirac Suzgun, Xinyun Chen, Aakanksha
  Chowdhery, Sharan Narang, Gaurav Mishra, Adams Yu, Vincent~Y. Zhao, Yanping
  Huang, Andrew~M. Dai, Hongkun Yu, Slav Petrov, Ed~H. Chi, Jeff Dean, Jacob
  Devlin, Adam Roberts, Denny Zhou, Quoc~V. Le, and Jason Wei. 2022.
\newblock \href {https://doi.org/10.48550/ARXIV.2210.11416} {Scaling
  instruction-finetuned language models}.
\newblock \emph{CoRR}, abs/2210.11416.

\bibitem[{Clark et~al.(2020)Clark, Luong, Le, and Manning}]{clark2020electra}
Kevin Clark, Minh{-}Thang Luong, Quoc~V. Le, and Christopher~D. Manning. 2020.
\newblock \href {https://openreview.net/forum?id=r1xMH1BtvB} {{ELECTRA:}
  pre-training text encoders as discriminators rather than generators}.
\newblock In \emph{8th International Conference on Learning Representations,
  {ICLR} 2020, Addis Ababa, Ethiopia, April 26-30, 2020}. OpenReview.net.

\bibitem[{Cobbe et~al.(2021)Cobbe, Kosaraju, Bavarian, Chen, Jun, Kaiser,
  Plappert, Tworek, Hilton, Nakano, Hesse, and Schulman}]{cobbe2021gsm8k}
Karl Cobbe, Vineet Kosaraju, Mohammad Bavarian, Mark Chen, Heewoo Jun, Lukasz
  Kaiser, Matthias Plappert, Jerry Tworek, Jacob Hilton, Reiichiro Nakano,
  Christopher Hesse, and John Schulman. 2021.
\newblock \href {http://arxiv.org/abs/2110.14168} {Training verifiers to solve
  math word problems}.
\newblock \emph{CoRR}, abs/2110.14168.

\bibitem[{Conneau et~al.(2018)Conneau, Kruszewski, Lample, Barrault, and
  Baroni}]{conneau-etal-2018-cram}
Alexis Conneau, German Kruszewski, Guillaume Lample, Lo{\"\i}c Barrault, and
  Marco Baroni. 2018.
\newblock \href {https://doi.org/10.18653/v1/P18-1198} {What you can cram into
  a single {\$}{\&}!{\#}* vector: Probing sentence embeddings for linguistic
  properties}.
\newblock In \emph{Proceedings of the 56th Annual Meeting of the Association
  for Computational Linguistics (Volume 1: Long Papers)}, pages 2126--2136,
  Melbourne, Australia. Association for Computational Linguistics.

\bibitem[{Conover et~al.(2023)Conover, Hayes, Mathur, Xie, Wan, Shah, Ghodsi,
  Wendell, Zaharia, and Xin}]{DatabricksBlog2023DollyV2}
Mike Conover, Matt Hayes, Ankit Mathur, Jianwei Xie, Jun Wan, Sam Shah, Ali
  Ghodsi, Patrick Wendell, Matei Zaharia, and Reynold Xin. 2023.
\newblock \href
  {https://www.databricks.com/blog/2023/04/12/dolly-first-open-commercially-viable-instruction-tuned-llm}
  {Free dolly: Introducing the world's first truly open instruction-tuned llm}.

\bibitem[{Deng et~al.(2009)Deng, Dong, Socher, Li, Li, and
  Fei-Fei}]{Deng2009ImageNetAL}
Jia Deng, Wei Dong, Richard Socher, Li-Jia Li, K.~Li, and Li~Fei-Fei. 2009.
\newblock \href {https://api.semanticscholar.org/CorpusID:57246310} {Imagenet:
  A large-scale hierarchical image database}.
\newblock \emph{2009 IEEE Conference on Computer Vision and Pattern
  Recognition}, pages 248--255.

\bibitem[{Devlin et~al.(2019)Devlin, Chang, Lee, and
  Toutanova}]{devlin-etal-2019-bert}
Jacob Devlin, Ming-Wei Chang, Kenton Lee, and Kristina Toutanova. 2019.
\newblock \href {https://doi.org/10.18653/v1/N19-1423} {{BERT}: Pre-training of
  deep bidirectional transformers for language understanding}.
\newblock In \emph{Proceedings of the 2019 Conference of the North {A}merican
  Chapter of the Association for Computational Linguistics: Human Language
  Technologies, Volume 1 (Long and Short Papers)}, pages 4171--4186,
  Minneapolis, Minnesota. Association for Computational Linguistics.

\bibitem[{Elazar et~al.(2021)Elazar, Ravfogel, Jacovi, and
  Goldberg}]{elazar-etal-2021-amnesic}
Yanai Elazar, Shauli Ravfogel, Alon Jacovi, and Yoav Goldberg. 2021.
\newblock \href {https://doi.org/10.1162/tacl_a_00359} {Amnesic probing:
  Behavioral explanation with amnesic counterfactuals}.
\newblock \emph{Transactions of the Association for Computational Linguistics},
  9:160--175.

\bibitem[{Flesch(1948)}]{flesch1948new}
Rudolf~Franz Flesch. 1948.
\newblock \href {https://api.semanticscholar.org/CorpusID:39344661} {A new
  readability yardstick.}
\newblock \emph{The Journal of applied psychology}, 32(3):221–233.

\bibitem[{Gantt et~al.(2022)Gantt, Glass, and
  White}]{gantt-etal-2022-decomposing}
William Gantt, Lelia Glass, and Aaron~Steven White. 2022.
\newblock \href {https://doi.org/10.1162/tacl_a_00445} {Decomposing and
  recomposing event structure}.
\newblock \emph{Transactions of the Association for Computational Linguistics},
  10:17--34.

\bibitem[{Gautam et~al.(2024)Gautam, Bingert, Zhu, Lauscher, and
  Klakow}]{Gautam2024RobustPU}
Vagrant Gautam, Eileen Bingert, D.~Zhu, Anne Lauscher, and Dietrich Klakow.
  2024.
\newblock \href {https://doi.org/10.48550/ARXIV.2404.03134} {Robust pronoun use
  fidelity with english llms: Are they reasoning, repeating, or just biased?}
\newblock \emph{CoRR}, abs/2404.03134.

\bibitem[{Giulianelli et~al.(2018)Giulianelli, Harding, Mohnert, Hupkes, and
  Zuidema}]{giulianelli-etal-2018-hood}
Mario Giulianelli, Jack Harding, Florian Mohnert, Dieuwke Hupkes, and Willem
  Zuidema. 2018.
\newblock \href {https://doi.org/10.18653/v1/W18-5426} {Under the hood: Using
  diagnostic classifiers to investigate and improve how language models track
  agreement information}.
\newblock In \emph{Proceedings of the 2018 {EMNLP} Workshop {B}lackbox{NLP}:
  Analyzing and Interpreting Neural Networks for {NLP}}, pages 240--248,
  Brussels, Belgium. Association for Computational Linguistics.

\bibitem[{Govindarajan et~al.(2019)Govindarajan, Van~Durme, and
  White}]{govindarajan-etal-2019-decomposing}
Venkata Govindarajan, Benjamin Van~Durme, and Aaron~Steven White. 2019.
\newblock \href {https://doi.org/10.1162/tacl_a_00285} {Decomposing
  generalization: Models of generic, habitual, and episodic statements}.
\newblock \emph{Transactions of the Association for Computational Linguistics},
  7:501--517.

\bibitem[{Gupta et~al.(2015)Gupta, Boleda, Baroni, and
  Pad{\'o}}]{gupta-etal-2015-distributional}
Abhijeet Gupta, Gemma Boleda, Marco Baroni, and Sebastian Pad{\'o}. 2015.
\newblock \href {https://doi.org/10.18653/v1/D15-1002} {Distributional vectors
  encode referential attributes}.
\newblock In \emph{Proceedings of the 2015 Conference on Empirical Methods in
  Natural Language Processing}, pages 12--21, Lisbon, Portugal. Association for
  Computational Linguistics.

\bibitem[{Harris(1954)}]{harris1954distributional}
Zellig~S Harris. 1954.
\newblock \href {https://api.semanticscholar.org/CorpusID:86680084}
  {Distributional structure}.
\newblock \emph{Word}, 10(2-3):146--162.

\bibitem[{He et~al.(2023)He, Gao, and Chen}]{he2022debertav3}
Pengcheng He, Jianfeng Gao, and Weizhu Chen. 2023.
\newblock \href {https://openreview.net/pdf?id=sE7-XhLxHA} {Debertav3:
  Improving deberta using electra-style pre-training with gradient-disentangled
  embedding sharing}.
\newblock In \emph{The Eleventh International Conference on Learning
  Representations, {ICLR} 2023, Kigali, Rwanda, May 1-5, 2023}. OpenReview.net.

\bibitem[{He et~al.(2021)He, Liu, Gao, and Chen}]{he2021deberta}
Pengcheng He, Xiaodong Liu, Jianfeng Gao, and Weizhu Chen. 2021.
\newblock \href {https://openreview.net/forum?id=XPZIaotutsD} {Deberta:
  decoding-enhanced bert with disentangled attention}.
\newblock In \emph{9th International Conference on Learning Representations,
  {ICLR} 2021, Virtual Event, Austria, May 3-7, 2021}. OpenReview.net.

\bibitem[{Hendrickx et~al.(2010)Hendrickx, Kim, Kozareva, Nakov,
  {\'O}~S{\'e}aghdha, Pad{\'o}, Pennacchiotti, Romano, and
  Szpakowicz}]{hendrickx-etal-2010-semeval}
Iris Hendrickx, Su~Nam Kim, Zornitsa Kozareva, Preslav Nakov, Diarmuid
  {\'O}~S{\'e}aghdha, Sebastian Pad{\'o}, Marco Pennacchiotti, Lorenza Romano,
  and Stan Szpakowicz. 2010.
\newblock \href {https://aclanthology.org/S10-1006} {{S}em{E}val-2010 task 8:
  Multi-way classification of semantic relations between pairs of nominals}.
\newblock In \emph{Proceedings of the 5th International Workshop on Semantic
  Evaluation}, pages 33--38, Uppsala, Sweden. Association for Computational
  Linguistics.

\bibitem[{Hendrycks et~al.(2021)Hendrycks, Burns, Basart, Zou, Mazeika, Song,
  and Steinhardt}]{Hendrycks2020MeasuringMM}
Dan Hendrycks, Collin Burns, Steven Basart, Andy Zou, Mantas Mazeika, Dawn
  Song, and Jacob Steinhardt. 2021.
\newblock \href {https://openreview.net/forum?id=d7KBjmI3GmQ} {Measuring
  massive multitask language understanding}.
\newblock In \emph{9th International Conference on Learning Representations,
  {ICLR} 2021, Virtual Event, Austria, May 3-7, 2021}. OpenReview.net.

\bibitem[{Hershcovitch et~al.(2024)Hershcovitch, Choshen, Wood, Enmouri, Chin,
  Sundararaman, and Harnik}]{hershcovitch2024lossless}
Moshik Hershcovitch, Leshem Choshen, Andrew Wood, Ilias Enmouri, Peter Chin,
  Swaminathan Sundararaman, and Danny Harnik. 2024.
\newblock \href {https://doi.org/10.48550/arXiv.2404.15198} {Lossless and
  near-lossless compression for foundation models}.
\newblock \emph{CoRR}, abs/2404.15198.

\bibitem[{Hewitt and Liang(2019)}]{hewitt-liang-2019-designing}
John Hewitt and Percy Liang. 2019.
\newblock \href {https://doi.org/10.18653/v1/D19-1275} {Designing and
  interpreting probes with control tasks}.
\newblock In \emph{Proceedings of the 2019 Conference on Empirical Methods in
  Natural Language Processing and the 9th International Joint Conference on
  Natural Language Processing (EMNLP-IJCNLP)}, pages 2733--2743, Hong Kong,
  China. Association for Computational Linguistics.

\bibitem[{Hewitt and Manning(2019)}]{hewitt-manning-2019-structural}
John Hewitt and Christopher~D. Manning. 2019.
\newblock \href {https://doi.org/10.18653/v1/N19-1419} {{A} structural probe
  for finding syntax in word representations}.
\newblock In \emph{Proceedings of the 2019 Conference of the North {A}merican
  Chapter of the Association for Computational Linguistics: Human Language
  Technologies, Volume 1 (Long and Short Papers)}, pages 4129--4138,
  Minneapolis, Minnesota. Association for Computational Linguistics.

\bibitem[{Hou(2018)}]{hou-2018-enhanced}
Yufang Hou. 2018.
\newblock \href {https://doi.org/10.18653/v1/N18-2001} {Enhanced word
  representations for bridging anaphora resolution}.
\newblock In \emph{Proceedings of the 2018 Conference of the North {A}merican
  Chapter of the Association for Computational Linguistics: Human Language
  Technologies, Volume 2 (Short Papers)}, pages 1--7, New Orleans, Louisiana.
  Association for Computational Linguistics.

\bibitem[{Hou(2020)}]{hou-2020-bridging}
Yufang Hou. 2020.
\newblock \href {https://doi.org/10.18653/v1/2020.acl-main.132} {Bridging
  anaphora resolution as question answering}.
\newblock In \emph{Proceedings of the 58th Annual Meeting of the Association
  for Computational Linguistics}, pages 1428--1438, Online. Association for
  Computational Linguistics.

\bibitem[{Hou et~al.(2024)Hou, Pascale, Carnerero-Cano, Tchrakian, Marinescu,
  Daly, Padhi, and Sattigeri}]{hou2024wikicontradictbenchmarkevaluatingllms}
Yufang Hou, Alessandra Pascale, Javier Carnerero-Cano, Tigran Tchrakian, Radu
  Marinescu, Elizabeth Daly, Inkit Padhi, and Prasanna Sattigeri. 2024.
\newblock \href {http://arxiv.org/abs/2406.13805} {Wikicontradict: A benchmark
  for evaluating llms on real-world knowledge conflicts from wikipedia}.
\newblock \emph{arXiv preprint arXiv:2406.13805}.

\bibitem[{Hu and Levy(2023)}]{hu-levy-2023-prompting}
Jennifer Hu and Roger Levy. 2023.
\newblock \href {https://doi.org/10.18653/v1/2023.emnlp-main.306} {Prompting is
  not a substitute for probability measurements in large language models}.
\newblock In \emph{Proceedings of the 2023 Conference on Empirical Methods in
  Natural Language Processing}, pages 5040--5060, Singapore. Association for
  Computational Linguistics.

\bibitem[{Huebner et~al.(2021)Huebner, Sulem, Cynthia, and
  Roth}]{huebner-etal-2021-babyberta}
Philip~A. Huebner, Elior Sulem, Fisher Cynthia, and Dan Roth. 2021.
\newblock \href {https://doi.org/10.18653/v1/2021.conll-1.49} {{B}aby{BERT}a:
  Learning more grammar with small-scale child-directed language}.
\newblock In \emph{Proceedings of the 25th Conference on Computational Natural
  Language Learning}, pages 624--646, Online. Association for Computational
  Linguistics.

\bibitem[{Hupkes and Zuidema(2018)}]{ijcai2018p796}
Dieuwke Hupkes and Willem Zuidema. 2018.
\newblock \href {https://doi.org/10.24963/ijcai.2018/796} {Visualisation and
  'diagnostic classifiers' reveal how recurrent and recursive neural networks
  process hierarchical structure (extended abstract)}.
\newblock In \emph{Proceedings of the Twenty-Seventh International Joint
  Conference on Artificial Intelligence, {IJCAI-18}}, pages 5617--5621.
  International Joint Conferences on Artificial Intelligence Organization.

\bibitem[{Immer et~al.(2022)Immer, Torroba~Hennigen, Fortuin, and
  Cotterell}]{immer-etal-2022-probing}
Alexander Immer, Lucas Torroba~Hennigen, Vincent Fortuin, and Ryan Cotterell.
  2022.
\newblock \href {https://doi.org/10.18653/v1/2022.acl-long.129} {Probing as
  quantifying inductive bias}.
\newblock In \emph{Proceedings of the 60th Annual Meeting of the Association
  for Computational Linguistics (Volume 1: Long Papers)}, pages 1839--1851,
  Dublin, Ireland. Association for Computational Linguistics.

\bibitem[{Jiang et~al.(2023)Jiang, Sablayrolles, Mensch, Bamford, Chaplot,
  de~Las~Casas, Bressand, Lengyel, Lample, Saulnier, Lavaud, Lachaux, Stock,
  Scao, Lavril, Wang, Lacroix, and Sayed}]{DBLP:journals/corr/abs-2310-06825}
Albert~Q. Jiang, Alexandre Sablayrolles, Arthur Mensch, Chris Bamford,
  Devendra~Singh Chaplot, Diego de~Las~Casas, Florian Bressand, Gianna Lengyel,
  Guillaume Lample, Lucile Saulnier, L{\'{e}}lio~Renard Lavaud, Marie{-}Anne
  Lachaux, Pierre Stock, Teven~Le Scao, Thibaut Lavril, Thomas Wang,
  Timoth{\'{e}}e Lacroix, and William~El Sayed. 2023.
\newblock \href {https://doi.org/10.48550/ARXIV.2310.06825} {Mistral 7b}.
\newblock \emph{CoRR}, abs/2310.06825.

\bibitem[{Jiang et~al.(2024)Jiang, Sablayrolles, Roux, Mensch, Savary, Bamford,
  Chaplot, de~Las~Casas, Hanna, Bressand, Lengyel, Bour, Lample, Lavaud,
  Saulnier, Lachaux, Stock, Subramanian, Yang, Antoniak, Scao, Gervet, Lavril,
  Wang, Lacroix, and Sayed}]{jiang2024mixtral}
Albert~Q. Jiang, Alexandre Sablayrolles, Antoine Roux, Arthur Mensch, Blanche
  Savary, Chris Bamford, Devendra~Singh Chaplot, Diego de~Las~Casas, Emma~Bou
  Hanna, Florian Bressand, Gianna Lengyel, Guillaume Bour, Guillaume Lample,
  L{\'{e}}lio~Renard Lavaud, Lucile Saulnier, Marie{-}Anne Lachaux, Pierre
  Stock, Sandeep Subramanian, Sophia Yang, Szymon Antoniak, Teven~Le Scao,
  Th{\'{e}}ophile Gervet, Thibaut Lavril, Thomas Wang, Timoth{\'{e}}e Lacroix,
  and William~El Sayed. 2024.
\newblock \href {https://doi.org/10.48550/ARXIV.2401.04088} {Mixtral of
  experts}.
\newblock \emph{CoRR}, abs/2401.04088.

\bibitem[{Kassner et~al.(2021)Kassner, Dufter, and
  Sch{\"u}tze}]{kassner-etal-2021-multilingual}
Nora Kassner, Philipp Dufter, and Hinrich Sch{\"u}tze. 2021.
\newblock \href {https://doi.org/10.18653/v1/2021.eacl-main.284} {Multilingual
  {LAMA}: Investigating knowledge in multilingual pretrained language models}.
\newblock In \emph{Proceedings of the 16th Conference of the European Chapter
  of the Association for Computational Linguistics: Main Volume}, pages
  3250--3258, Online. Association for Computational Linguistics.

\bibitem[{Klafka and Ettinger(2020)}]{klafka-ettinger-2020-spying}
Josef Klafka and Allyson Ettinger. 2020.
\newblock \href {https://doi.org/10.18653/v1/2020.acl-main.434} {Spying on your
  neighbors: Fine-grained probing of contextual embeddings for information
  about surrounding words}.
\newblock In \emph{Proceedings of the 58th Annual Meeting of the Association
  for Computational Linguistics}, pages 4801--4811, Online. Association for
  Computational Linguistics.

\bibitem[{K{\"o}hn(2015)}]{kohn-2015-whats}
Arne K{\"o}hn. 2015.
\newblock \href {https://doi.org/10.18653/v1/D15-1246} {What{'}s in an
  embedding? analyzing word embeddings through multilingual evaluation}.
\newblock In \emph{Proceedings of the 2015 Conference on Empirical Methods in
  Natural Language Processing}, pages 2067--2073, Lisbon, Portugal. Association
  for Computational Linguistics.

\bibitem[{Konstantinova et~al.(2012)Konstantinova, de~Sousa, Cruz, Ma{\~n}a,
  Taboada, and Mitkov}]{konstantinova-etal-2012-review}
Natalia Konstantinova, Sheila~C.M. de~Sousa, Noa~P. Cruz, Manuel~J. Ma{\~n}a,
  Maite Taboada, and Ruslan Mitkov. 2012.
\newblock \href
  {http://www.lrec-conf.org/proceedings/lrec2012/pdf/533_Paper.pdf} {A review
  corpus annotated for negation, speculation and their scope}.
\newblock In \emph{Proceedings of the Eighth International Conference on
  Language Resources and Evaluation ({LREC}'12)}, pages 3190--3195, Istanbul,
  Turkey. European Language Resources Association (ELRA).

\bibitem[{Koto et~al.(2021)Koto, Lau, and Baldwin}]{koto-etal-2021-discourse}
Fajri Koto, Jey~Han Lau, and Timothy Baldwin. 2021.
\newblock \href {https://doi.org/10.18653/v1/2021.naacl-main.301} {Discourse
  probing of pretrained language models}.
\newblock In \emph{Proceedings of the 2021 Conference of the North American
  Chapter of the Association for Computational Linguistics: Human Language
  Technologies}, pages 3849--3864, Online. Association for Computational
  Linguistics.

\bibitem[{Krasnowska-Kiera{\'s} and
  Wr{\'o}blewska(2019)}]{krasnowska-kieras-wroblewska-2019-empirical}
Katarzyna Krasnowska-Kiera{\'s} and Alina Wr{\'o}blewska. 2019.
\newblock \href {https://doi.org/10.18653/v1/P19-1573} {Empirical linguistic
  study of sentence embeddings}.
\newblock In \emph{Proceedings of the 57th Annual Meeting of the Association
  for Computational Linguistics}, pages 5729--5739, Florence, Italy.
  Association for Computational Linguistics.

\bibitem[{Kurfal{\i} and {\"O}stling(2021)}]{kurfali-ostling-2021-probing}
Murathan Kurfal{\i} and Robert {\"O}stling. 2021.
\newblock \href {https://doi.org/10.18653/v1/2021.repl4nlp-1.2} {Probing
  multilingual language models for discourse}.
\newblock In \emph{Proceedings of the 6th Workshop on Representation Learning
  for NLP (RepL4NLP-2021)}, pages 8--19, Online. Association for Computational
  Linguistics.

\bibitem[{Lan et~al.(2020)Lan, Chen, Goodman, Gimpel, Sharma, and
  Soricut}]{Lan2020ALBERTAL}
Zhenzhong Lan, Mingda Chen, Sebastian Goodman, Kevin Gimpel, Piyush Sharma, and
  Radu Soricut. 2020.
\newblock \href {https://openreview.net/forum?id=H1eA7AEtvS} {{ALBERT:} {A}
  lite {BERT} for self-supervised learning of language representations}.
\newblock In \emph{8th International Conference on Learning Representations,
  {ICLR} 2020, Addis Ababa, Ethiopia, April 26-30, 2020}. OpenReview.net.

\bibitem[{Lauscher et~al.(2023)Lauscher, Nozza, Miltersen, Crowley, and
  Hovy}]{lauscher-etal-2023-em}
Anne Lauscher, Debora Nozza, Ehm Miltersen, Archie Crowley, and Dirk Hovy.
  2023.
\newblock \href {https://doi.org/10.18653/v1/2023.acl-long.23} {What about
  {``}em{''}? how commercial machine translation fails to handle
  (neo-)pronouns}.
\newblock In \emph{Proceedings of the 61st Annual Meeting of the Association
  for Computational Linguistics (Volume 1: Long Papers)}, pages 377--392,
  Toronto, Canada. Association for Computational Linguistics.

\bibitem[{Lewis et~al.(2020)Lewis, Liu, Goyal, Ghazvininejad, Mohamed, Levy,
  Stoyanov, and Zettlemoyer}]{lewis-etal-2020-bart}
Mike Lewis, Yinhan Liu, Naman Goyal, Marjan Ghazvininejad, Abdelrahman Mohamed,
  Omer Levy, Veselin Stoyanov, and Luke Zettlemoyer. 2020.
\newblock \href {https://doi.org/10.18653/v1/2020.acl-main.703} {{BART}:
  Denoising sequence-to-sequence pre-training for natural language generation,
  translation, and comprehension}.
\newblock In \emph{Proceedings of the 58th Annual Meeting of the Association
  for Computational Linguistics}, pages 7871--7880, Online. Association for
  Computational Linguistics.

\bibitem[{Liang et~al.(2023)Liang, Bommasani, Lee, Tsipras, Soylu, Yasunaga,
  Zhang, Narayanan, Wu, Kumar, Newman, Yuan, Yan, Zhang, Cosgrove, Manning, Re,
  Acosta-Navas, Hudson, Zelikman, Durmus, Ladhak, Rong, Ren, Yao, WANG,
  Santhanam, Orr, Zheng, Yuksekgonul, Suzgun, Kim, Guha, Chatterji, Khattab,
  Henderson, Huang, Chi, Xie, Santurkar, Ganguli, Hashimoto, Icard, Zhang,
  Chaudhary, Wang, Li, Mai, Zhang, and Koreeda}]{liang2023holistic}
Percy Liang, Rishi Bommasani, Tony Lee, Dimitris Tsipras, Dilara Soylu,
  Michihiro Yasunaga, Yian Zhang, Deepak Narayanan, Yuhuai Wu, Ananya Kumar,
  Benjamin Newman, Binhang Yuan, Bobby Yan, Ce~Zhang, Christian~Alexander
  Cosgrove, Christopher~D Manning, Christopher Re, Diana Acosta-Navas,
  Drew~Arad Hudson, Eric Zelikman, Esin Durmus, Faisal Ladhak, Frieda Rong,
  Hongyu Ren, Huaxiu Yao, Jue WANG, Keshav Santhanam, Laurel Orr, Lucia Zheng,
  Mert Yuksekgonul, Mirac Suzgun, Nathan Kim, Neel Guha, Niladri~S. Chatterji,
  Omar Khattab, Peter Henderson, Qian Huang, Ryan~Andrew Chi, Sang~Michael Xie,
  Shibani Santurkar, Surya Ganguli, Tatsunori Hashimoto, Thomas Icard, Tianyi
  Zhang, Vishrav Chaudhary, William Wang, Xuechen Li, Yifan Mai, Yuhui Zhang,
  and Yuta Koreeda. 2023.
\newblock \href {https://openreview.net/forum?id=iO4LZibEqW} {Holistic
  evaluation of language models}.
\newblock \emph{Transactions on Machine Learning Research}.
\newblock Featured Certification, Expert Certification.

\bibitem[{Linzen et~al.(2016)Linzen, Dupoux, and
  Goldberg}]{linzen-etal-2016-assessing}
Tal Linzen, Emmanuel Dupoux, and Yoav Goldberg. 2016.
\newblock \href {https://doi.org/10.1162/tacl_a_00115} {Assessing the ability
  of {LSTM}s to learn syntax-sensitive dependencies}.
\newblock \emph{Transactions of the Association for Computational Linguistics},
  4:521--535.

\bibitem[{Liu et~al.(2019)Liu, Ott, Goyal, Du, Joshi, Chen, Levy, Lewis,
  Zettlemoyer, and Stoyanov}]{Liu2019RoBERTaAR}
Yinhan Liu, Myle Ott, Naman Goyal, Jingfei Du, Mandar Joshi, Danqi Chen, Omer
  Levy, Mike Lewis, Luke Zettlemoyer, and Veselin Stoyanov. 2019.
\newblock \href {http://arxiv.org/abs/1907.11692} {Roberta: {A} robustly
  optimized {BERT} pretraining approach}.
\newblock \emph{CoRR}, abs/1907.11692.

\bibitem[{Loshchilov and Hutter(2019)}]{adamW2019}
Ilya Loshchilov and Frank Hutter. 2019.
\newblock \href {https://openreview.net/forum?id=Bkg6RiCqY7} {Decoupled weight
  decay regularization}.
\newblock In \emph{7th International Conference on Learning Representations,
  {ICLR} 2019, New Orleans, LA, USA, May 6-9, 2019}. OpenReview.net.

\bibitem[{Lu et~al.(2023)Lu, Bigoulaeva, Sachdeva, Madabushi, and
  Gurevych}]{lu2023emergent}
Sheng Lu, Irina Bigoulaeva, Rachneet Sachdeva, Harish~Tayyar Madabushi, and
  Iryna Gurevych. 2023.
\newblock \href {https://doi.org/10.48550/ARXIV.2309.01809} {Are emergent
  abilities in large language models just in-context learning?}
\newblock \emph{CoRR}, abs/2309.01809.

\bibitem[{Mahowald et~al.(2024)Mahowald, Ivanova, Blank, Kanwisher, Tenenbaum,
  and Fedorenko}]{mahowald2023dissociating}
Kyle Mahowald, Anna~A. Ivanova, Idan~A. Blank, Nancy Kanwisher, Joshua~B.
  Tenenbaum, and Evelina Fedorenko. 2024.
\newblock \href {https://doi.org/https://doi.org/10.1016/j.tics.2024.01.011}
  {Dissociating language and thought in large language models}.
\newblock \emph{Trends in Cognitive Sciences}.

\bibitem[{Matthews(2014)}]{Matthews1998TheCO}
Peter~Hugoe Matthews. 2014.
\newblock \href
  {https://www.oxfordreference.com/display/10.1093/acref/9780199675128.001.0001/acref-9780199675128}
  {\emph{The concise Oxford dictionary of linguistics}}.
\newblock Oxford University Press, USA.

\bibitem[{Miller(1995)}]{wordnet-1995}
George~A. Miller. 1995.
\newblock \href {https://doi.org/10.1145/219717.219748} {Wordnet: {A} lexical
  database for english}.
\newblock \emph{Communications of the ACM}, 38(11):39--41.

\bibitem[{Min et~al.(2022)Min, Lyu, Holtzman, Artetxe, Lewis, Hajishirzi, and
  Zettlemoyer}]{min-etal-2022-rethinking}
Sewon Min, Xinxi Lyu, Ari Holtzman, Mikel Artetxe, Mike Lewis, Hannaneh
  Hajishirzi, and Luke Zettlemoyer. 2022.
\newblock \href {https://doi.org/10.18653/v1/2022.emnlp-main.759} {Rethinking
  the role of demonstrations: What makes in-context learning work?}
\newblock In \emph{Proceedings of the 2022 Conference on Empirical Methods in
  Natural Language Processing}, pages 11048--11064, Abu Dhabi, United Arab
  Emirates. Association for Computational Linguistics.

\bibitem[{Mitra et~al.(2023)Mitra, Corro, Mahajan, Codas, Sim{\~{o}}es,
  Agrawal, Chen, Razdaibiedina, Jones, Aggarwal, Palangi, Zheng, Rosset,
  Khanpour, and Awadallah}]{mitra2023orca}
Arindam Mitra, Luciano~Del Corro, Shweti Mahajan, Andr{\'{e}}s Codas, Clarisse
  Sim{\~{o}}es, Sahaj Agrawal, Xuxi Chen, Anastasia Razdaibiedina, Erik Jones,
  Kriti Aggarwal, Hamid Palangi, Guoqing Zheng, Corby Rosset, Hamed Khanpour,
  and Ahmed Awadallah. 2023.
\newblock \href {https://doi.org/10.48550/ARXIV.2311.11045} {Orca 2: Teaching
  small language models how to reason}.
\newblock \emph{CoRR}, abs/2311.11045.

\bibitem[{Mizrahi et~al.(2024)Mizrahi, Kaplan, Malkin, Dror, Shahaf, and
  Stanovsky}]{mizrahi2023state}
Moran Mizrahi, Guy Kaplan, Dan Malkin, Rotem Dror, Dafna Shahaf, and Gabriel
  Stanovsky. 2024.
\newblock \href {https://doi.org/10.48550/ARXIV.2401.00595} {State of what art?
  {A} call for multi-prompt {LLM} evaluation}.
\newblock \emph{CoRR}, abs/2401.00595.

\bibitem[{Mohler et~al.(2016)Mohler, Brunson, Rink, and
  Tomlinson}]{mohler-etal-2016-introducing}
Michael Mohler, Mary Brunson, Bryan Rink, and Marc Tomlinson. 2016.
\newblock \href {https://aclanthology.org/L16-1668} {Introducing the {LCC}
  metaphor datasets}.
\newblock In \emph{Proceedings of the Tenth International Conference on
  Language Resources and Evaluation ({LREC}'16)}, pages 4221--4227,
  Portoro{\v{z}}, Slovenia. European Language Resources Association (ELRA).

\bibitem[{Morante and Blanco(2012)}]{morante-blanco-2012-sem}
Roser Morante and Eduardo Blanco. 2012.
\newblock \href {https://aclanthology.org/S12-1035} {*{SEM} 2012 shared task:
  Resolving the scope and focus of negation}.
\newblock In \emph{*{SEM} 2012: The First Joint Conference on Lexical and
  Computational Semantics {--} Volume 1: Proceedings of the main conference and
  the shared task, and Volume 2: Proceedings of the Sixth International
  Workshop on Semantic Evaluation ({S}em{E}val 2012)}, pages 265--274,
  Montr{\'e}al, Canada. Association for Computational Linguistics.

\bibitem[{Mosbach et~al.(2020{\natexlab{a}})Mosbach, Khokhlova, Hedderich, and
  Klakow}]{mosbach-etal-2020-interplay}
Marius Mosbach, Anna Khokhlova, Michael~A. Hedderich, and Dietrich Klakow.
  2020{\natexlab{a}}.
\newblock \href {https://doi.org/10.18653/v1/2020.blackboxnlp-1.7} {On the
  interplay between fine-tuning and sentence-level probing for linguistic
  knowledge in pre-trained transformers}.
\newblock In \emph{Proceedings of the Third BlackboxNLP Workshop on Analyzing
  and Interpreting Neural Networks for NLP}, pages 68--82, Online. Association
  for Computational Linguistics.

\bibitem[{Mosbach et~al.(2020{\natexlab{b}})Mosbach, Khokhlova, Hedderich, and
  Klakow}]{mosbach-etal-2020-interplay-fine}
Marius Mosbach, Anna Khokhlova, Michael~A. Hedderich, and Dietrich Klakow.
  2020{\natexlab{b}}.
\newblock \href {https://doi.org/10.18653/v1/2020.findings-emnlp.227} {{O}n the
  {I}nterplay {B}etween {F}ine-tuning and {S}entence-level {P}robing for
  {L}inguistic {K}nowledge in {P}re-trained {T}ransformers}.
\newblock In \emph{Findings of the Association for Computational Linguistics:
  EMNLP 2020}, pages 2502--2516, Online. Association for Computational
  Linguistics.

\bibitem[{Muennighoff et~al.(2023)Muennighoff, Tazi, Magne, and
  Reimers}]{muennighoff-etal-2023-mteb}
Niklas Muennighoff, Nouamane Tazi, Loic Magne, and Nils Reimers. 2023.
\newblock \href {https://aclanthology.org/2023.eacl-main.148} {{MTEB}: Massive
  text embedding benchmark}.
\newblock In \emph{Proceedings of the 17th Conference of the European Chapter
  of the Association for Computational Linguistics}, pages 2014--2037,
  Dubrovnik, Croatia. Association for Computational Linguistics.

\bibitem[{Narayan et~al.(2018)Narayan, Cohen, and
  Lapata}]{narayan-etal-2018-dont}
Shashi Narayan, Shay~B. Cohen, and Mirella Lapata. 2018.
\newblock \href {https://doi.org/10.18653/v1/D18-1206} {Don{'}t give me the
  details, just the summary! topic-aware convolutional neural networks for
  extreme summarization}.
\newblock In \emph{Proceedings of the 2018 Conference on Empirical Methods in
  Natural Language Processing}, pages 1797--1807, Brussels, Belgium.
  Association for Computational Linguistics.

\bibitem[{Nie et~al.(2019)Nie, Bennett, and Goodman}]{nie-etal-2019-dissent}
Allen Nie, Erin Bennett, and Noah Goodman. 2019.
\newblock \href {https://doi.org/10.18653/v1/P19-1442} {{D}is{S}ent: Learning
  sentence representations from explicit discourse relations}.
\newblock In \emph{Proceedings of the 57th Annual Meeting of the Association
  for Computational Linguistics}, pages 4497--4510, Florence, Italy.
  Association for Computational Linguistics.

\bibitem[{Nie et~al.(2020)Nie, Williams, Dinan, Bansal, Weston, and
  Kiela}]{nie-etal-2020-adversarial}
Yixin Nie, Adina Williams, Emily Dinan, Mohit Bansal, Jason Weston, and Douwe
  Kiela. 2020.
\newblock \href {https://doi.org/10.18653/v1/2020.acl-main.441} {Adversarial
  {NLI}: A new benchmark for natural language understanding}.
\newblock In \emph{Proceedings of the 58th Annual Meeting of the Association
  for Computational Linguistics}, pages 4885--4901, Online. Association for
  Computational Linguistics.

\bibitem[{Ouyang et~al.(2022)Ouyang, Wu, Jiang, Almeida, Wainwright, Mishkin,
  Zhang, Agarwal, Slama, Ray, Schulman, Hilton, Kelton, Miller, Simens, Askell,
  Welinder, Christiano, Leike, and Lowe}]{ouyang2022training}
Long Ouyang, Jeffrey Wu, Xu~Jiang, Diogo Almeida, Carroll~L. Wainwright, Pamela
  Mishkin, Chong Zhang, Sandhini Agarwal, Katarina Slama, Alex Ray, John
  Schulman, Jacob Hilton, Fraser Kelton, Luke Miller, Maddie Simens, Amanda
  Askell, Peter Welinder, Paul~F. Christiano, Jan Leike, and Ryan Lowe. 2022.
\newblock \href
  {http://papers.nips.cc/paper\_files/paper/2022/hash/b1efde53be364a73914f58805a001731-Abstract-Conference.html}
  {Training language models to follow instructions with human feedback}.
\newblock In \emph{Advances in Neural Information Processing Systems 35: Annual
  Conference on Neural Information Processing Systems 2022, NeurIPS 2022, New
  Orleans, LA, USA, November 28 - December 9, 2022}.

\bibitem[{Paetzold and Specia(2016)}]{paetzold-specia-2016-semeval}
Gustavo Paetzold and Lucia Specia. 2016.
\newblock \href {https://doi.org/10.18653/v1/S16-1085} {{S}em{E}val 2016 task
  11: Complex word identification}.
\newblock In \emph{Proceedings of the 10th International Workshop on Semantic
  Evaluation ({S}em{E}val-2016)}, pages 560--569, San Diego, California.
  Association for Computational Linguistics.

\bibitem[{Pandit and Hou(2021)}]{pandit-hou-2021-probing}
Onkar Pandit and Yufang Hou. 2021.
\newblock \href {https://doi.org/10.18653/v1/2021.naacl-main.327} {Probing for
  bridging inference in transformer language models}.
\newblock In \emph{Proceedings of the 2021 Conference of the North American
  Chapter of the Association for Computational Linguistics: Human Language
  Technologies}, pages 4153--4163, Online. Association for Computational
  Linguistics.

\bibitem[{Peng et~al.(2022)Peng, Wang, Hu, Jin, Hou, Li, Liu, and
  Liu}]{peng-etal-2022-copen}
Hao Peng, Xiaozhi Wang, Shengding Hu, Hailong Jin, Lei Hou, Juanzi Li, Zhiyuan
  Liu, and Qun Liu. 2022.
\newblock \href {https://doi.org/10.18653/v1/2022.emnlp-main.335} {{COPEN}:
  Probing conceptual knowledge in pre-trained language models}.
\newblock In \emph{Proceedings of the 2022 Conference on Empirical Methods in
  Natural Language Processing}, pages 5015--5035, Abu Dhabi, United Arab
  Emirates. Association for Computational Linguistics.

\bibitem[{Pennington et~al.(2014)Pennington, Socher, and
  Manning}]{pennington-etal-2014-glove}
Jeffrey Pennington, Richard Socher, and Christopher Manning. 2014.
\newblock \href {https://doi.org/10.3115/v1/D14-1162} {{G}lo{V}e: Global
  vectors for word representation}.
\newblock In \emph{Proceedings of the 2014 Conference on Empirical Methods in
  Natural Language Processing ({EMNLP})}, pages 1532--1543, Doha, Qatar.
  Association for Computational Linguistics.

\bibitem[{Perlitz et~al.(2023)Perlitz, Bandel, Gera, Arviv, Ein{-}Dor, Shnarch,
  Slonim, Shmueli{-}Scheuer, and Choshen}]{perlitz2023efficient}
Yotam Perlitz, Elron Bandel, Ariel Gera, Ofir Arviv, Liat Ein{-}Dor, Eyal
  Shnarch, Noam Slonim, Michal Shmueli{-}Scheuer, and Leshem Choshen. 2023.
\newblock \href {https://doi.org/10.48550/ARXIV.2308.11696} {Efficient
  benchmarking (of language models)}.
\newblock \emph{CoRR}, abs/2308.11696.

\bibitem[{Perlitz et~al.(2024)Perlitz, Gera, Arviv, Yehudai, Bandel, Shnarch,
  Shmueli-Scheuer, and Choshen}]{perlitz2024benchmarkagreementtestingright}
Yotam Perlitz, Ariel Gera, Ofir Arviv, Asaf Yehudai, Elron Bandel, Eyal
  Shnarch, Michal Shmueli-Scheuer, and Leshem Choshen. 2024.
\newblock \href {https://doi.org/10.48550/ARXIV.2407.13696} {Benchmark
  agreement testing done right: A guide for llm benchmark evaluation}.
\newblock \emph{CoRR}, abs/2407.13696.

\bibitem[{Petroni et~al.(2020)Petroni, Lewis, Piktus, Rockt{\"{a}}schel, Wu,
  Miller, and Riedel}]{petroni2020how}
Fabio Petroni, Patrick S.~H. Lewis, Aleksandra Piktus, Tim Rockt{\"{a}}schel,
  Yuxiang Wu, Alexander~H. Miller, and Sebastian Riedel. 2020.
\newblock \href {https://doi.org/10.24432/C5201W} {How context affects language
  models' factual predictions}.
\newblock In \emph{Conference on Automated Knowledge Base Construction, {AKBC}
  2020, Virtual, June 22-24, 2020}.

\bibitem[{Petroni et~al.(2019{\natexlab{a}})Petroni, Rockt{\"a}schel, Riedel,
  Lewis, Bakhtin, Wu, and Miller}]{petroni-etal-2019-language}
Fabio Petroni, Tim Rockt{\"a}schel, Sebastian Riedel, Patrick Lewis, Anton
  Bakhtin, Yuxiang Wu, and Alexander Miller. 2019{\natexlab{a}}.
\newblock \href {https://doi.org/10.18653/v1/D19-1250} {Language models as
  knowledge bases?}
\newblock In \emph{Proceedings of the 2019 Conference on Empirical Methods in
  Natural Language Processing and the 9th International Joint Conference on
  Natural Language Processing (EMNLP-IJCNLP)}, pages 2463--2473, Hong Kong,
  China. Association for Computational Linguistics.

\bibitem[{Petroni et~al.(2019{\natexlab{b}})Petroni, Rockt{\"{a}}schel, Riedel,
  Lewis, Bakhtin, Wu, and Miller}]{DBLP:conf/emnlp/PetroniRRLBWM19}
Fabio Petroni, Tim Rockt{\"{a}}schel, Sebastian Riedel, Patrick S.~H. Lewis,
  Anton Bakhtin, Yuxiang Wu, and Alexander~H. Miller. 2019{\natexlab{b}}.
\newblock \href {https://doi.org/10.18653/V1/D19-1250} {Language models as
  knowledge bases?}
\newblock In \emph{Proceedings of the 2019 Conference on Empirical Methods in
  Natural Language Processing and the 9th International Joint Conference on
  Natural Language Processing, {EMNLP-IJCNLP} 2019, Hong Kong, China, November
  3-7, 2019}, pages 2463--2473. Association for Computational Linguistics.

\bibitem[{Pimentel et~al.(2020)Pimentel, Valvoda, Maudslay, Zmigrod, Williams,
  and Cotterell}]{pimentel2020information}
Tiago Pimentel, Josef Valvoda, Rowan~Hall Maudslay, Ran Zmigrod, Adina
  Williams, and Ryan Cotterell. 2020.
\newblock \href {https://doi.org/10.18653/v1/2020.acl-main.420}
  {Information-theoretic probing for linguistic structure}.
\newblock In \emph{Proceedings of the 58th Annual Meeting of the Association
  for Computational Linguistics}, pages 4609--4622, Online. Association for
  Computational Linguistics.

\bibitem[{Radford et~al.(2019)Radford, Wu, Child, Luan, Amodei, Sutskever
  et~al.}]{radford_language_2019}
Alec Radford, Jeffrey Wu, Rewon Child, David Luan, Dario Amodei, Ilya
  Sutskever, et~al. 2019.
\newblock \href
  {https://www.semanticscholar.org/paper/Language-Models-are-Unsupervised-Multitask-Learners-Radford-Wu/9405cc0d6169988371b2755e573cc28650d14dfe}
  {Language models are unsupervised multitask learners}.
\newblock \emph{OpenAI blog}, 1(8):9.

\bibitem[{Raffel et~al.(2020)Raffel, Shazeer, Roberts, Lee, Narang, Matena,
  Zhou, Li, and Liu}]{raffel2020exploring}
Colin Raffel, Noam Shazeer, Adam Roberts, Katherine Lee, Sharan Narang, Michael
  Matena, Yanqi Zhou, Wei Li, and Peter~J. Liu. 2020.
\newblock \href {http://jmlr.org/papers/v21/20-074.html} {Exploring the limits
  of transfer learning with a unified text-to-text transformer}.
\newblock \emph{Journal of Machine Learning Research}, 21(140):1--67.

\bibitem[{Robins(2013)}]{robins2013short}
Robert~Henry Robins. 2013.
\newblock \href {https://doi.org/10.4324/9781315843186} {\emph{A short history
  of linguistics}}.
\newblock Routledge.

\bibitem[{Rodriguez et~al.(2021)Rodriguez, Barrow, Hoyle, Lalor, Jia, and
  Boyd-Graber}]{rodriguez-etal-2021-evaluation}
Pedro Rodriguez, Joe Barrow, Alexander~Miserlis Hoyle, John~P. Lalor, Robin
  Jia, and Jordan Boyd-Graber. 2021.
\newblock \href {https://doi.org/10.18653/v1/2021.acl-long.346} {Evaluation
  examples are not equally informative: How should that change {NLP}
  leaderboards?}
\newblock In \emph{Proceedings of the 59th Annual Meeting of the Association
  for Computational Linguistics and the 11th International Joint Conference on
  Natural Language Processing (Volume 1: Long Papers)}, pages 4486--4503,
  Online. Association for Computational Linguistics.

\bibitem[{Rogers et~al.(2020)Rogers, Kovaleva, and
  Rumshisky}]{rogers-etal-2020-primer}
Anna Rogers, Olga Kovaleva, and Anna Rumshisky. 2020.
\newblock \href {https://doi.org/10.1162/tacl_a_00349} {A primer in
  {BERT}ology: What we know about how {BERT} works}.
\newblock \emph{Transactions of the Association for Computational Linguistics},
  8:842--866.

\bibitem[{Rudinger et~al.(2018{\natexlab{a}})Rudinger, Teichert, Culkin, Zhang,
  and Van~Durme}]{rudinger-etal-2018-neural}
Rachel Rudinger, Adam Teichert, Ryan Culkin, Sheng Zhang, and Benjamin
  Van~Durme. 2018{\natexlab{a}}.
\newblock \href {https://doi.org/10.18653/v1/D18-1114} {Neural-{D}avidsonian
  semantic proto-role labeling}.
\newblock In \emph{Proceedings of the 2018 Conference on Empirical Methods in
  Natural Language Processing}, pages 944--955, Brussels, Belgium. Association
  for Computational Linguistics.

\bibitem[{Rudinger et~al.(2018{\natexlab{b}})Rudinger, White, and
  Van~Durme}]{rudinger-etal-2018-neural-models}
Rachel Rudinger, Aaron~Steven White, and Benjamin Van~Durme.
  2018{\natexlab{b}}.
\newblock \href {https://doi.org/10.18653/v1/N18-1067} {Neural models of
  factuality}.
\newblock In \emph{Proceedings of the 2018 Conference of the North {A}merican
  Chapter of the Association for Computational Linguistics: Human Language
  Technologies, Volume 1 (Long Papers)}, pages 731--744, New Orleans,
  Louisiana. Association for Computational Linguistics.

\bibitem[{de~Saussure(1916)}]{saussure1916course}
Ferdinand de~Saussure. 1916.
\newblock \emph{Cours de linguistique g{\'e}n{\'e}rale}.
\newblock Payot, Paris.

\bibitem[{Schaeffer et~al.(2023)Schaeffer, Miranda, and
  Koyejo}]{DBLP:conf/nips/SchaefferMK23}
Rylan Schaeffer, Brando Miranda, and Sanmi Koyejo. 2023.
\newblock \href
  {http://papers.nips.cc/paper\_files/paper/2023/hash/adc98a266f45005c403b8311ca7e8bd7-Abstract-Conference.html}
  {Are emergent abilities of large language models a mirage?}
\newblock In \emph{Advances in Neural Information Processing Systems 36: Annual
  Conference on Neural Information Processing Systems 2023, NeurIPS 2023, New
  Orleans, LA, USA, December 10 - 16, 2023}.

\bibitem[{Shapiro et~al.(2021)Shapiro, Paullada, and
  Steinert-Threlkeld}]{shapiro-etal-2021-multilabel-approach}
Naomi Shapiro, Amandalynne Paullada, and Shane Steinert-Threlkeld. 2021.
\newblock \href {https://doi.org/10.18653/v1/2021.findings-emnlp.382} {A
  multilabel approach to morphosyntactic probing}.
\newblock In \emph{Findings of the Association for Computational Linguistics:
  EMNLP 2021}, pages 4486--4524, Punta Cana, Dominican Republic. Association
  for Computational Linguistics.

\bibitem[{Sharma et~al.(2023)Sharma, Ash, and Misra}]{sharma2023truth}
Pratyusha Sharma, Jordan~T. Ash, and Dipendra Misra. 2023.
\newblock \href {https://doi.org/10.48550/ARXIV.2312.13558} {The truth is in
  there: Improving reasoning in language models with layer-selective rank
  reduction}.
\newblock \emph{CoRR}, abs/2312.13558.

\bibitem[{Shi et~al.(2016)Shi, Padhi, and Knight}]{shi-etal-2016-string}
Xing Shi, Inkit Padhi, and Kevin Knight. 2016.
\newblock \href {https://doi.org/10.18653/v1/D16-1159} {Does string-based
  neural {MT} learn source syntax?}
\newblock In \emph{Proceedings of the 2016 Conference on Empirical Methods in
  Natural Language Processing}, pages 1526--1534, Austin, Texas. Association
  for Computational Linguistics.

\bibitem[{Silveira et~al.(2014)Silveira, Dozat, de~Marneffe, Bowman, Connor,
  Bauer, and Manning}]{silveira-etal-2014-gold}
Natalia Silveira, Timothy Dozat, Marie-Catherine de~Marneffe, Samuel Bowman,
  Miriam Connor, John Bauer, and Chris Manning. 2014.
\newblock \href
  {http://www.lrec-conf.org/proceedings/lrec2014/pdf/1089_Paper.pdf} {A gold
  standard dependency corpus for {E}nglish}.
\newblock In \emph{Proceedings of the Ninth International Conference on
  Language Resources and Evaluation ({LREC}'14)}, pages 2897--2904, Reykjavik,
  Iceland. European Language Resources Association (ELRA).

\bibitem[{Socher et~al.(2013)Socher, Perelygin, Wu, Chuang, Manning, Ng, and
  Potts}]{socher-etal-2013-recursive}
Richard Socher, Alex Perelygin, Jean Wu, Jason Chuang, Christopher~D. Manning,
  Andrew Ng, and Christopher Potts. 2013.
\newblock \href {https://aclanthology.org/D13-1170} {Recursive deep models for
  semantic compositionality over a sentiment treebank}.
\newblock In \emph{Proceedings of the 2013 Conference on Empirical Methods in
  Natural Language Processing}, pages 1631--1642, Seattle, Washington, USA.
  Association for Computational Linguistics.

\bibitem[{Srivastava et~al.(2022)Srivastava, Rastogi, Rao, Shoeb, Abid, Fisch,
  Brown, Santoro, Gupta, Garriga{-}Alonso, Kluska, Lewkowycz, Agarwal, Power,
  Ray, Warstadt, Kocurek, Safaya, Tazarv, Xiang, Parrish, Nie, Hussain, Askell,
  Dsouza, Rahane, Iyer, Andreassen, Santilli, Stuhlm{\"{u}}ller, Dai, La,
  Lampinen, Zou, Jiang, Chen, Vuong, Gupta, Gottardi, Norelli, Venkatesh,
  Gholamidavoodi, Tabassum, Menezes, Kirubarajan, Mullokandov, Sabharwal,
  Herrick, Efrat, Erdem, Karakas, and et~al.}]{Srivastava2022BeyondTI}
Aarohi Srivastava, Abhinav Rastogi, Abhishek Rao, Abu Awal~Md Shoeb, Abubakar
  Abid, Adam Fisch, Adam~R. Brown, Adam Santoro, Aditya Gupta, Adri{\`{a}}
  Garriga{-}Alonso, Agnieszka Kluska, Aitor Lewkowycz, Akshat Agarwal, Alethea
  Power, Alex Ray, Alex Warstadt, Alexander~W. Kocurek, Ali Safaya, Ali Tazarv,
  Alice Xiang, Alicia Parrish, Allen Nie, Aman Hussain, Amanda Askell, Amanda
  Dsouza, Ameet Rahane, Anantharaman~S. Iyer, Anders Andreassen, Andrea
  Santilli, Andreas Stuhlm{\"{u}}ller, Andrew~M. Dai, Andrew La, Andrew~K.
  Lampinen, Andy Zou, Angela Jiang, Angelica Chen, Anh Vuong, Animesh Gupta,
  Anna Gottardi, Antonio Norelli, Anu Venkatesh, Arash Gholamidavoodi, Arfa
  Tabassum, Arul Menezes, Arun Kirubarajan, Asher Mullokandov, Ashish
  Sabharwal, Austin Herrick, Avia Efrat, Aykut Erdem, Ayla Karakas, and et~al.
  2022.
\newblock \href {https://doi.org/10.48550/ARXIV.2206.04615} {Beyond the
  imitation game: Quantifying and extrapolating the capabilities of language
  models}.
\newblock \emph{CoRR}, abs/2206.04615.

\bibitem[{Steen et~al.(2010)Steen, Dorst, Herrmann, Kaal, Krennmayr, Pasma
  et~al.}]{Steen2010AMF}
Gerard~J Steen, Aletta~G Dorst, J~Berenike Herrmann, Anna~A Kaal, Tina
  Krennmayr, Tryntje Pasma, et~al. 2010.
\newblock \href {https://doi.org/10.1075/celcr.14} {\emph{A method for
  linguistic metaphor identification}}.
\newblock Converging evidence in language and communication research. John
  Benjamins Publishing Company Amsterdam.

\bibitem[{Sudalairaj et~al.(2024)Sudalairaj, Bhandwaldar, Pareja, Xu, Cox, and
  Srivastava}]{Sudalairaj2024LABLA}
Shivchander Sudalairaj, Abhishek Bhandwaldar, Aldo Pareja, Kai Xu, David~D.
  Cox, and Akash Srivastava. 2024.
\newblock \href {https://doi.org/10.48550/ARXIV.2403.01081} {{LAB:} large-scale
  alignment for chatbots}.
\newblock \emph{CoRR}, abs/2403.01081.

\bibitem[{Szarvas et~al.(2008)Szarvas, Vincze, Farkas, and
  Csirik}]{vincze2008bioscope}
Gy{\"o}rgy Szarvas, Veronika Vincze, Rich{\'a}rd Farkas, and J{\'a}nos Csirik.
  2008.
\newblock \href {https://aclanthology.org/W08-0606} {The {B}io{S}cope corpus:
  annotation for negation, uncertainty and their scope in biomedical texts}.
\newblock In \emph{Proceedings of the Workshop on Current Trends in Biomedical
  Natural Language Processing}, pages 38--45, Columbus, Ohio. Association for
  Computational Linguistics.

\bibitem[{Talmor et~al.(2020{\natexlab{a}})Talmor, Elazar, Goldberg, and
  Berant}]{talmor-etal-2020-olmpics}
Alon Talmor, Yanai Elazar, Yoav Goldberg, and Jonathan Berant.
  2020{\natexlab{a}}.
\newblock \href {https://doi.org/10.1162/tacl_a_00342} {o{LM}pics-on what
  language model pre-training captures}.
\newblock \emph{Transactions of the Association for Computational Linguistics},
  8:743--758.

\bibitem[{Talmor et~al.(2020{\natexlab{b}})Talmor, Elazar, Goldberg, and
  Berant}]{10.1162/tacl_a_00342}
Alon Talmor, Yanai Elazar, Yoav Goldberg, and Jonathan Berant.
  2020{\natexlab{b}}.
\newblock \href {https://doi.org/10.1162/tacl_a_00342} {{oLMpics-On What
  Language Model Pre-training Captures}}.
\newblock \emph{Transactions of the Association for Computational Linguistics},
  8:743--758.

\bibitem[{Tay et~al.(2023)Tay, Dehghani, Tran, Garcia, Wei, Wang, Chung, Bahri,
  Schuster, Zheng, Zhou, Houlsby, and Metzler}]{tay2022ul2}
Yi~Tay, Mostafa Dehghani, Vinh~Q. Tran, Xavier Garcia, Jason Wei, Xuezhi Wang,
  Hyung~Won Chung, Dara Bahri, Tal Schuster, Huaixiu~Steven Zheng, Denny Zhou,
  Neil Houlsby, and Donald Metzler. 2023.
\newblock \href {https://openreview.net/pdf?id=6ruVLB727MC} {{UL2:} unifying
  language learning paradigms}.
\newblock In \emph{The Eleventh International Conference on Learning
  Representations, {ICLR} 2023, Kigali, Rwanda, May 1-5, 2023}. OpenReview.net.

\bibitem[{Tenney et~al.(2019{\natexlab{a}})Tenney, Das, and
  Pavlick}]{tenney-etal-2019-bert}
Ian Tenney, Dipanjan Das, and Ellie Pavlick. 2019{\natexlab{a}}.
\newblock \href {https://doi.org/10.18653/v1/P19-1452} {{BERT} rediscovers the
  classical {NLP} pipeline}.
\newblock In \emph{Proceedings of the 57th Annual Meeting of the Association
  for Computational Linguistics}, pages 4593--4601, Florence, Italy.
  Association for Computational Linguistics.

\bibitem[{Tenney et~al.(2019{\natexlab{b}})Tenney, Xia, Chen, Wang, Poliak,
  McCoy, Kim, Durme, Bowman, Das, and Pavlick}]{tenney2018what}
Ian Tenney, Patrick Xia, Berlin Chen, Alex Wang, Adam Poliak, R.~Thomas McCoy,
  Najoung Kim, Benjamin~Van Durme, Samuel~R. Bowman, Dipanjan Das, and Ellie
  Pavlick. 2019{\natexlab{b}}.
\newblock \href {https://openreview.net/forum?id=SJzSgnRcKX} {What do you learn
  from context? probing for sentence structure in contextualized word
  representations}.
\newblock In \emph{7th International Conference on Learning Representations,
  {ICLR} 2019, New Orleans, LA, USA, May 6-9, 2019}. OpenReview.net.

\bibitem[{Thakur et~al.(2021)Thakur, Reimers, R{\"{u}}ckl{\'{e}}, Srivastava,
  and Gurevych}]{thakurbeir}
Nandan Thakur, Nils Reimers, Andreas R{\"{u}}ckl{\'{e}}, Abhishek Srivastava,
  and Iryna Gurevych. 2021.
\newblock \href
  {https://datasets-benchmarks-proceedings.neurips.cc/paper/2021/hash/65b9eea6e1cc6bb9f0cd2a47751a186f-Abstract-round2.html}
  {{BEIR:} {A} heterogeneous benchmark for zero-shot evaluation of information
  retrieval models}.
\newblock In \emph{Proceedings of the Neural Information Processing Systems
  Track on Datasets and Benchmarks 1, NeurIPS Datasets and Benchmarks 2021,
  December 2021, virtual}.

\bibitem[{Torroba~Hennigen et~al.(2020)Torroba~Hennigen, Williams, and
  Cotterell}]{torroba-hennigen-etal-2020-intrinsic}
Lucas Torroba~Hennigen, Adina Williams, and Ryan Cotterell. 2020.
\newblock \href {https://doi.org/10.18653/v1/2020.emnlp-main.15} {Intrinsic
  probing through dimension selection}.
\newblock In \emph{Proceedings of the 2020 Conference on Empirical Methods in
  Natural Language Processing (EMNLP)}, pages 197--216, Online. Association for
  Computational Linguistics.

\bibitem[{Touvron et~al.(2023)Touvron, Martin, Stone, Albert, Almahairi,
  Babaei, Bashlykov, Batra, Bhargava, Bhosale, Bikel, Blecher, Canton{-}Ferrer,
  Chen, Cucurull, Esiobu, Fernandes, Fu, Fu, Fuller, Gao, Goswami, Goyal,
  Hartshorn, Hosseini, Hou, Inan, Kardas, Kerkez, Khabsa, Kloumann, Korenev,
  Koura, Lachaux, Lavril, Lee, Liskovich, Lu, Mao, Martinet, Mihaylov, Mishra,
  Molybog, Nie, Poulton, Reizenstein, Rungta, Saladi, Schelten, Silva, Smith,
  Subramanian, Tan, Tang, Taylor, Williams, Kuan, Xu, Yan, Zarov, Zhang, Fan,
  Kambadur, Narang, Rodriguez, Stojnic, Edunov, and Scialom}]{touvron2023llama}
Hugo Touvron, Louis Martin, Kevin Stone, Peter Albert, Amjad Almahairi, Yasmine
  Babaei, Nikolay Bashlykov, Soumya Batra, Prajjwal Bhargava, Shruti Bhosale,
  Dan Bikel, Lukas Blecher, Cristian Canton{-}Ferrer, Moya Chen, Guillem
  Cucurull, David Esiobu, Jude Fernandes, Jeremy Fu, Wenyin Fu, Brian Fuller,
  Cynthia Gao, Vedanuj Goswami, Naman Goyal, Anthony Hartshorn, Saghar
  Hosseini, Rui Hou, Hakan Inan, Marcin Kardas, Viktor Kerkez, Madian Khabsa,
  Isabel Kloumann, Artem Korenev, Punit~Singh Koura, Marie{-}Anne Lachaux,
  Thibaut Lavril, Jenya Lee, Diana Liskovich, Yinghai Lu, Yuning Mao, Xavier
  Martinet, Todor Mihaylov, Pushkar Mishra, Igor Molybog, Yixin Nie, Andrew
  Poulton, Jeremy Reizenstein, Rashi Rungta, Kalyan Saladi, Alan Schelten, Ruan
  Silva, Eric~Michael Smith, Ranjan Subramanian, Xiaoqing~Ellen Tan, Binh Tang,
  Ross Taylor, Adina Williams, Jian~Xiang Kuan, Puxin Xu, Zheng Yan, Iliyan
  Zarov, Yuchen Zhang, Angela Fan, Melanie Kambadur, Sharan Narang,
  Aur{\'{e}}lien Rodriguez, Robert Stojnic, Sergey Edunov, and Thomas Scialom.
  2023.
\newblock \href {https://doi.org/10.48550/ARXIV.2307.09288} {Llama 2: Open
  foundation and fine-tuned chat models}.
\newblock \emph{CoRR}, abs/2307.09288.

\bibitem[{Vahtola et~al.(2022)Vahtola, Creutz, and
  Tiedemann}]{vahtola-etal-2022-easy}
Teemu Vahtola, Mathias Creutz, and J{\"o}rg Tiedemann. 2022.
\newblock \href {https://doi.org/10.18653/v1/2022.blackboxnlp-1.20} {It is not
  easy to detect paraphrases: Analysing semantic similarity with antonyms and
  negation using the new {S}em{A}nto{N}eg benchmark}.
\newblock In \emph{Proceedings of the Fifth BlackboxNLP Workshop on Analyzing
  and Interpreting Neural Networks for NLP}, pages 249--262, Abu Dhabi, United
  Arab Emirates (Hybrid). Association for Computational Linguistics.

\bibitem[{Vashishtha et~al.(2019)Vashishtha, Van~Durme, and
  White}]{vashishtha-etal-2019-fine}
Siddharth Vashishtha, Benjamin Van~Durme, and Aaron~Steven White. 2019.
\newblock \href {https://doi.org/10.18653/v1/P19-1280} {Fine-grained temporal
  relation extraction}.
\newblock In \emph{Proceedings of the 57th Annual Meeting of the Association
  for Computational Linguistics}, pages 2906--2919, Florence, Italy.
  Association for Computational Linguistics.

\bibitem[{Veldhoen et~al.(2016)Veldhoen, Hupkes, and
  Zuidema}]{Veldhoen2016DiagnosticCR}
Sara Veldhoen, Dieuwke Hupkes, and Willem~H. Zuidema. 2016.
\newblock \href {https://ceur-ws.org/Vol-1773/CoCoNIPS\_2016\_paper6.pdf}
  {Diagnostic classifiers revealing how neural networks process hierarchical
  structure}.
\newblock In \emph{Proceedings of the Workshop on Cognitive Computation:
  Integrating neural and symbolic approaches 2016 co-located with the 30th
  Annual Conference on Neural Information Processing Systems {(NIPS} 2016),
  Barcelona, Spain, December 9, 2016}, volume 1773 of \emph{{CEUR} Workshop
  Proceedings}. CEUR-WS.org.

\bibitem[{Voita and Titov(2020)}]{voita-titov-2020-information}
Elena Voita and Ivan Titov. 2020.
\newblock \href {https://doi.org/10.18653/v1/2020.emnlp-main.14}
  {Information-theoretic probing with minimum description length}.
\newblock In \emph{Proceedings of the 2020 Conference on Empirical Methods in
  Natural Language Processing (EMNLP)}, pages 183--196, Online. Association for
  Computational Linguistics.

\bibitem[{Waldis et~al.(2024{\natexlab{a}})Waldis, Hou, and
  Gurevych}]{waldis2024dive}
Andreas Waldis, Yufang Hou, and Iryna Gurevych. 2024{\natexlab{a}}.
\newblock \href {https://aclanthology.org/2024.findings-eacl.146} {Dive into
  the chasm: Probing the gap between in- and cross-topic generalization}.
\newblock In \emph{Findings of the Association for Computational Linguistics:
  EACL 2024}, pages 2197--2214, St. Julian{'}s, Malta. Association for
  Computational Linguistics.

\bibitem[{Waldis et~al.(2024{\natexlab{b}})Waldis, Hou, and
  Gurevych}]{waldis2024handle}
Andreas Waldis, Yufang Hou, and Iryna Gurevych. 2024{\natexlab{b}}.
\newblock \href {https://doi.org/10.48550/ARXIV.2309.08316} {How to handle
  different types of out-of-distribution scenarios in computational
  argumentation? a comprehensive and fine-grained field study}.
\newblock \emph{CoRR}, abs/2309.08316.

\bibitem[{Wang et~al.(2019{\natexlab{a}})Wang, Pruksachatkun, Nangia, Singh,
  Michael, Hill, Levy, and Bowman}]{NEURIPS2019_4496bf24}
Alex Wang, Yada Pruksachatkun, Nikita Nangia, Amanpreet Singh, Julian Michael,
  Felix Hill, Omer Levy, and Samuel Bowman. 2019{\natexlab{a}}.
\newblock \href
  {https://proceedings.neurips.cc/paper/2019/file/4496bf24afe7fab6f046bf4923da8de6-Paper.pdf}
  {Superglue: A stickier benchmark for general-purpose language understanding
  systems}.
\newblock In \emph{Advances in Neural Information Processing Systems},
  volume~32. Curran Associates, Inc.

\bibitem[{Wang et~al.(2019{\natexlab{b}})Wang, Singh, Michael, Hill, Levy, and
  Bowman}]{DBLP:conf/iclr/WangSMHLB19}
Alex Wang, Amanpreet Singh, Julian Michael, Felix Hill, Omer Levy, and
  Samuel~R. Bowman. 2019{\natexlab{b}}.
\newblock \href {https://openreview.net/forum?id=rJ4km2R5t7} {{GLUE:} {A}
  multi-task benchmark and analysis platform for natural language
  understanding}.
\newblock In \emph{7th International Conference on Learning Representations,
  {ICLR} 2019, New Orleans, LA, USA, May 6-9, 2019}. OpenReview.net.

\bibitem[{Wang et~al.(2021)Wang, Xu, Wang, Gan, Cheng, Gao, Awadallah, and
  Li}]{wang2021adversarial}
Boxin Wang, Chejian Xu, Shuohang Wang, Zhe Gan, Yu~Cheng, Jianfeng Gao,
  Ahmed~Hassan Awadallah, and Bo~Li. 2021.
\newblock \href
  {https://datasets-benchmarks-proceedings.neurips.cc/paper/2021/hash/335f5352088d7d9bf74191e006d8e24c-Abstract-round2.html}
  {Adversarial {GLUE:} {A} multi-task benchmark for robustness evaluation of
  language models}.
\newblock In \emph{Proceedings of the Neural Information Processing Systems
  Track on Datasets and Benchmarks 1, NeurIPS Datasets and Benchmarks 2021,
  December 2021, virtual}.

\bibitem[{Wang et~al.(2023)Wang, Ivison, Dasigi, Hessel, Khot, Chandu, Wadden,
  MacMillan, Smith, Beltagy, and Hajishirzi}]{wang2023far}
Yizhong Wang, Hamish Ivison, Pradeep Dasigi, Jack Hessel, Tushar Khot, Khyathi
  Chandu, David Wadden, Kelsey MacMillan, Noah~A. Smith, Iz~Beltagy, and
  Hannaneh Hajishirzi. 2023.
\newblock \href
  {http://papers.nips.cc/paper\_files/paper/2023/hash/ec6413875e4ab08d7bc4d8e225263398-Abstract-Datasets\_and\_Benchmarks.html}
  {How far can camels go? exploring the state of instruction tuning on open
  resources}.
\newblock In \emph{Advances in Neural Information Processing Systems 36: Annual
  Conference on Neural Information Processing Systems 2023, NeurIPS 2023, New
  Orleans, LA, USA, December 10 - 16, 2023}.

\bibitem[{Wang et~al.(2022)Wang, Mishra, Alipoormolabashi, Kordi, Mirzaei,
  Naik, Ashok, Dhanasekaran, Arunkumar, Stap, Pathak, Karamanolakis, Lai,
  Purohit, Mondal, Anderson, Kuznia, Doshi, Pal, Patel, Moradshahi, Parmar,
  Purohit, Varshney, Kaza, Verma, Puri, Karia, Doshi, Sampat, Mishra, Reddy~A,
  Patro, Dixit, and Shen}]{wang-etal-2022-super}
Yizhong Wang, Swaroop Mishra, Pegah Alipoormolabashi, Yeganeh Kordi, Amirreza
  Mirzaei, Atharva Naik, Arjun Ashok, Arut~Selvan Dhanasekaran, Anjana
  Arunkumar, David Stap, Eshaan Pathak, Giannis Karamanolakis, Haizhi Lai,
  Ishan Purohit, Ishani Mondal, Jacob Anderson, Kirby Kuznia, Krima Doshi,
  Kuntal~Kumar Pal, Maitreya Patel, Mehrad Moradshahi, Mihir Parmar, Mirali
  Purohit, Neeraj Varshney, Phani~Rohitha Kaza, Pulkit Verma, Ravsehaj~Singh
  Puri, Rushang Karia, Savan Doshi, Shailaja~Keyur Sampat, Siddhartha Mishra,
  Sujan Reddy~A, Sumanta Patro, Tanay Dixit, and Xudong Shen. 2022.
\newblock \href {https://doi.org/10.18653/v1/2022.emnlp-main.340}
  {Super-{N}atural{I}nstructions: Generalization via declarative instructions
  on 1600+ {NLP} tasks}.
\newblock In \emph{Proceedings of the 2022 Conference on Empirical Methods in
  Natural Language Processing}, pages 5085--5109, Abu Dhabi, United Arab
  Emirates. Association for Computational Linguistics.

\bibitem[{Warstadt et~al.(2020)Warstadt, Parrish, Liu, Mohananey, Peng, Wang,
  and Bowman}]{warstadt-etal-2020-blimp-benchmark}
Alex Warstadt, Alicia Parrish, Haokun Liu, Anhad Mohananey, Wei Peng, Sheng-Fu
  Wang, and Samuel~R. Bowman. 2020.
\newblock \href {https://doi.org/10.1162/tacl_a_00321} {{BL}i{MP}: The
  benchmark of linguistic minimal pairs for {E}nglish}.
\newblock \emph{Transactions of the Association for Computational Linguistics},
  8:377--392.

\bibitem[{Webber et~al.(2019)Webber, Prasad, Lee, and Joshi}]{webber2019penn}
Bonnie Webber, Rashmi Prasad, Alan Lee, and Aravind Joshi. 2019.
\newblock \href {https://doi.org/10.35111/qebf-gk47} {The penn discourse
  treebank 3.0 annotation manual}.
\newblock \emph{Philadelphia, University of Pennsylvania}.

\bibitem[{Weischedel et~al.(2013)Weischedel, Palmer, Marcus, Hovy, Pradhan,
  Ramshaw, Xue, Taylor, Kaufman, Franchini et~al.}]{weischedel2013ontonotes}
Ralph Weischedel, Martha Palmer, Mitchell Marcus, Eduard Hovy, Sameer Pradhan,
  Lance Ramshaw, Nianwen Xue, Ann Taylor, Jeff Kaufman, Michelle Franchini,
  et~al. 2013.
\newblock \href {https://doi.org/10.35111/xmhb-2b84} {Ontonotes release 5.0}.
\newblock \emph{Linguistic Data Consortium, Philadelphia, PA}, 23:170.

\bibitem[{White et~al.(2016)White, Reisinger, Sakaguchi, Vieira, Zhang,
  Rudinger, Rawlins, and Van~Durme}]{white-etal-2016-universal}
Aaron~Steven White, Drew Reisinger, Keisuke Sakaguchi, Tim Vieira, Sheng Zhang,
  Rachel Rudinger, Kyle Rawlins, and Benjamin Van~Durme. 2016.
\newblock \href {https://doi.org/10.18653/v1/D16-1177} {Universal
  decompositional semantics on {U}niversal {D}ependencies}.
\newblock In \emph{Proceedings of the 2016 Conference on Empirical Methods in
  Natural Language Processing}, pages 1713--1723, Austin, Texas. Association
  for Computational Linguistics.

\bibitem[{Wu et~al.(2020)Wu, Chen, Kao, and Liu}]{wu-etal-2020-perturbed}
Zhiyong Wu, Yun Chen, Ben Kao, and Qun Liu. 2020.
\newblock \href {https://doi.org/10.18653/v1/2020.acl-main.383} {Perturbed
  masking: Parameter-free probing for analyzing and interpreting {BERT}}.
\newblock In \emph{Proceedings of the 58th Annual Meeting of the Association
  for Computational Linguistics}, pages 4166--4176, Online. Association for
  Computational Linguistics.

\bibitem[{Xu et~al.(2023)Xu, Sun, Zheng, Geng, Zhao, Feng, Tao, and
  Jiang}]{DBLP:journals/corr/abs-2304-12244}
Can Xu, Qingfeng Sun, Kai Zheng, Xiubo Geng, Pu~Zhao, Jiazhan Feng, Chongyang
  Tao, and Daxin Jiang. 2023.
\newblock \href {https://doi.org/10.48550/ARXIV.2304.12244} {Wizardlm:
  Empowering large language models to follow complex instructions}.
\newblock \emph{CoRR}, abs/2304.12244.

\bibitem[{Yadav et~al.(2023)Yadav, Choshen, Raffel, and
  Bansal}]{Yadav2023ComPEFTCF}
Prateek Yadav, Leshem Choshen, Colin Raffel, and Mohit Bansal. 2023.
\newblock \href {https://doi.org/10.48550/ARXIV.2311.13171} {Compeft:
  Compression for communicating parameter efficient updates via sparsification
  and quantization}.
\newblock \emph{CoRR}, abs/2311.13171.

\bibitem[{Yang et~al.(2023)Yang, Zhang, Qin, Li, Wang, Liu, Wang, Xie, and
  Zhang}]{yang-etal-2023-glue}
Linyi Yang, Shuibai Zhang, Libo Qin, Yafu Li, Yidong Wang, Hanmeng Liu, Jindong
  Wang, Xing Xie, and Yue Zhang. 2023.
\newblock \href {https://doi.org/10.18653/v1/2023.findings-acl.806}
  {{GLUE}-{X}: Evaluating natural language understanding models from an
  out-of-distribution generalization perspective}.
\newblock In \emph{Findings of the Association for Computational Linguistics:
  ACL 2023}, pages 12731--12750, Toronto, Canada. Association for Computational
  Linguistics.

\bibitem[{Yuan et~al.(2024)Yuan, Whitehouse, Chamoun, Aly, and
  Vlachos}]{yuan2024probelmplausibilityrankingevaluation}
Zhangdie Yuan, Chenxi Whitehouse, Eric Chamoun, Rami Aly, and Andreas Vlachos.
  2024.
\newblock \href {https://doi.org/10.48550/ARXIV.2404.03818} {Probelm:
  Plausibility ranking evaluation for language models}.
\newblock \emph{CoRR}, abs/2404.03818.

\bibitem[{Zeldes(2017)}]{Zeldes2017}
Amir Zeldes. 2017.
\newblock \href {https://doi.org/http://dx.doi.org/10.1007/s10579-016-9343-x}
  {The {GUM} corpus: Creating multilayer resources in the classroom}.
\newblock \emph{Language Resources and Evaluation}, 51(3):581--612.

\bibitem[{Zhang et~al.(2020)Zhang, Ramachandran, Tenney, Elazar, and
  Roth}]{zhang-etal-2020-language}
Xikun Zhang, Deepak Ramachandran, Ian Tenney, Yanai Elazar, and Dan Roth. 2020.
\newblock \href {https://doi.org/10.18653/v1/2020.blackboxnlp-1.27} {Do
  language embeddings capture scales?}
\newblock In \emph{Proceedings of the Third BlackboxNLP Workshop on Analyzing
  and Interpreting Neural Networks for NLP}, pages 292--299, Online.
  Association for Computational Linguistics.

\bibitem[{Zhang et~al.(2021)Zhang, Warstadt, Li, and
  Bowman}]{zhang-etal-2021-need}
Yian Zhang, Alex Warstadt, Xiaocheng Li, and Samuel~R. Bowman. 2021.
\newblock \href {https://doi.org/10.18653/v1/2021.acl-long.90} {When do you
  need billions of words of pretraining data?}
\newblock In \emph{Proceedings of the 59th Annual Meeting of the Association
  for Computational Linguistics and the 11th International Joint Conference on
  Natural Language Processing (Volume 1: Long Papers)}, pages 1112--1125,
  Online. Association for Computational Linguistics.

\bibitem[{Zheng et~al.(2023)Zheng, Chiang, Sheng, Zhuang, Wu, Zhuang, Lin, Li,
  Li, Xing, Zhang, Gonzalez, and Stoica}]{heng2023judging}
Lianmin Zheng, Wei{-}Lin Chiang, Ying Sheng, Siyuan Zhuang, Zhanghao Wu,
  Yonghao Zhuang, Zi~Lin, Zhuohan Li, Dacheng Li, Eric~P. Xing, Hao Zhang,
  Joseph~E. Gonzalez, and Ion Stoica. 2023.
\newblock \href
  {http://papers.nips.cc/paper\_files/paper/2023/hash/91f18a1287b398d378ef22505bf41832-Abstract-Datasets\_and\_Benchmarks.html}
  {Judging llm-as-a-judge with mt-bench and chatbot arena}.
\newblock In \emph{Advances in Neural Information Processing Systems 36: Annual
  Conference on Neural Information Processing Systems 2023, NeurIPS 2023, New
  Orleans, LA, USA, December 10 - 16, 2023}.

\bibitem[{Zhou et~al.(2023)Zhou, Liu, Xu, Iyer, Sun, Mao, Ma, Efrat, Yu, Yu,
  Zhang, Ghosh, Lewis, Zettlemoyer, and Levy}]{zhou2024lima}
Chunting Zhou, Pengfei Liu, Puxin Xu, Srinivasan Iyer, Jiao Sun, Yuning Mao,
  Xuezhe Ma, Avia Efrat, Ping Yu, Lili Yu, Susan Zhang, Gargi Ghosh, Mike
  Lewis, Luke Zettlemoyer, and Omer Levy. 2023.
\newblock \href
  {http://papers.nips.cc/paper\_files/paper/2023/hash/ac662d74829e4407ce1d126477f4a03a-Abstract-Conference.html}
  {{LIMA:} less is more for alignment}.
\newblock In \emph{Advances in Neural Information Processing Systems 36: Annual
  Conference on Neural Information Processing Systems 2023, NeurIPS 2023, New
  Orleans, LA, USA, December 10 - 16, 2023}.

\bibitem[{Zhu et~al.(2022{\natexlab{a}})Zhu, Shahtalebi, and
  Rudzicz}]{zhu-etal-2022-predicting}
Zining Zhu, Soroosh Shahtalebi, and Frank Rudzicz. 2022{\natexlab{a}}.
\newblock \href {https://doi.org/10.18653/v1/2022.emnlp-main.793} {Predicting
  fine-tuning performance with probing}.
\newblock In \emph{Proceedings of the 2022 Conference on Empirical Methods in
  Natural Language Processing}, pages 11534--11547, Abu Dhabi, United Arab
  Emirates. Association for Computational Linguistics.

\bibitem[{Zhu et~al.(2022{\natexlab{b}})Zhu, Wang, Li, and
  Rudzicz}]{zhu-etal-2022-data}
Zining Zhu, Jixuan Wang, Bai Li, and Frank Rudzicz. 2022{\natexlab{b}}.
\newblock \href {https://doi.org/10.18653/v1/2022.findings-acl.326} {On the
  data requirements of probing}.
\newblock In \emph{Findings of the Association for Computational Linguistics:
  ACL 2022}, pages 4132--4147, Dublin, Ireland. Association for Computational
  Linguistics.

\end{thebibliography}
\bibliographystyle{acl_natbib}

\clearpage
\appendix
\section{Additional Details of \Holmes}

\subsection{Additional Details on the Evolution of Probing Literature}\label{app:evolution}

We analyze publication trends by year and venue as shown in \autoref{tab:year_venue}.
Less work was published between 2015-2018 (\textit{earlier}) focusing on LSTM-based \citep{linzen-etal-2016-assessing,conneau-etal-2018-cram} and static LMs \citep{kohn-2015-whats,linzen-etal-2016-assessing,belinkov-etal-2017-neural,conneau-etal-2018-cram}.
With the release of BERT \citep{devlin-etal-2019-bert} in 2019, we note increasing attention to analyzing linguistic abilities within LMs, with a peak of 90 papers in 2022.
Considering the venue, more than half of the relevant work (149 papers) was published at major conferences (ACL and EMNLP), and 68 papers were published at AACL, EACL, NAACL, and COLING.\footnote{Note that EMNLP-23 and AACL-23 proceedings were not published when conducting this meta-study.} 
In addition, we observe a constant contribution of TACL, various workshops, such as \href{https://blackboxnlp.github.io/}{Analyzing and Interpreting Neural Networks for NLP} or \href{https://www.sigrep.org/}{Representation Learning for NLP}. 

\subsection{Experimental Details}\label{app:experiments}
\paragraph{Probing Hyperparameters}\label{app:hyperparameters}
Following previous work \citep{hewitt-liang-2019-designing, voita-titov-2020-information}, we use fixed hyperparameters for training the probes: 20 epochs, where we find the best one using dev instances; AdamW \citep{adamW2019} as optimizer; a batch size of 64; a learning rate of 0.0005; a dropout rate of 0.2; a warmup rate of 10\% of the steps; random seeds: $[0,1,2,3,4]$

\paragraph{Hardware}\label{app:hardware}
We run all of our experiments using 12 Nvidia RTX A6000 GPUs.
Every GPU provides 48GB of memory and 10752 CUDA Cores.

\paragraph{Considered LMs}\label{app:lms}

\autoref{tab:models} outlines the details of the LMs we evaluate on \Holmes{} in this work.

\subsection{Probing Datasets Categorization}\label{app:task-categorization}

We show in \autoref{tab:scenarios-morphology}, \autoref{tab:scenarios-syntax}, \autoref{tab:scenarios-semantics}, \autoref{tab:scenarios-reasoning}, and \autoref{tab:scenarios-discourse} which resources \Holmes{} use to cover \textit{morphology}, \textit{syntax}, \textit{semantics}, \textit{reasoning}, and \textit{discourse}  phenomena.
Further, we provide illustrative examples of the phenomena. 
We rely on 33 works providing the data, the specific linguistic phenomena, or both.
For example, for \textit{readability}, we use the data of \citet{weischedel2013ontonotes} and calculated the Flesch score \citep{flesch1948new}.

\paragraph{Morphology}\label{subsec:morphology}
First, we feature 19 tasks verifying \textit{morphology} phenomena:
\textit{Anaphor agreement} , \textit{determiner noun agreement}, \textit{subject-verb agreement} and \textit{irregular forms} \citep{warstadt-etal-2020-blimp-benchmark,huebner-etal-2021-babyberta}.

\paragraph{Syntax}\label{subsec:syntax}
The second group of 75 tasks verifies the following \textit{syntax} phenomena:
\textit{Part-of-speech} and \textit{constituent labeling} \citep{weischedel2013ontonotes}; \textit{dependency labeling} \citep{silveira-etal-2014-gold}; \textit{bigram-shift} (whether two words were shifted), \textit{tree-depth} (the depth of a sentence constituency tree), \textit{top-constituent-task} (top constituency tag), and \textit{sentence-length} \citep{conneau-etal-2018-cram}; \textit{subject-} \& \textit{object-number} (singular/plural), and \textit{deoncausative-inchoative alternation} (interaction of a verb with its context) based on \citet{klafka-ettinger-2020-spying}; \textit{binding}, \textit{control/raising}, \textit{negative polarity item licensing}, \textit{island-effects}, \textit{argument-structure}, \textit{ellipsis}, and \textit{filler-gap} \citep{warstadt-etal-2020-blimp-benchmark,huebner-etal-2021-babyberta}.

\begin{table}[t]
\centering
  \setlength{\tabcolsep}{3pt}
  \resizebox{0.46\textwidth}{!}{%
    \begin{tabular}{l|cccccc|c}
    \toprule
       & \it earlier & \it 2019 & \it 2020 & \it 2021 & \it 2022 & \it 2023 & Total \\
    \midrule
      \texttt{ACL} & 2 & 10 & 12 & 9 & 34 & 25 & 92  \\
      \texttt{AACL} & - & - & - & - & 1 & - & 1  \\
      \texttt{COLING} & - & - & 10 & - & 9 & - & 19  \\
      \texttt{EACL} & - & - & - & 7 & - & 15 & 22  \\
      \texttt{EMNLP} & 2 & 4 & 13 & 17 & 21 & - & 57  \\
      \texttt{NAACL} & - & 3 & - & 9 & 14 & - & 26  \\
      \texttt{TACL} & 1 & 1 & 2 & 3 & 3 & 1 & 11  \\
      \texttt{Workshops} & 4 & 4 & 10 & 10 & 7 & 1 & 36  \\
      \texttt{Other} & 1 & 2 & 1 & 1 & 1 & 4 & 10  \\
    \midrule
      \texttt{Probing}  & 10 & 24 & 48 & 56 & 90 & 46 & 274  \\
      \texttt{All Papers}  & 8,056 & 3,111 & 3,822 & 4,294 & 5,133 & 3,647 & 28,063  \\
    \midrule
    \bottomrule
    \end{tabular}
  }
  \caption{Evolution of probing studies. Note that EMNLP-23 and AACL-23 proceedings were not published when conducting this meta-study.}
  \label{tab:year_venue}
\end{table}

\paragraph{Semantics}\label{subsec:semantic}
Third, consider 67 datasets covering \textit{semantics} phenomena:
\textit{Named-entity labeling} and \textit{semantic-role labeling} \citep{weischedel2013ontonotes}; \textit{tense}, \textit{semantic odd man out}, \textit{word content}, and \textit{coordination inversion} \citep{conneau-etal-2018-cram}; \textit{semantic relation classification} \citep{hendrickx-etal-2010-semeval}; \textit{semantic proto-roles} \citep{rudinger-etal-2018-neural}; \textit{factuality} (if a span is factual or not) \citep{rudinger-etal-2018-neural-models}; \textit{genericity} (whether a span is generic or not) \citep{govindarajan-etal-2019-decomposing}; \textit{event structure} \citep{gantt-etal-2022-decomposing}; \textit{time} (time dimension of a span) \citep{vashishtha-etal-2019-fine}; \textit{word sense} \citep{white-etal-2016-universal}; \textit{sentiment analysis} \citep{socher-etal-2013-recursive}; \textit{object-} and \textit{subject-animacy} (whether a entity is animate, like human, or not, such as cars), \textit{object-} and \textit{subject-gender} (male/female), \textit{verb-tense}, and \textit{verb-dynamic} \citet{klafka-ettinger-2020-spying}; \textit{metaphor} \citep{mohler-etal-2016-introducing, Birke2006ACA, Steen2010AMF}; \textit{complex word identification} (whether the word is complex or not) \citep{paetzold-specia-2016-semeval}; and \textit{passive} \citep{krasnowska-kieras-wroblewska-2019-empirical}.
In addition, we derive an \textit{synonym-/antonym-detection} dataset using WordNet \citep{wordnet-1995} and the texts from OntoNotes v5 \citet{weischedel2013ontonotes}.

\paragraph{Reasoning}\label{subsec:reasoning}
Forth, 19 datasets cover \textit{reasoning} phenomena:
\textit{Paraphrasticity} with negation and antonyms \citep{vahtola-etal-2022-easy}; \textit{negation detection} \citep{vincze2008bioscope, konstantinova-etal-2012-review, morante-blanco-2012-sem}; \textit{negation-span classification} (does a span cause a negation) \citep{vincze2008bioscope, konstantinova-etal-2012-review}; \textit{negation-correspondence} (the target span of a negation) \citep{vincze2008bioscope, konstantinova-etal-2012-review}; \textit{speculation detection}, \textit{speculation-span classification}, and \textit{speculation-correspondence} (the target span of a sepculation) \citep{vincze2008bioscope}; and \textit{always-never}, \textit{age comparison}, \textit{objects comparison}, \textit{antonym negation}, \textit{property conjunction}, \textit{taxonomy connection}, and \textit{multi-hop composition} \citep{10.1162/tacl_a_00342}.

\paragraph{Discourse}\label{subsec:discourse}
Finally, \Holmes{} embodies 28 datasets addressing \textit{discourse} phenomena:
\textit{Co-reference resolution} \citet{weischedel2013ontonotes}; \textit{bridging} \citep{hou-2018-enhanced,hou-2020-bridging,pandit-hou-2021-probing}; \textit{discourse connective} \citep{nie-etal-2019-dissent}; \textit{sentence order} and \textit{next-sentence prediction} \citep{narayan-etal-2018-dont};
Given discourse tree, whether two nodes correspond (\textit{discourse correspondence}), the correct order of two nodes (\textit{discourse order}), node-node relation (\textit{discourse relation}), distance between two nodes (\textit{discourse distance}), explicit node class \textit{discourse explicit classes}, implicit node class \textit{discourse implicit classes} \citep{webber2019penn, kurfali-ostling-2021-probing}; and given a rhetorical tree with the number of child nodes (\textit{rst-count}), the node depth (\textit{rst-depth}), distance between two nodes \textit{rst-distance}, node-node relation (\textit{rst-relation}), node-node relation group (\textit{rst-relation-group}), appear two nodes after each other (\textit{rst-successively}), node type (\textit{rst-type}) \citep{carlson-etal-2001, koto-etal-2021-discourse,kurfali-ostling-2021-probing,Zeldes2017}.

\subsection{Details of Probing Dataset Composition}\label{app:details-probing-datasets}

\begin{figure*}[]
    \centering
    \includegraphics[width=0.95\textwidth]{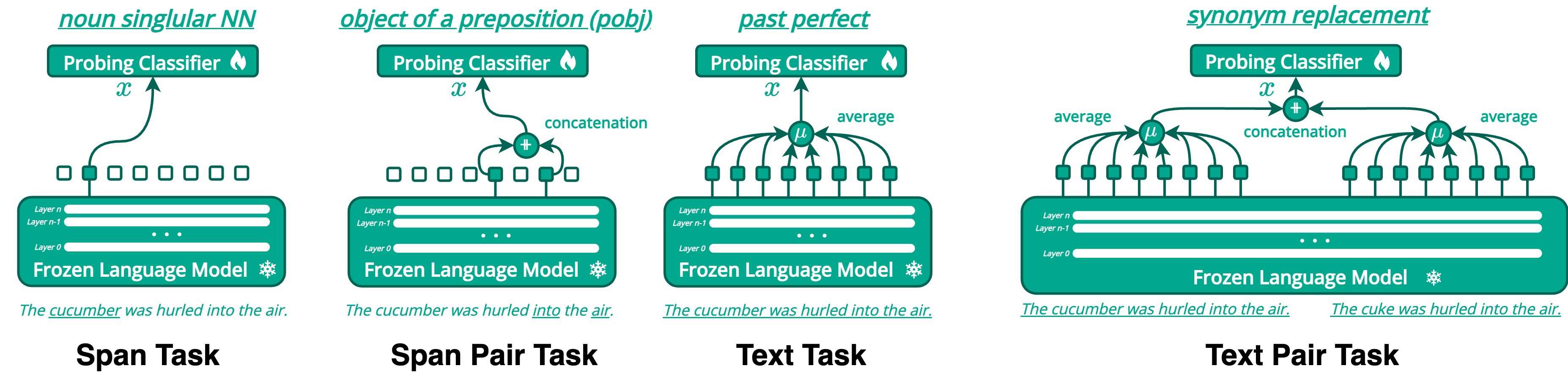}
    \caption{Overview of the composition of the probing input $x$ based on the given text the four types of tasks using concatenating and averaging. In case a tokenizer splits one word into multiple tokens and applies additional averaging operations, such as when probing the part-of-speech phenomenon.} 
    \label{fig:app:tasks}
\end{figure*}

Whenever possible, we rely on established probing datasets and transform instances into a unified format: \textbf{1)} an input $x$ which is either one or a pair of span(s) or sentence(s), including the string and an optional starting and ending index in the context $c$ when task type is either a span or span-pair classification; \textbf{2)} an optional textual context $c$ to encode $x$, for example the sentence in which a span occurs; and \textbf{3)} a corresponding label $y$.
\autoref{fig:app:tasks} shows the composition of the specific probing input $x$ for these four tasks using the internal representation of the last layer of LMs.
Note that additional averaging operations are required if words are tokenized into multiple tokens to get one average vector representing one word, for example, when probing for the part-of-speech tag of a rare word.

If given, we use the original train/dev/test splits.
However, if this division does not exist, we use a 70/10/20 ratio to form these splits. 
Furthermore, we adapted the design of some data to map our dataset format.
Exemplary, for the oLMmpics \citep{10.1162/tacl_a_00342} dataset, we transform the mask-filling tasks into a binary classification where the \textit{correct} label corresponds to a sentence with a correctly filled mask and \textit{incorrect} to a sentence where the mask was filled wrongly. 

\paragraph{OnToNotes}
Following \citet{tenney2018what, tenney-etal-2019-bert}, we use the \textit{OntoNotes} \citep{weischedel2013ontonotes} dataset to derive \textit{part-of-speech tagging}, \textit{constituent labeling}, \textit{named-entity labeling}, \textit{semantic role}, and \textit{co-reference resolution} probing datasets.
Further, we consider with \textit{constituent maximum depth} and \textit{constituent node length} further properties of the constituent tree this dataset \textit{OntoNotes}.

\paragraph{Dependency Corpus}
As in \citet{tenney2018what, tenney-etal-2019-bert}, we use Universal Dependencies annotations of the English Web Treebank to form a \textit{dependency labeling} datasets.

\paragraph{Context Probes}
Presented in \citet{klafka-ettinger-2020-spying}, we compose the nine datasets verifying LMs' knowledge about the context of words. For example, is a word animate (like animals or humans) or inanimate (like buildings or vehicles), or is a verb static or dynamic.

\paragraph{BLiMP Dataset}
Using the data presented in the BLiMP benchmark \citep{warstadt-etal-2020-blimp-benchmark}, we derive 67 probing datasets verifying specific phenomena, like \textit{island effect}, covering \textit{morphology}, \textit{syntax}, and \textit{semantics}.
Unlike the original version, we compose a binary classification task for every phenomenon, either accepting a valid sentence or rejecting one that violates the given linguistic phenomenon.

\paragraph{Zorro Dataset}
As for the BLiMP tasks, we convert the 21 distinct Zorro tasks into a binary classification task on whether a sentence accepts or rejects the given linguistic phenomena is violated.

\paragraph{SemEval-2010 Task 8}
For \textit{semantic relation classification}, we rely on the dataset of \citet{hendrickx-etal-2010-semeval}.

\paragraph{Decompositional Semantics Initiative}
The \textit{Decompositional Semantics Initiative}\footnote{\href{https://decomp.io/}{https://decomp.io/}} provides a large number of datasets to verify semantic phenomena.  
Apart from the common use \textit{semantic proto-roles} \citep{rudinger-etal-2018-neural}, we use their collection of works to compose probing datasets for \textit{factuality} \citep{rudinger-etal-2018-neural-models}, genericity \citep{govindarajan-etal-2019-decomposing}, event structure \citep{vashishtha-etal-2019-fine}, time \citep{vashishtha-etal-2019-fine}, and word sense \citep{white-etal-2016-universal}.

\paragraph{Sentiment Analysis}
We use the commonly used work of \citet{socher-etal-2013-recursive} and form a probing dataset targeting sentiment.

\paragraph{Metaphor}
As in \citet{aghazadeh-etal-2022-metaphors}, we use the data from \citet{mohler-etal-2016-introducing,Birke2006ACA,Steen2010AMF} to form three metaphor datasets.

\paragraph{Complex Word Identification}
We consider word complexity for the first time and use the data presented in \citet{paetzold-specia-2016-semeval}. It provides annotations for different complexity levels of words.

\paragraph{Passive}
We use data from \citet{krasnowska-kieras-wroblewska-2019-empirical} to form a probing dataset assessing knowledge about passive language. 

\paragraph{Synonym / Antonym Replacement}
Using the text of the \textit{OntoNotes} \citep{weischedel2013ontonotes} and Wordnet \citep{wordnet-1995}, we form a probing dataset to detect synonym and antonym replacement. 
Specifically, the binary classification task is: given two texts (the original and an updated one), was the updated one changed by replacing a word with its synonym or antonym?

\paragraph{Negation}
With this work, we verify for the first time \textit{negation} based on human annotated datasets \citep{vahtola-etal-2022-easy, vincze2008bioscope,konstantinova-etal-2012-review}.
Specifically, we form different probing datasets.
\begin{itemize}
    \item Is a text negated or not?
    \item Given two text spans, does the negation within the first one correspond to the second one?
    \item Given a text span, is it the cue or the scope of the negation?
\end{itemize}

\paragraph{oLMmpics}
We form probing datasets addressing different lexical reasoning using the data presented in \citet{10.1162/tacl_a_00342}.
As they provide multiple choices, we form \textit{correct} instances by filling the gap with the correct option and \textit{wrong} ones by filling in the other options. 
Specifically, we form dataset for \textit{always-never}, \textit{age comparison}, \textit{objects comparison}, \textit{antonym-negation}, \textit{multi-hop composition} \textit{property conjunction}, \textit{taxonomy conjunction}, and \textit{encyclopedic composition}.

\paragraph{Bridging}
We rely on the data presented in \citet{pandit-hou-2021-probing} and form two probing datasets. One is to verify whether a text is linguistically applicable, considering bridging (antecedent matches anaphora). And a second one to verify whether an antecedent and anaphora match. 

\paragraph{Discourse Connective}
Using data from \citet{nie-etal-2019-dissent}, we form a probing dataset to assess whether a given connective marker matches the discourse of the given text.

\paragraph{Sentence Order and Next Sentence Prediction}
Following \citet{narayan-etal-2018-dont}, we form two datasets to verify the order of good or badness of a given sentence and whether two sentences occur after each other. 

\paragraph{Discourse Representation Theory}

We use data from \citet{webber2019penn} to compose eight probing datasets addressing \textit{discourse representation theory}: 
\begin{itemize}
    \item Four probing dataset predicting the class of a given span. We distinguish between \textit{implicit}, \textit{explicit}, \textit{implicit-coarse}, and \textit{explicit-coarse}.
    \item The absolute distance, number of words, between two spans in the text.
    \item Whether the order of two spans is correct or not.
    \item Whether two spans have discourse relation or not.
    \item The specific discourse relation of two spans. 
\end{itemize}

\paragraph{Rhetorical Structure Theory}
Using annotations from \citet{carlson-etal-2001,Zeldes2017}, we compose 14 probing datasets addressing \textit{rhetorical theory}. Specifically, we compose the following seven types of datasets for both works:

\begin{itemize}
    \item The rhetorical type of a text span, either nucleus or satellite. 
    \item The number of children of a text span within the rhetorical tree of the text.
    \item The depth of a text span within the rhetorical tree of the text.  
    \item The number of edges between two text spans within the rhetorical tree. 
    \item The specific rhetorical relation between two text spans like \textit{conclusion}.
    \item The relation group of a specific rhetorical relation between two text spans like \textit{evaluation} for the relation \textit{conclusion}.
    \item Whether two text spans occur after each other in the rhetorical tree.
\end{itemize}

\begin{figure*}[]
    \centering
    \includegraphics[width=0.95\textwidth]{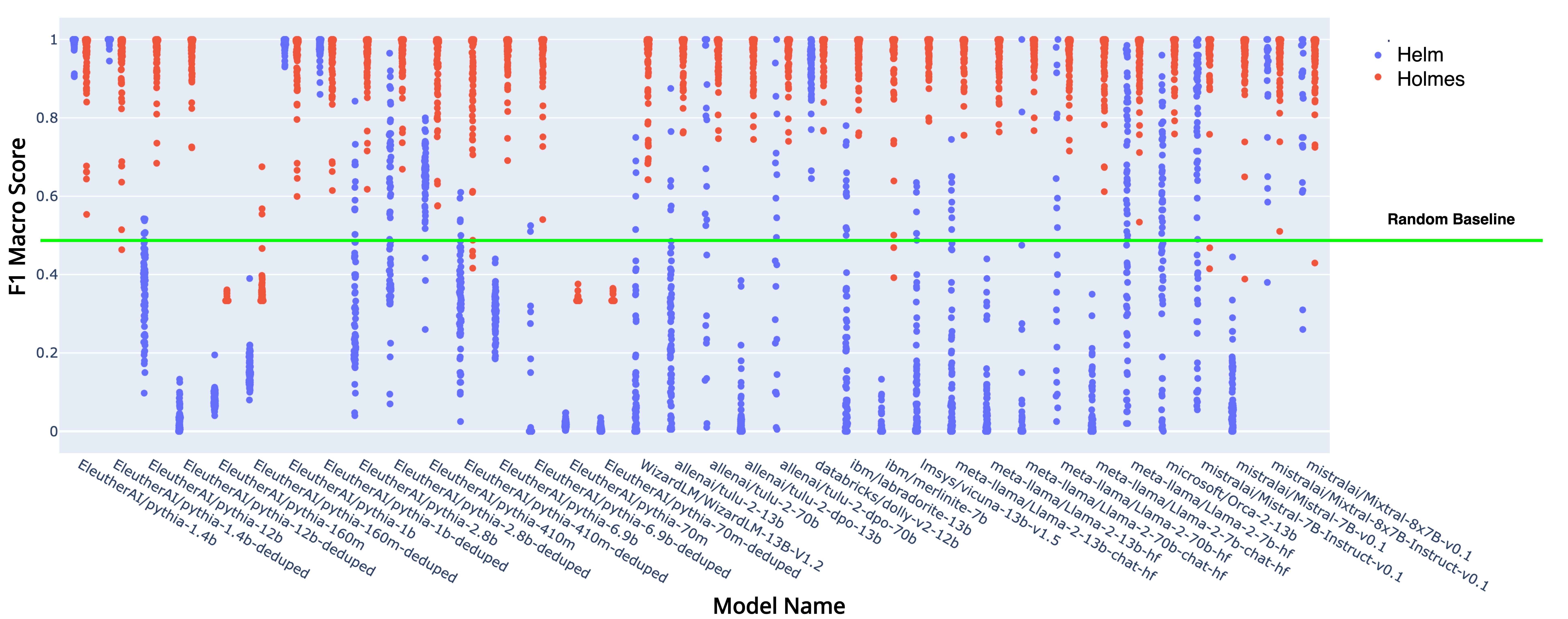}
    \caption{Detailed Holmes vs. HELM \citep{liang2023holistic} comparison for 40 open decoder models and 22 Blimp datasets covering \textit{quantifier}, \textit{island effects}, \textit{irregular forms}, and \textit{binding} phenomena. We use the evaluation code of HELM and run the prompting-based adaption (\textit{multiple joice joined}). The \Holmes{} and Helm results for 40 open decoder models. These results show the advantage of disentangled evaluation (\Holmes{}) over entanglement evaluations (like in HELM), which intertwine the understanding of specific linguistic phenomena and other abilities (like following instructions or answering precisely) in HELM. Most HELM results are below the random baseline, underscoring the necessity to measure linguistic phenomena directly in isolation within LMs.} 
    \label{fig:app:helm-holmes}
\end{figure*}

\begin{table*}[h!]
\centering
\setlength{\tabcolsep}{5pt}
\resizebox{0.9\textwidth}{!}{%

\begin{tabular}{l|c|cccc|ll}\toprule
\bf Phenomena&\bf Illustrative Example & \rotatebox{90}{Text}&  \rotatebox{90}{Text-Pair} & \rotatebox{90}{Span} &  \rotatebox{90}{Span-Pair} &  \rotatebox{90}{\citet{warstadt-etal-2020-blimp-benchmark}}& \rotatebox{90}{\citet{huebner-etal-2021-babyberta}}\\\midrule
 
\rowcolor{lightgray}\textit{anaphor agreement} & Katherine can’t help \underline{herself*}/himself. & 3 & & & & \checkmark&\checkmark\\
\textit{determiner noun agreement} & Craig explored that grocery \underline{store*}/stores. & 10 & & & & \checkmark&\checkmark\\
\rowcolor{lightgray}\textit{irregular forms} & Edward \underline{hid*}/hidden the cats. & 3 & & & & \checkmark&\checkmark\\ 
\textit{subject-verb agreement} & A sketch of lights \underline{does not*}/do not appear. & 10 & & & & \checkmark&\checkmark\\\bottomrule
\end{tabular}
}
\caption{Overview of resources and linguistic phenomena mapping for \textit{morphology}. We give an illustrative example for each phenomenon (*indicates the right option, if options are given) and the number of datasets for the phenomenon by dataset type.}
\label{tab:scenarios-morphology}
\end{table*}

\begin{table*}[h!]
\centering
\setlength{\tabcolsep}{5pt}
\resizebox{0.98\textwidth}{!}{%

\begin{tabular}{l|c|cccc|lllllll}\toprule
\bf Phenomena& \bf Illustrative Example & \rotatebox{90}{Text}&  \rotatebox{90}{Text-Pair} & \rotatebox{90}{Span} &  \rotatebox{90}{Span-Pair} &   \rotatebox{90}{\citet{weischedel2013ontonotes}}& \rotatebox{90}{\citet{silveira-etal-2014-gold}}& \rotatebox{90}{\citet{conneau-etal-2018-cram}}& 
\rotatebox{90}{\citet{flesch1948new}} & \rotatebox{90}{\citet{klafka-ettinger-2020-spying}}& \rotatebox{90}{\citet{warstadt-etal-2020-blimp-benchmark}}& \rotatebox{90}{\citet{huebner-etal-2021-babyberta}}\\\midrule

\rowcolor{lightgray}\textit{argument-structure} &Most cashiers are \underline{disliked*}/flirted.  & 20 &  & & &&&&&&\checkmark&\checkmark\\
\textit{bigram-shift} & What \underline{are you*}/you are doing out there?& &1 & & & &&&\checkmark\\
\rowcolor{lightgray}\textit{binding} & Carlos said that Lori helped \underline{him*}/himself. & 8 &  & & &&&&&&\checkmark&\checkmark\\
\textit{case-subjective-pronoun}& \underline{He brought the pig this suit.*}/The pig brought he this suit. &1  &  & & &&&&&&&\checkmark\\ 
\rowcolor{lightgray}\textit{constituent parsing} & sees Bill $\Rightarrow$ VP & 2 & & & 1 &\checkmark&&&&&&\\
\textit{control/raising}& Julia wasn’t \underline{fun*}/unlikely to talk to. & 5 &  & & &&&&&&\checkmark&\checkmark\\
\rowcolor{lightgray}\textit{deoncausative-inchoative alternation}& \underline{The warden melted the ice.*}/The warden bought the ice. & 1 &  & & &&&&&\checkmark&&\\ 
\textit{dependency parsing} & (into, air) $\Rightarrow$ pobj & & & &1 &&\checkmark\\
\rowcolor{lightgray}\textit{ellipsis} &\underline{He cleans one important book and Stacey cleans a few.*}/He cleans one book and Stacey cleans a few important. & 3 &  & & &&&&&&\checkmark&\checkmark\\
\textit{filler-gap} &\underline{Brett knew what many waiters find.*}/Brett knew that many waiters find. & 9&  & & &&&&&&\checkmark&\checkmark\\ 
\rowcolor{lightgray}\textit{island-effects} & \underline{Which bikes is John fixing?*}/Which is John fixing bikes? & 10 &  & & &&&&&&\checkmark&\checkmark\\
\textit{local attractor} & \underline{Can the access work?*}/ Can the access works? & 1  &  & & &&&&&&&\checkmark\\ 
\rowcolor{lightgray}\textit{object-number} & Oh gods! $\Rightarrow$ Plural & 2& & & &&&\checkmark&&&\checkmark&\\
\textit{part-of-speech}& cucumber $\Rightarrow$ NN (Noun Singular) & & & 3 & & \checkmark&\checkmark\\
\rowcolor{lightgray}\textit{readability} & Curriculums need selling points. $\Rightarrow$ 50.5 (middle) &1 & & & &\checkmark&&&\checkmark&&&\\
\textit{sentence-length} &Oh gods! $\Rightarrow$ 3 words &1 & & & &&&\checkmark\\
\rowcolor{lightgray}\textit{subject-number} &Things are going to be noticed. $Rightarrow$ Plural & 2& & & &&&\checkmark&&&\checkmark&\\
\textit{top-constituent} &Did it all matter? $Rightarrow$ VBD NP VP &1 & & & &&&\checkmark\\
\rowcolor{lightgray}\textit{tree-depth} &Where do you want it? $\Rightarrow$ 6& 1& & & &&&\checkmark&&&&\\\bottomrule
\end{tabular}
}
\caption{Overview of resources and linguistic phenomena mapping for \textit{syntax}. We give an illustrative example for each phenomenon (*indicates the right option, if options are given) and the number of datasets for the phenomenon by dataset type.}
\label{tab:scenarios-syntax}
\end{table*}

\begin{table*}[h]
\centering
\setlength{\tabcolsep}{5pt}
\resizebox{0.95\textwidth}{!}{%
\begin{tabular}{l|c|cccc|lllll}\toprule
\bf  Phenomena& \bf  Illustrative Example & \rotatebox{90}{Text}&  \rotatebox{90}{Text-Pair} & \rotatebox{90}{Span} &  \rotatebox{90}{Span-Pair} &  \rotatebox{90}{\citet{vahtola-etal-2022-easy}}& \rotatebox{90}{\citet{vincze2008bioscope}}& \rotatebox{90}{\citet{konstantinova-etal-2012-review}}& \rotatebox{90}{\citet{morante-blanco-2012-sem}}& \rotatebox{90}{\citet{10.1162/tacl_a_00342}}\\\midrule
 
\rowcolor{lightgray}\textit{age comparison} & \underline{21 years old is older than 35 years fold.*}/21 years old is younger than 35 years fold.  & 1 &  & & &&&&&\checkmark\\
\textit{always-never} & Horses have \underline{always*}/never four legs.& 1 & & & &&&&&\checkmark\\
\rowcolor{lightgray}\textit{antonym negation} & It was \underline{not*}/really hot, it was cold. & 1 & & & &&&&&\checkmark\\
\textit{multi-hop composition} & Comparing a 23, a 38 and a 31 year old, the \underline{last*}/first is oldest. & 1 && & &&&&&\checkmark\\
\rowcolor{lightgray}\textit{negation} & I don't like bananas. $\Rightarrow$ Negation&& 3 & 1 & 2 & 2  &\checkmark&\checkmark&\checkmark&\checkmark\\
\textit{objects comparison} &An airplane is \underline{bigger*}/smaller than a pen. & 1 &  & & &&&&&\checkmark\\
\rowcolor{lightgray}\textit{property conjunction} &A \underline{pen*}/computer is usually located at hand and used for writing. & 1 &  & & &&&&&\checkmark\\
\textit{speculation} &Just about every PC can be upgraded. $\Rightarrow$ Speculation& 1 & & 1 & 1 &&\checkmark\\
\rowcolor{lightgray}\textit{taxonomy connection} &Ferry and floatplane are both \underline{boats*}/airplaines. & 1 & & & &&&&&\checkmark\\ \bottomrule
\end{tabular}
}
\caption{Overview of resources and linguistic phenomena mapping for \textit{reasoning}. We give an illustrative example for each phenomenon (*indicates the right option, if options are given) and the number of datasets for the phenomenon by dataset type.}
\label{tab:scenarios-reasoning}
\end{table*}

\begin{table*}[h]
\centering
\setlength{\tabcolsep}{5pt}
\resizebox{0.95\textwidth}{!}{%

\begin{tabular}{l|c|cccc|llllllllllllllllllllllllllllllll}\toprule
\bf Phenomena & \bf Illustrative Example & \rotatebox{90}{Text}&  \rotatebox{90}{Text-Pair} & \rotatebox{90}{Span} &  \rotatebox{90}{Span-Pair} &     \rotatebox{90}{\citet{weischedel2013ontonotes}} & \rotatebox{90}{\citet{pandit-hou-2021-probing}}& \rotatebox{90}{\citet{nie-etal-2019-dissent}}& \rotatebox{90}{\citet{narayan-etal-2018-dont}} &  \rotatebox{90}{\citet{webber2019penn}} &  \rotatebox{90}{\citet{carlson-etal-2001}} &  \rotatebox{90}{\citet{Zeldes2017}}\\\midrule
\rowcolor{lightgray}\textit{bridging} & The \underline{disease} and symptoms of advanced \underline{infection}. $\Rightarrow$ Valid Bridge & 1 & & & 1 &&\checkmark&&&&&\\
\textit{co-reference resolution} & \underline{National Taiwan University} opened the doors of five of \underline{its} graduate schools. $\Rightarrow$ Valid Co-Reference & & & & 1 &\checkmark\\
\rowcolor{lightgray}\textit{discourse connective} & Leaning against his hip. He reclined with his feet up on the table. $\Rightarrow$ when & & 1 & & &&&\checkmark&&&&\\
\textit{discourse representation theory} & \underline{This is an old story. We're talking about years ago.} $\Rightarrow$ Implicit Relation & & & 8 &&&&&&&\checkmark\\
\rowcolor{lightgray}\textit{next-sentence prediction} & Sentence A, Sentence B $\Rightarrow$ Valid Next Sentence& & 1 & & &&&&\checkmark&&&\\
\textit{rethorical structure theory} & \underline{The statistics} \underline{quoted by the " new " Census Bureau report }$\Rightarrow$ Elaboration & & & 6 & 8 &&&&&\checkmark&&\checkmark\\ 
\rowcolor{lightgray}\textit{sentence order} & Given Sentence B, C, and D $\Rightarrow$ C is at position 2 & 1 & & &&&&&\checkmark&&&\\\bottomrule
\end{tabular}
}
\caption{Overview of resources and linguistic phenomena mapping for \textit{discourse}. We give an illustrative example for each phenomenon (*indicates the right option, if options are given) and the number of datasets for the phenomenon by dataset type.}
\label{tab:scenarios-discourse}
\end{table*}

\begin{table*}[]
\centering
\setlength{\tabcolsep}{5pt}
\resizebox{1.00\textwidth}{!}{%

\begin{tabular}{l|c|cccc|lllllllllllllllllll}\toprule
\bf Phenomena& \bf Illustrative Example & \rotatebox{90}{Text}&  \rotatebox{90}{Text-Pair} & \rotatebox{90}{Span} &  \rotatebox{90}{Span-Pair} &   \rotatebox{90}{\citet{weischedel2013ontonotes}}& \rotatebox{90}{\citet{conneau-etal-2018-cram}} & \rotatebox{90}{\citet{klafka-ettinger-2020-spying}}& \rotatebox{90}{\citet{warstadt-etal-2020-blimp-benchmark}}& \rotatebox{90}{\citet{huebner-etal-2021-babyberta}}& \rotatebox{90}{\citet{hendrickx-etal-2010-semeval}}& \rotatebox{90}{\citet{rudinger-etal-2018-neural}}& \rotatebox{90}{\citet{rudinger-etal-2018-neural-models}}& \rotatebox{90}{\citet{govindarajan-etal-2019-decomposing}}& \rotatebox{90}{\citet{gantt-etal-2022-decomposing}}& \rotatebox{90}{\citet{vashishtha-etal-2019-fine}}& \rotatebox{90}{\citet{white-etal-2016-universal}}& \rotatebox{90}{\citet{socher-etal-2013-recursive}} &\rotatebox{90}{\citet{mohler-etal-2016-introducing}}& \rotatebox{90}{\citet{Birke2006ACA}}& \rotatebox{90}{\citet{Steen2010AMF}}& \rotatebox{90}{\citet{paetzold-specia-2016-semeval}}& \rotatebox{90}{\citet{krasnowska-kieras-wroblewska-2019-empirical}}& \rotatebox{90}{\citet{wordnet-1995}}\\\midrule

\rowcolor{lightgray}\textit{complex word identification} & membrane $\Rightarrow$ Complex, his $\Rightarrow$ Simple& & & 1 & &&&&&&&&&&&&&&&&&\checkmark&&\\
\textit{coordination inversion} & He knew it, and he deserved no answer. $\Rightarrow$ Inversion& 1 & & & &&\checkmark\\
\rowcolor{lightgray}\textit{event structure} & Give them to a library or \underline{burn them}. $\Rightarrow$ Distributive & & & 4 & 2 &&&&&&&&&&\checkmark&&&&&&&&&\\
\textit{factuality} & I \underline{ran} across this item on the Internet. $\Rightarrow$ Factual & & & & 1 &&&&&&&&\checkmark\\
\rowcolor{lightgray}\textit{genericity} & I assume you \underline{mean} the crazy horse memorial. $\Rightarrow$ Not Dynamic & & & 6 & &&&&&&&&&\checkmark&&&&&&&&&&\\
\textit{metaphor} &After all, \underline{morons} pay taxes, too. $\Rightarrow$ Valid Metaphor& & & 4 & &&&&&&&&&&&&&&\checkmark& \checkmark& \checkmark\\
\rowcolor{lightgray}\textit{named-entity labeling} & Paris $\Rightarrow$ City & & & 1 & &\checkmark&&&&&&&&&&&&&&&&&&\\
\textit{negative polarity item licensing} &\underline{Only}/Even Bill would ever complain.&4  &  & & &&&&\checkmark&\checkmark\\ 
\rowcolor{lightgray}\textit{object-animacy} & The rhino fined the pumpkin. $\Rightarrow$ Animate & 1 & & & &&&\checkmark&&&&&&&&&&&&&&&&\\
\textit{object-gender} &The princess uncovered the heiress. $\Rightarrow$ Feminine & 1 & & & &&&\checkmark\\
\rowcolor{lightgray}\textit{passive} & He is considered a European poet through and through. $\Rightarrow$ Passive Sentence& 1 & & & &&&&&&&&&&&&&&&&&&\checkmark&\\
\textit{quantifiers} &There aren’t \underline{many*}/all lights darkening. &6  &  & & &&&&&\checkmark\\ 
\rowcolor{lightgray}\textit{semantic relation classification} & Those cancers were caused by radiation exposures. $\Rightarrow$ Cause-Effect & & 1 & & &&&&&&\checkmark&&&&&&&&&&&&&\\
\textit{semantic proto-roles} & \underline{These look} fine to me. $\Rightarrow$ Exists as physical&  & & & 20  &&&&&&&\checkmark\\
\rowcolor{lightgray}\textit{semantic odd man out} & I wanted to know if it was real or a ploy. $\Rightarrow$ Original & 1 & & & &&\checkmark&&&&&&&&&&&&&&&&&\\
\textit{semantic-role labeling} & And what effect does their return \underline{have} \underline{on campus}? $\Rightarrow$ ARGM-ADV & & & & 1 &\checkmark\\
\rowcolor{lightgray}\textit{sentiment analysis} &You 'll probably love it. $\Rightarrow$ Positive & 1 & & & &&&&&&&&&&&&&\checkmark&&&&&&\\
\textit{subject-animacy} & The turtle betrayed the judge. $\Rightarrow$ Animate  & 1 & & & &&&\checkmark\\
\rowcolor{lightgray}\textit{subject-gender} & The waitress betrayed the judge. $\Rightarrow$ Feminine & 1 & & & &&&\checkmark&&&&&&&&&&&&&&&&\\
\textit{synonym-/antonym-detection} & Is the degree really that \underline{important} $\rightarrow$ \underline{unimportant} to them? $\Rightarrow$ Antonym Replacement & 1 & & & &&&&&&&&&&&&&&&&&&&\checkmark\\ 
\rowcolor{lightgray}\textit{tense} &I quietly snuck up to him and pulled at his sleeve. $\rightarrow$ Present& 2 & & & &&\checkmark&\checkmark&&&&&&&&&&&&&&&&\\
\textit{time} &His mother \underline{was also killed in} the attack. $\Rightarrow$ Minutes & & & 1 & &&&&&&&&&&&\checkmark\\
\rowcolor{lightgray}\textit{verb-dynamic} & The lawyer \underline{found} the judge. $\Rightarrow$ Dynamic Verb& 1 & & & &&&\checkmark&&&&&&&&&&&&&&&&\\ 
\textit{word content} &You mean Alice. $\Rightarrow$ Contains Word \textit{Alice}& 1 & & & &&\checkmark\\
\rowcolor{lightgray}\textit{word sense} &\underline{His mother} was also killed in the attack. $\Rightarrow$ Supersense Noun Person & & & 1 & &&&&&&&&&&&&\checkmark&&&&&&&\\\bottomrule
\end{tabular}
}
\caption{Overview of resources and linguistic phenomena mapping for \textit{semantics}. We give an illustrative example for each phenomenon (*indicates the right option, if options are given) and the number of datasets for the phenomenon by dataset type.}
\label{tab:scenarios-semantics}
\end{table*}

\begin{table*}[t]
\centering
\setlength{\tabcolsep}{5pt}
\resizebox{1.0\textwidth}{!}{%

\begin{tabular}{lccccc}\toprule
\bf Model & \bf Citation  & \bf Size & \bf Pre-Training Objective & \bf Pre-Training Data & \bf Huggingface Tag \\ \toprule

  \rowcolor{lightgray}\multicolumn{6}{c}{\textit{Encoder-Only Language Models}}\\ 
ALBERT & \citet{Lan2020ALBERTAL} & 10 million & MLM+SOP & 16GB & \href{https://huggingface.co/albert-base-v2}{\texttt{albert-base-v2}} \\
BERT & \citet{tenney-etal-2019-bert} & 110 million &   MLM+NSP & 16GB & \href{https://huggingface.co/bert-base-uncased}{\texttt{bert-base-uncased}} \\
DeBERTa & \citet{he2021deberta} & 100	 million & MLM & 80GB & \href{https://huggingface.co/microsoft/deberta-base}{\texttt{microsoft/deberta-base}} \\
DeBERTa-v3 & \citet{he2022debertav3} & 86 million & MLM+DISC & 160GB & \href{https://huggingface.co/microsoft/deberta-v3-base}{\texttt{microsoft/deberta-v3-base}} \\
ELECTRA & \citet{clark2020electra} & 110 million & MLM & 16GB &  \href{https://huggingface.co/google/electra-base-discriminator}{\texttt{google/electra-base-discriminator}} \\
RoBERTa & \citet{Liu2019RoBERTaAR} & 110 million & MLM+DISC & 160GB & \href{https://huggingface.co/roberta-base}{\texttt{roberta-base}} \\ \midrule

  \rowcolor{lightgray} \multicolumn{6}{c}{\textit{Decoder-Only Language Models}}\\ 

GPT2 & \citet{radford_language_2019} & 117 million & LM & 40GB & \href{https://huggingface.co/gpt2}{\texttt{gpt2}} \\
Pythia-70m & \citet{biderman2023pythia} & 70 million & LM & 300 billion tokens & \href{https://huggingface.co/EleutherAI/pythia-70m}{\texttt{EleutherAI/pythia-70m}} \\
Pythia-160m & \citet{biderman2023pythia} & 160 million & LM & 300 billion tokens & \href{https://huggingface.co/EleutherAI/pythia-160m}{\texttt{EleutherAI/pythia-160m}} \\
Pythia-410m & \citet{biderman2023pythia} & 410 million & LM & 300 billion tokens & \href{https://huggingface.co/EleutherAI/pythia-410m}{\texttt{EleutherAI/pythia-410m}} \\
Pythia-1b & \citet{biderman2023pythia} & 1 billion & LM & 300 billion tokens & \href{https://huggingface.co/EleutherAI/pythia-1B}{\texttt{EleutherAI/pythia-1B}} \\
Pythia-1.4b & \citet{biderman2023pythia} & 1.4 billion & LM & 300 billion tokens & \href{https://huggingface.co/EleutherAI/pythia-1.4B}{\texttt{EleutherAI/pythia-1.4B}} \\
Pythia-2.8b& \citet{biderman2023pythia} & 2.8 billion & LM & 300 billion tokens & \href{https://huggingface.co/EleutherAI/pythia-2.8B}{\texttt{EleutherAI/pythia-2.8B}} \\
Pythia-6.9b & \citet{biderman2023pythia} & 6.9 billion & LM & 300 billion tokens & \href{https://huggingface.co/EleutherAI/pythia-6.9B}{\texttt{EleutherAI/pythia-6.9B}} \\
Pythia-12b & \citet{biderman2023pythia} & 12 billion & LM & 300 billion tokens & \href{https://huggingface.co/EleutherAI/pythia-12B}{\texttt{EleutherAI/pythia-12B}} \\

Pythia-70m-dedup & \citet{biderman2023pythia} & 70 million & LM & 207 billion tokens & \href{https://huggingface.co/EleutherAI/pythia-70m-deduped}{\texttt{EleutherAI/pythia-70m-deduped}} \\
Pythia-160m-dedup & \citet{biderman2023pythia} & 160 million & LM & 207 billion tokens & \href{https://huggingface.co/EleutherAI/pythia-160m-deduped}{\texttt{EleutherAI/pythia-160m-deduped}} \\
Pythia-410m-dedup & \citet{biderman2023pythia} & 410 million & LM & 207 billion tokens & \href{https://huggingface.co/EleutherAI/pythia-410m-deduped}{\texttt{EleutherAI/pythia-410m-deduped}} \\
Pythia-1b-dedup & \citet{biderman2023pythia} & 1 billion & LM & 207 billion tokens & \href{https://huggingface.co/EleutherAI/pythia-1B-deduped}{\texttt{EleutherAI/pythia-1B-deduped}} \\
Pythia-1.4b-dedup & \citet{biderman2023pythia} & 1.4 billion & LM & 207 billion tokens & \href{https://huggingface.co/EleutherAI/pythia-1.4B-deduped}{\texttt{EleutherAI/pythia-1.4B-deduped}} \\
Pythia-2.8b-dedup & \citet{biderman2023pythia} & 2.8 billion & LM & 207 billion tokens & \href{https://huggingface.co/EleutherAI/pythia-2.8B-deduped}{\texttt{EleutherAI/pythia-2.8B-deduped}} \\
Pythia-6.9b-dedup & \citet{biderman2023pythia} & 6.9 billion & LM & 207 billion tokens & \href{https://huggingface.co/EleutherAI/pythia-6.9B-deduped}{\texttt{EleutherAI/pythia-6.9B-deduped}} \\
Pythia-12b-dedup & \citet{biderman2023pythia} & 12 billion & LM & 207 billion tokens & \href{https://huggingface.co/EleutherAI/pythia-12B-deduped}{\texttt{EleutherAI/pythia-12B-deduped}} \\

Dolly-v2 & \citet{DatabricksBlog2023DollyV2} & 12 billion & LM+IT & 300 billion token + 15K instructions & \href{https://huggingface.co/databricks/dolly-v2-12b}{\texttt{databricks/dolly-v2-12b}} \\
Llama-2-7b & \citet{touvron2023llama} & 7 billion & LM & 2.4 trillion tokens & \href{https://huggingface.co/meta-llama/Llama-2-7b-hf}{\texttt{meta-llama/Llama-2-7b-hf}} \\
Llama-2-13b & \citet{touvron2023llama} & 13 billion & LM & 2.4 trillion tokens & \href{https://huggingface.co/meta-llama/Llama-2-13b-hf}{\texttt{meta-llama/Llama-2-13b-hf}} \\
Llama-2-70b & \citet{touvron2023llama} & 70 billion & LM & 2.4 trillion tokens & \href{https://huggingface.co/meta-llama/Llama-2-70b-hf}{\texttt{meta-llama/Llama-2-70b-hf}} \\
Llama-2-7b-chat & \citet{touvron2023llama} & 7 billion & LM+IT & 2.4 trillion tokens + 27,5K instructions & \href{https://huggingface.co/meta-llama/Llama-2-7b-chat-hf}{\texttt{meta-llama/Llama-2-7b-chat-hf}} \\
Llama-2-13b-chat & \citet{touvron2023llama} & 13 billion & LM+IT & 2.4 trillion tokens + 27,5K instructions & \href{https://huggingface.co/meta-llama/Llama-2-13b-chat-hf}{\texttt{meta-llama/Llama-2-13b-chat-hf}} \\
Llama-2-70b-chat & \citet{touvron2023llama} & 70 billion & LM+IT & 2.4 trillion tokens + 27,5K instructions & \href{https://huggingface.co/meta-llama/Llama-2-70b-chat-hf}{\texttt{meta-llama/Llama-2-70b-chat-hf}} \\

IBM-Merlinite & \citet{Sudalairaj2024LABLA}   & 7 billion & LM+IT & 2.4 trillion tokens + 1400k instructions & \href{https://huggingface.co/ibm/merlinite-7b}{\texttt{ibm/merlinite-7b}} \\
IBM-Labradorite & \citet{Sudalairaj2024LABLA}   & 13 billion & LM+IT & 2.4 trillion tokens + 1400k instructions & \href{https://huggingface.co/ibm/labradorite-13b}{\texttt{ibm/labradorite-13b}} \\

Vicuna-13b-v1.5 & \citet{heng2023judging}   & 13 billion & LM+IT & 2.4 trillion tokens + 125k instructions & \href{https://huggingface.co/lmsys/vicuna-13b-v1.5}{\texttt{lmsys/vicuna-13b-v1.5}} \\
Orca-2-13b & \citet{mitra2023orca}  & 13 billion & LM+IT & 2.4 trillion tokens + 817K instructions & \href{https://huggingface.co/microsoft/Orca-2-13b}{\texttt{microsoft/Orca-2-13b}} \\
Wizard-13B-v1.2 & \citet{DBLP:journals/corr/abs-2304-12244}  & 13 billion & LM & unknown & \href{WizardLM/WizardLM-13B-V1.2}{\texttt{WizardLM/WizardLM-13B-V1.2}} \\ 
Tülu-2-13b & \citet{wang2023far}  & 13 billion & LM+IT & 2.4 trillion tokens + 330k instructions & \href{https://huggingface.co/allenai/tulu-2-13b}{\texttt{allenai/tulu-2-13b}} \\
Tülu-2-dpo-13b & \citet{wang2023far}  & 13 billion & LM+IT & 2.4 trillion tokens + 330k instructions & \href{https://huggingface.co/allenai/tulu-2-dpo-13b}{\texttt{tulu-2-dpo-13b}} \\
Tülu-2-70b & \citet{wang2023far}  & 70 billion & LM+IT & 2.4 trillion tokens + 330k instructions & \href{https://huggingface.co/allenai/tulu-2-70b}{\texttt{allenai/tulu-2-70b}} \\
Tülu-2-dpo-70b & \citet{wang2023far}  & 70 billion & LM+IT & 2.4 trillion tokens + 330k instructions & \href{https://huggingface.co/allenai/tulu-2-dpo-70b}{\texttt{tulu-2-dpo-70b}} \\

Mistral-7b & \citet{DBLP:journals/corr/abs-2310-06825}  & 7 billion & LM & unknown & \href{https://huggingface.co/mistralai/Mistral-7B-v0.1}{\texttt{mistralai/Mistral-7B-v0.1}} \\ 
Mistral-7b-Inst & \citet{DBLP:journals/corr/abs-2310-06825}  & 7 billion & LM & unknown & \href{https://huggingface.co/mistralai/Mistral-7B-Instruct-v0.1}{\texttt{mistralai/Mistral-7B-Instruct-v0.1}} \\

Mixtral-8x7b & \citet{jiang2024mixtral}  & 47 billion & LM & unknown & \href{https://huggingface.co/mistralai/Mixtral-8x7B-v0.1}{\texttt{mistralai/Mixtral-8x7B-v0.1}} \\ 
Mixtral-8x7b-Inst & \citet{jiang2024mixtral}  & 47 billion & LM & unknown & \href{https://huggingface.co/mistralai/Mistral-7B-v0.1}{\texttt{mistralai/Mistral-7B-v0.1}} \\

\midrule
  \rowcolor{lightgray}\multicolumn{6}{c}{\textit{Encoder-Decoder Language Models}}\\

BART & \citet{lewis-etal-2020-bart} & 121 million & DAE & 160GB & \href{https://huggingface.co/facebook/bart-base}{\texttt{google/facebook/bart-base}} \\

T5-small & \citet{raffel2020exploring} & 60 million & DAE & 800GB & \href{https://huggingface.co/google/t5-small-lm-adapt}{\texttt{google/t5-small-lm-adapt}} \\
T5-base & \citet{raffel2020exploring} & 220 million & DAE & 800GB & \href{https://huggingface.co/google/t5-base-lm-adapt}{\texttt{google/t5-base-lm-adapt}} \\
T5-large & \citet{raffel2020exploring} & 770 million & DAE & 800GB & \href{https://huggingface.co/google/t5-large-lm-adapt}{\texttt{google/t5-large-lm-adapt}} \\
T5-xl & \citet{raffel2020exploring} & 3 billion & DAE & 800GB & \href{https://huggingface.co/google/t5-xl-lm-adapt}{\texttt{google/t5-xl-lm-adapt}} \\
T5-xxl & \citet{raffel2020exploring} & 11 billion & DAE & 800GB & \href{https://huggingface.co/google/t5-xxl-lm-adapt}{\texttt{google/t5-xxl-lm-adapt}} \\

FLAN-T5-small & \citet{raffel2020exploring} & 60 million & DAE+IT & 800GB + 1.8k tasks & \href{https://huggingface.co/google/t5-small-lm-adapt}{\texttt{google/t5-small-lm-adapt}} \\
FLAN-T5-base & \citet{raffel2020exploring} & 220 million & DAE+IT & 800GB + 1.8k tasks & \href{https://huggingface.co/google/t5-base-lm-adapt}{\texttt{google/t5-base-lm-adapt}} \\
FLAN-T5-large & \citet{raffel2020exploring} & 770 million & DAE+IT & 800GB + 1.8k tasks & \href{https://huggingface.co/google/t5-large-lm-adapt}{\texttt{google/t5-large-lm-adapt}} \\
FLAN-T5-xl & \citet{raffel2020exploring} & 3 billion & DAE+IT & 800GB + 1.8k tasks & \href{https://huggingface.co/google/t5-xl-lm-adapt}{\texttt{google/t5-xl-lm-adapt}} \\
FLAN-T5-xxl & \citet{raffel2020exploring} & 11 billion & DAE+IT & 800GB + 1.8k tasks & \href{https://huggingface.co/google/t5-xxl-lm-adapt}{\texttt{google/t5-xxl-lm-adapt}} \\
TK-Instruct & \citet{wang-etal-2022-super} & 11 billion billion & DAE+IT & 800GB + 1.6k tasks & \href{https://huggingface.co/allenai/tk-instruct-11b-def}{\texttt{allenai/tk-instruct-11b-def}} \\
UL2 & \citet{tay2022ul2} & 20 billion & DAE & 800GB & \href{https://huggingface.co/google/ul2}{\texttt{google/ul2}} \\
FLAN-UL2 & \citet{tay2022ul2} & 20 billion & DAE+IT & 800GB + 100k instructions & \href{https://huggingface.co/google/flan-ul2}{\texttt{google/flan-ul2}} \\

\midrule
  \rowcolor{lightgray}\multicolumn{6}{c}{\textit{Static Language Models}}\\ 
Glove-6B & \citet{pennington-etal-2014-glove} & - & WP & 6 billion tokens & \href{https://huggingface.co/sentence-transformers/average_word_embeddings_glove.6B.300d}{\texttt{glove.6B.300d}} \\
Glove-840B & \citet{pennington-etal-2014-glove} & - & WP & 840 billion tokens & \href{https://huggingface.co/sentence-transformers/average_word_embeddings_glove.840B.300d}{\texttt{glove.840B.300d}} \\

\bottomrule
\end{tabular}
}
\caption{Overview of the evaluated LMS covering the corresponding citation, model size, model architecture, pre-training objective \& data, and the Huggingface model tag. Regarding the pre-training objective, we distinguish between masked language modeling (MLM), sentence order prediction (SOP), next sentence prediction (NSP), next word prediction (LM), instruction fine-tuning (IT), word denoising (DAE), and word probabilities from word co-occurrences (WP). For pre-training data, we report known numbers, either as the size of the corpora in gigabytes (GB), the number of pre-training tokens, the number of instructions for fine-tuning, or the number of tasks for instruction fine-tuning.}
\label{tab:models}
\end{table*}

\end{document}